%% file: main.tex
\documentclass{article}
\usepackage{iclr2027_conference,times}

\usepackage{amsmath}
\usepackage{amssymb}
\usepackage{amsthm}
\usepackage{mathtools}
\usepackage{physics}
\usepackage{bm}
\usepackage{empheq}

\usepackage[table]{xcolor}
\usepackage{graphicx}
\usepackage{wrapfig}
\usepackage{subcaption}
\usepackage{booktabs}
\usepackage{multirow}
\usepackage{array}
\usepackage{tabularx}
\usepackage{tikz}

\usepackage{algorithm}
\usepackage{algpseudocode}

\usepackage{enumitem}
\usepackage[most]{tcolorbox}
\usepackage{changepage}
\usepackage{titletoc}

\usepackage{url}
\usepackage{hyperref}
\usepackage{cleveref}

\hypersetup{
    colorlinks=false,
    citebordercolor=green,
    linkbordercolor=red,
    urlbordercolor=cyan
}

\makeatletter
\AtBeginDocument{%
    \let\oldhyper@natlinkstart\hyper@natlinkstart
    \let\oldhyper@natlinkend\hyper@natlinkend

    \renewcommand{\hyper@natlinkstart}[1]{%
        \begingroup
        \color{green!30!black}%
        \oldhyper@natlinkstart{#1}%
    }

    \renewcommand{\hyper@natlinkend}{%
        \oldhyper@natlinkend
        \endgroup
    }
}
\makeatother

\input{math_commands}

\input{custom_notation}

\definecolor{stageblue}{HTML}{4C78A8}

\definecolor{takeblue}{HTML}{2B5C8A}
\definecolor{takebluebg}{HTML}{F1F6FA}
\definecolor{pastelblue}{HTML}{BFD7EA}

\definecolor{remarkgreen}{HTML}{2F8F6B}
\definecolor{remarkbg}{HTML}{EEEEEB}

\definecolor{controlblue}{RGB}{220,235,247}
\definecolor{assemblygreen}{RGB}{226,239,218}
\definecolor{rewardpink}{RGB}{252,235,235}
\definecolor{controlleryellow}{RGB}{255,242,204}
\definecolor{boxgray}{RGB}{238,238,238}
\definecolor{ingredientblue}{HTML}{D9EAF5}

\newtcolorbox{takehome}{
    enhanced,
    frame hidden,
    colback=takebluebg,
    fontupper=\small,
    arc=0mm,
    outer arc=0mm,
    boxsep=0pt,
    left=7pt,
    right=7pt,
    top=5pt,
    bottom=5pt,
    width=\linewidth,
    opacityback=0.45,
    borderline west={2.5pt}{0pt}{takeblue}
}

\theoremstyle{remark}
\newtheorem{remark}{Remark}

\tcolorboxenvironment{remark}{
    enhanced,
    breakable,
    colback=remarkbg,
    colframe=remarkbg,
    boxrule=0pt,
    arc=0pt,
    left=9pt,
    right=8pt,
    top=7pt,
    bottom=7pt,
    before skip=6pt,
    after skip=6pt,
    borderline west={2.5pt}{0pt}{remarkgreen}
}

\newtcolorbox{eqbox}{
    colback=boxgray,
    colframe=boxgray,
    boxrule=0pt,
    arc=0pt,
    left=2pt,
    right=2pt,
    top=2pt,
    bottom=2pt,
    before skip=5pt,
    after skip=3pt,
}

\newtcolorbox{researchquestion}{
    enhanced,
    sharp corners,
    frame hidden,
    boxrule=0pt,
    colback=black!8,
    borderline west={1.5pt}{0pt}{black!55},
    left=8pt,
    right=8pt,
    top=5pt,
    bottom=5pt,
    before skip=7pt,
    after skip=7pt
}

\newcommand{\mathhl}[2]{%
    \begingroup
    \setlength{\fboxsep}{1.5pt}%
    \colorbox{#1}{$\displaystyle #2$}%
    \endgroup
}

\newcommand{\hlblue}[1]{\mathhl{controlblue}{#1}}

\newcommand{\hlpink}[1]{\mathhl{rewardpink}{#1}}
\newcommand{\hlyellow}[1]{\mathhl{controlleryellow}{#1}}

\newcommand{\figPone}{\textcolor{blue!70!black}{\ensuremath{\bullet}}}
\newcommand{\figPtwo}{\textcolor{orange!90!black}{\ensuremath{\bullet}}}
\newcommand{\figPthree}{\textcolor{violet!80!black}{\ensuremath{\bullet}}}
\newcommand{\figPfour}{\textcolor{teal!80!black}{\ensuremath{\bullet}}}

\newcommand{\uncontrolledpath}{%
    \textcolor{gray}{%
        \raisebox{0.5ex}{%
            \rule{0.12cm}{1.0pt}\hspace{1pt}\rule{0.12cm}{1.0pt}%
        }%
    }%
}

\newcommand{\controlledpath}{%
    \textcolor{gray}{%
        \raisebox{0.5ex}{\rule{0.25cm}{1.0pt}}%
    }%
}

\newcommand{\colorsquare}[1]{%
    \raisebox{0.1ex}{%
        \colorbox{#1}{\rule{0pt}{0.8ex}\rule{0.8ex}{0pt}}%
    }%
}

\newcommand{\ingredient}[1]{%
    \begingroup
    \setlength{\fboxsep}{1.5pt}%
    \mbox{%
        \colorbox{ingredientblue}{%
            \textbf{(#1)}%
        }%
    }%
    \endgroup
}

\crefname{algorithm}{Algorithm}{Algorithms}
\Crefname{algorithm}{Algorithm}{Algorithms}

\newcommand{\bluecomment}[1]{%
    \hspace{0.6em}%
    {\scriptsize\color{stageblue}\#~#1}%
}

\newcommand{\algphase}[1]{%
    \Statex \textcolor{blue}{\textit{#1}}%
}

\newcommand{\method}{\textsc{CMDS}}

\newcommand{\discadj}{\textsc{Disc. Adj.}}
\newcommand{\adjmatch}{\textsc{Adj. Match.}}

\newcommand{\Law}{\mathrm{Law}}

\iclrfinalcopy 

\newcommand{\equalcontrib}{\textsuperscript{*}}

\newcommand{\authorcell}[3]{%
    \parbox[t]{0.29\textwidth}{%
        \centering\normalfont
        \textbf{#1}\\[0.25em]
        #2%
        \if\relax\detokenize{#3}\relax
        \else\\[-0.05em]#3%
        \fi
    }%
}

\title{
    One for All, All for One: \\
    Coordinated Multi-Agent Diffusion Steering \\
    via Stochastic Optimal Control \\[0.25em]
}

\author{%
\begin{tabular}{@{}c@{\hspace{1.8em}}c@{\hspace{1.8em}}c@{}}
\authorcell{Riccardo Barbano\equalcontrib}{UCL}{}
&
\authorcell{Vincent Pauline\equalcontrib}{TUM \& MCML}{}
&
\authorcell{Runchang Li\equalcontrib}{CUHK}{}
\\[1.6em]
\authorcell{George Webber}{KCL}{}
&
\authorcell{Alexander Denker}{DESY}{}
&
\authorcell{\v{Z}eljko Kereta}{UCL}{}
\\[1.6em]
\authorcell{Stefan Bauer}{TUM \& MCML \& CIFAR}{}
&
\authorcell{Francisco Vargas}{Proxima Bio}{}
&
\authorcell{Esmeralda S. Whitammer}{University of Edinburgh \& CIFAR}{}
\end{tabular}%
}

\begin{document}

\maketitle

\begingroup
\makeatletter
\renewcommand{\thefootnote}{\fnsymbol{footnote}}
\renewcommand{\@makefntext}[1]{\noindent\@makefnmark\,#1}
\makeatother

\begingroup
\renewcommand{\thefootnote}{\fnsymbol{footnote}}
\footnotetext[1]{%
Equal contribution.\\ Correspondence to R.B. (\texttt{r.barbano@ucl.ac.uk}) and V.P. (\texttt{vincent.paulinef@gmail.com}).%
}
\endgroup

\vspace{-0.7em}

\begin{abstract}
    \input{_abstract}
\end{abstract}

\section{Introduction}
\input{_main}

\pagebreak

\subsection*{AI use statement}

In this work, we used generative AI tools to assist with text editing, reformatting, and polishing; \LaTeX{} formatting and styling;
and software-code review, debugging, and editing.
We also used generative AI to generate code for visualisation and presentation of results; all such code was reviewed by the authors and was not used to implement the methods or produce the experimental results reported in this paper.
All AI-assisted text and code were reviewed and edited by the authors.
AI-assisted code was manually inspected and tested before use.

We did not use generative AI tools to generate synthetic datasets, help develop theoretical models or conceptual frameworks, formulate mathematical claims, provide critical ingredients for proving mathematical claims, assist in the writing of proofs, propose or refine hypotheses, clean and reformat datasets, support qualitative and thematic data analysis, or interpret results; assisting with translation is not applicable to this work.

We take responsibility for the final content of this work, including text, claims, code, and artifacts produced with the aid of generative AI.


\section*{Acknowledgments}
RB acknowledges support from the EPSRC (EP/V026259/1).
GW acknowledges support from the EPSRC CDT in Smart Medical Imaging [EP/S022104/1] and by a GSK Studentship.
AD acknowledges support from DESY (Hamburg, Germany), a member of the Helmholtz Association HGF. 
VP acknowledges the financial support from the Munich Center for Machine Learning (MCML).
The authors gratefully acknowledge the Gauss Centre for Supercomputing e.\,V. (\url{https://www.gauss-centre.eu}) for funding this project by providing computing time through the John von Neumann Institute for Computing (NIC) on the GCS Supercomputer JUPITER---JUWELS at J\"ulich Supercomputing Centre (JSC). 
Furthermore, the authors appreciate the computational resources provided by the National High Performance Computing Centre
(\url{https://www.nhr.kit.edu}). 
Finally, the authors would also like to acknowledge the fruitful discussions with Lourdes de Agapito Vicente, Simon Arridge, Mirgahney Mohammed, Marcelo Pereyra, and Neill Campbell.


\bibliography{_refs}
\bibliographystyle{iclr2027_conference}


\newpage
\appendix
\startcontents[appendix]

\crefalias{section}{appendix}
\crefalias{subsection}{appendix}
\crefalias{subsubsection}{appendix}
\crefname{appendix}{Appendix}{Appendices}
\Crefname{appendix}{Appendix}{Appendices}


\begingroup
\hypersetup{linkbordercolor={1 0.8 0}}
\section*{Appendix Overview}
\printcontents[appendix]{}{1}{\setcounter{tocdepth}{2}}
\endgroup


\input{_related_work_apdx}
\input{_theory_apdx}

\newpage
\input{_exp_apdx}

\end{document}

%% file: math_commands.tex
\def\1{\bm{1}}

\DeclareMathAlphabet{\mathsfit}{\encodingdefault}{\sfdefault}{m}{sl}
\SetMathAlphabet{\mathsfit}{bold}{\encodingdefault}{\sfdefault}{bx}{n}

\def\sP{{\mathbb{P}}}

\newcommand{\E}{\mathbb{E}}

\newcommand{\R}{\mathbb{R}}

\newcommand{\KL}{D_{\mathrm{KL}}}



%% file: custom_notation.tex
\newcommand{\Rd}{\R^{d}}
\newcommand{\TT}{[0,1]}

\newcommand{\N}{\mathcal{N}}

\newcommand{\pt}{p_t}

\renewcommand{\grad}{\nabla}

\NewDocumentCommand{\scorenet}{ o o o }{%
  \ensuremath{%
    s_{\theta}%
    \IfValueT{#1}{^{#1}}%
    \IfValueTF{#2}
      {%
        \IfValueTF{#3}
          {\!\left(#2,#3\right)}%
          {\!\left(#2\right)}%
      }%
      {}%
  }%
}

\newcommand{\W}{W}

\NewDocumentCommand{\bdrift}{o}{%
  \ensuremath{%
    \bar{b}%
    \IfNoValueTF{#1}{}{(#1)}%
  }%
}


%% file: _abstract.tex
Deep generative models often produce structured outputs composed of interacting components.
Modelling these outputs with a single model requires learning both the component distributions and their interactions.
We pursue a modular alternative: reuse independently trained component generators and learn only how to coordinate them to produce coherent structured outputs.
Our framework, \textbf{C}oordinated \textbf{M}ulti-Agent \textbf{D}iffusion \textbf{S}teering (\method{}), treats frozen pretrained diffusion models as \textit{reusable generative primitives} and coordinates their reverse processes through a learned \textit{control}.
We formulate coordination as a stochastic optimal control problem, balancing an assembly-level reward that specifies the desired properties of the combined output against deviations from the pretrained dynamics.
The learned control amortises this optimisation, allowing reuse across new task instances.
Experiments show that \method{} can recover a known target distribution, satisfy different spatial constraints with the same trained control, and recover individual sources from degraded mixtures.
Across multi-agent maze navigation, articulated robot planning, and text-conditioned human motion, \method{} turns frozen models into coordinated multi-agent generators.

%% file: _main.tex
Producing coherent structured outputs involves two distinct problems: modelling individual components and coordinating them when they are used together.
End-to-end generative models absorb both into a single generative distribution.
We instead build on a long-standing tradition of modular and compositional generative modelling \citep{grenander1978pattern,bienenstock1996compositionality,williams2025structured,pmlr-v235-du24d}.
Component generators, pretrained without joint-system supervision, capture component-level variability, while coordination is learned separately at the level of the structured output.
Diffusion models \citep{ho2020denoising,song2020score,lai2025principles} are well suited to this decomposition, as each pretrained reverse process provides a reusable component generative prior whose dynamics can be steered via a learned control while the underlying model remains fixed.
Prior work on compositional diffusion generation typically resolves the required coordination at sampling time for each generated instance \citep{liu2022compositional,yang2023compositional,cao2025modality}.
This is sensible for one-off tasks, but becomes costly when related coordination problems recur across many instances.
In settings where related coordination problems recur across many instances--for example, across relational design specifications, source-separation observations, or multi-agent start-goal configurations--the associated computation can instead be amortised into a reusable control.
Building on recent work showing how controls or amortised samplers can steer pretrained diffusion processes towards task-specific target distributions
\citep{domingo2024stochastic,domingoenrich2025adjoint,venkatraman2024amortizing}, we thus ask:
\begin{researchquestion}
\itshape
How can we learn an amortised control that coordinates frozen component diffusion models using a reward on their assembled output, while
remaining close to their pretrained dynamics?
\end{researchquestion}
\paragraph{A framework for coordinated generation.}
The proposed framework begins with independently trained component generators, whose distributions form a reference product distribution over component configurations.
An assembly map combines their outputs into a candidate, while an assembly-level reward evaluates the candidate.
A learned control then coordinates the component generators towards high-reward joint outputs.
To make this concrete, consider planning the movements of several agents through a shared environment.
Each frozen copy of a single-agent trajectory model generates a path through the environment, using an agent-specific start and end as conditioning inputs.
Stacking these paths produces a candidate multi-agent plan.
Although each path may avoid walls in isolation, the assembled plan may still be invalid if agents collide.
This illustrates the role of the assembly-level reward. 
It expresses requirements not captured by the independently trained component models and provides the signal used to coordinate their generation.
Generally, the framework requires \emph{no training examples of valid structured outputs}, only \ingredient{i}~component-level data or pretrained component generators; \ingredient{ii}~an assembly map; and \ingredient{iii}~an assembly-level reward.

\paragraph{Continuous-time generative modelling.}
For each component \(i\), let \(G^i\) denote the generator induced by the reverse-time sampling process of a pretrained diffusion model \citep{anderson1982reverse,haussmann1986time,song2020score}.
We write \(\hat X_s^i\in\mathbb R^d\) for component \(i\)'s state and \(\hat X_s=(\hat X_s^1,\ldots,\hat X_s^N)\in\mathbb R^D\), \(D=Nd\), for the joint state; hats denote reverse-time quantities.
We use \(s \in [0,1]\) for reverse (generative) time with \(s=0\) corresponding to the noise state and \(s=1\) to the generated sample.
The reverse process for component \(i\) starts from \(\hat X^i_0\sim \hat{p}^i_0\), where \(\hat{p}^i_0\) is a standard Gaussian, and terminates at \(\hat X^i_1\sim\tilde{\pi}^i\), where \(\tilde{\pi}^i\) denotes the learned component distribution.
Suppressing the index \(i\), we write the reference reverse-time diffusion as,
\begin{equation}
\label{eqn:reverse_SDE}
    \dd \hat X_s
    =
    \hat f_s(\hat X_s)\,\dd s
    +
    \hat\sigma_s\,\dd \bar W_s,
    \qquad
    \hat X_0\sim \hat{p}_0.
\end{equation}
Here, \(\hat f_s\) denotes the learned reverse drift, \(\hat\sigma_s\) is the reverse-time diffusion scale, and \(\bar W_s\) is a Brownian motion in reverse time.
The terminal law of \(\hat X_1\) is the learned model distribution \(\tilde\pi\), which approximates the corresponding exact terminal law \(\pi\).
For \(N\) pretrained component samplers, their independent reference processes form the joint product reverse process
\(\hat X_s=(\hat X^1_s,\ldots,\hat X^N_s)\), with \(\hat X^i_0\sim \hat{p}^i_0\) and \(\hat X^i_1\sim\tilde\pi^i\).
Its terminal marginal is therefore the product law
\(\rho=\tilde\pi^1\otimes\cdots\otimes\tilde\pi^N.\)
\paragraph{Controlled diffusion and stochastic optimal control (SOC).}
The connection between stochastic control, path-space relative entropy, and exponential tilting is well established \citep{chetrite2015variational}.
Diffusion sampling admits a SOC interpretation in which a reward tilt is realised by controlling a reference diffusion under a quadratic path-action penalty \citep{nusken2021solving,berner2024optimal,uehara2024fine,domingoenrich2025adjoint}; see \cref{theory_path_view_coord}.
For a single diffusion model (\(N=1\)), the problem can be written as,
\begin{equation}
\label{eq:soc-control-problem}
\begin{aligned}
\min_{u \in \mathcal U}
\quad
&\mathbb E_{\hat{\mathbb P}^{u}}
\left[
    \int_0^1
    \frac12 \|u_s(\hat X^u_s)\|^2 \, \dd s
    -
    \lambda r(\hat X^u_1)
\right],
\\
\text{s.t.}\quad
&\dd\hat X^u_s
=
\Bigl[
    \hat f_s(\hat X^u_s)
    +
    \hat\sigma_{s}u_s(\hat X^u_s)
\Bigr]\dd s
+
\hat\sigma_{s}\dd \bar W_s,
\qquad
\hat X^u_0\sim \hat{p}_0.
\end{aligned}
\end{equation}
Here, \(u \in \mathcal U\) is an admissible control field, \(\hat{\mathbb P}^{u}\) is the path measure induced by the controlled process \(\hat X^u\), \(\lambda>0\) controls the reward strength, \(\hat f_s\) and \(\hat\sigma_s\) denote the reverse-time drift and diffusion scale of the pretrained model, respectively, \(\bar W_s\) is a Brownian motion, and \(\hat{p}_0\) is the initial noise distribution.
Under the Girsanov conditions and a shared initial law, the expected accumulated quadratic control cost equals the relative entropy between the controlled and reference path measures \citep{nusken2021solving,berner2024optimal}.
We apply this SOC formulation to the joint state \(\hat X_s\) of multiple frozen pretrained component diffusion models, using an assembly-level reward.

\textbf{Contributions.} 
Our contributions can be summarised as follows:
\begin{itemize}
    \item \textbf{Coordination as modular post-training.}
    We formulate compositional generation as stochastic optimal control over the joint reverse dynamics of frozen pretrained component diffusion models.
    The independent component processes define a reference product law, while an assembly-level reward can induce cross-component dependencies, realised by a learned control that couples their reverse dynamics.
    \item \textbf{Comprehensive empirical validation.}
    We demonstrate target-law recovery, amortised relational assembly, and source separation.
    In PointMaze, \method{} coordinates up to four agents and generalises to unseen start--goal regions.
    Using off-the-shelf KUKA and MDM checkpoints, it turns frozen single-agent models into coordinated multi-agent generators without joint training data.
    The learned KUKA control transfers to new geometries and additional arms without retraining.
    In four-person motion generation, \method{} achieves 90\% joint success versus 75\% for PCD++, with over $250\times$ faster sampling.
\end{itemize}
Further discussion on relevant background concepts can be found in Appendix~\ref{app:related-work}.
\section{Coordinated Multi-Agent Diffusion Steering (\method{})}
\begin{figure*}
    \centering
    \includegraphics[width=\linewidth]{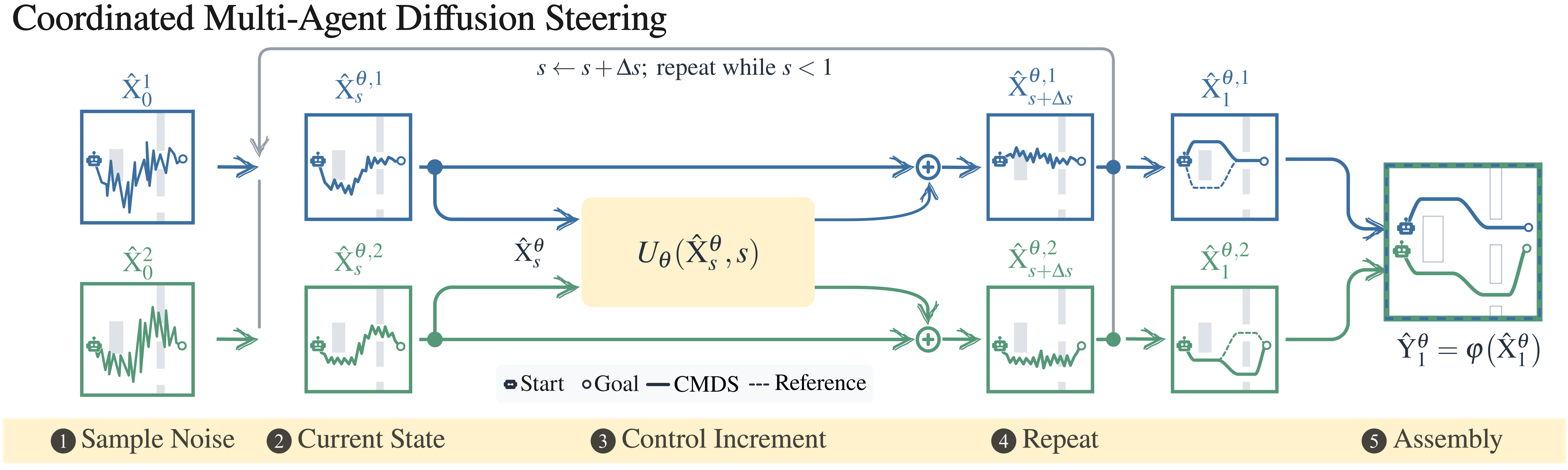}
    \caption{
    \textbf{Overview of \method{}.}
    Frozen component diffusion models are sampled concurrently, while a joint control conditions on the multi-agent state and applies component-wise controls throughout the reverse process, yielding a coordinated assembled output.
    }
    \label{fig:cmds-overview}
    \vspace{-1em}
\end{figure*}
We consider \(N\) pretrained diffusion agents, each generating one component of a structured output.
Using the joint state \(\hat X_s\) defined above, the assembled state is \(\hat Y_s=\varphi(\hat X_s) \in \mathbb{R}^m\), where
\(\varphi:\mathbb R^D\to\mathbb R^m\) is a fixed assembly operator.
Without coordination, the component processes evolve independently and need not produce a globally consistent assembly.
We extend the single-process SOC formulation in \cref{eq:soc-control-problem} to the joint state, with \(u_s:\mathbb R^D\to\mathbb R^D\), denoting the coordination control.

Importantly, \method{} is non-autoregressive across components.
All \(N\) reverse processes evolve concurrently along the same generative-time axis, and at each time \(s\), the control
\(u_s^i(\hat X_s)\) applied to agent \(i\) may depend on the current state of every agent.
No component is therefore generated and fixed before the others.
This avoids imposing an arbitrary ordering over components and enables corrections throughout generation, which is particularly important when the assembly-level criterion depends on their joint configuration.
The component updates may also be evaluated in parallel, subject to the cost of the coordination modules.

With this joint representation, coordination amounts to steering the combined reverse process so that its assembled terminal sample satisfies an assembly-level criterion while remaining close to the independent pretrained dynamics.
Writing \(x = (x^1, \cdots, x^N)\in \mathbb{R}^D\), we denote the joint reference reverse drift by
\(
\hat f_s(x)
=
(\hat f_s^1(x^1),\ldots,\hat f_s^N(x^N)).
\)
The controlled terminal assembly is
\(
\hat Y_1^u
=
\varphi(\hat X_1^u)
=
\varphi(\hat X_1^{u,1},\ldots,\hat X_1^{u,N}),
\)
and we apply the SOC objective in \cref{eq:soc-control-problem} to the reference product process, using the assembly-level reward \(r\circ\varphi\), 
\begin{eqbox}
    \begin{equation}\label{eq:multi_agent_soc}
        \min_{u\in \mathcal U} \quad
        \mathbb E_{\hat{\mathbb P}^{u}}
        \left[
        \int_0^1
        \hlblue{
        \frac{1}{2}
        \sum_{i=1}^N
        \left\|
        u_s^i(\hat X_s^u)
        \right\|^2
        }
        \,\mathrm ds
        -
        \hlpink{
        \lambda r\!\left(
        \varphi(\hat X_1^u)
        \right)
        }
        \right],
    \end{equation}
\end{eqbox}
where the \colorsquare{rewardpink}\, reward of the assembled terminal output measures the global consistency, and the \colorsquare{controlblue}\, quadratic control cost penalises deviation from the pretrained agents' distribution.
The central hypothesis of \method{} is that locally valid components can be learned, while the dependencies required for global validity can be introduced through an assembly-level criterion.
Further details on the forward noising process, its time reversal, the associated score-based parameterisation, and the controlled reverse-time SDE are provided in \Cref{apdx:more_theory_DM,apdx:controlled_time_reverse_SDE}.

\subsection{Learning Coordinated Reverse Dynamics}\label{method:multi_agent_adjoint}
The exact optimal path distribution $\hat{\mathbb P}^{u^\star}$ is generally intractable.
We therefore approximate the non-parametric optimal control $u^\star_s(x)$ by learning a \colorsquare{controlleryellow}\, parametrised neural control field,
\[
    \hlyellow{U_\theta(x,s)}
    =
    \left(U_\theta^1(x,s),\ldots,U_\theta^N(x,s)\right).
\]
The component controls may either share parameters across agents or use agent-specific parameters; in both cases, each \(U_\theta^i\) conditions on the full joint state.
The resulting controlled reverse diffusion implicitly defines a stochastic generator $G_\theta=(G_\theta^1,\ldots,G_\theta^N)$, with component dynamics,
\begin{eqbox}
\begin{align}
    \dd \hat X_s^{\theta,i}
    =
    \left[
    \hat f^i_s(\hat X_s^{\theta,i})
    +
    \hat\sigma_{s}\hlyellow{U_{\theta}^i(\hat X_s^{\theta},s)}
    \right]\dd s
    +
    \hat\sigma_{s}\,\dd\bar{\W}_s^i,
    \qquad
    i=1,\ldots,N.
\end{align}
\end{eqbox}
We denote the corresponding assembled process by \(\hat Y_s^\theta:=\varphi(\hat X_s^\theta)\).
Equivalently, \(G_\theta^i:\hat X_0\mapsto \hat X_1^{\theta,i}\) denotes the stochastic solution map induced by the controlled reverse dynamics, with \(\hat X_0^i\sim \hat{p}_0^i\).
The corresponding training objective balances control cost and global reward,
\begin{align}
    \mathcal{J}(\theta)
    =
    \mathbb{E}_{\hat{\mathbb P}^{\theta}}
    \left[
    \int^1_0
    \frac12\sum_{i=1}^N
    \|U_{\theta}^i(\hat X_s^\theta,s)\|^2
    \dd s
    -
    \lambda r(\varphi(\hat X^\theta_1))
    \right],
    \label{eq:soc-objective}
\end{align}
where $\hat{\mathbb P}^{\theta}$ is the path distribution induced by the parametrised controlled process.
\paragraph{Adjoint matching \method{} instantiation \citep{domingoenrich2025adjoint}.}
To train the control efficiently, we use an adjoint matching (AM) version of the original objective.
Along sampled controlled trajectories, an adjoint variable estimates the direction in which each component should move to improve the final global reward.
The control is trained to match this optimal correction,
\begin{eqbox}
\begin{align}
    \mathcal L_{\rm AM}(\theta)
    =
    \frac12
    \E_{\hat{\sP}^{\bar{\theta}}}\left[
    \int_0^1
    \left\|
    \hlyellow{U_{\theta}(\hat X_s^{\bar{\theta}}, s)}
    +
    \hat \sigma_s a_s^{\bar{\theta}}
    \right\|^2
    \,\dd s
    \right], 
    \qquad
    \bar{\theta}=\operatorname{stopgrad}(\theta)
    \label{eq:cmad-basic-am}
\end{align}
\end{eqbox}
where \(a_s^{\bar\theta}\) denotes the adjoint along the sampled controlled trajectory \(\hat X^{\bar\theta}\), obtained by solving the following terminal-value problem backwards from \(s=1\) to \(s=0\), 
\begin{equation}
\label{eq:cmad-adjoint}
    \begin{aligned}
    \frac{\dd}{\dd s}a^{\bar{\theta}}_s
    &=
    -D_x\left(\hat f_s(\hat X_s^{\bar{\theta}}) + \hat\sigma_s U_{\bar{\theta}}(\hat X_s^{\bar{\theta}}, s)\right)^\top a_s^{\bar{\theta}}
    -\frac12\grad_x\sum_{i=1}^N\|U_{\bar{\theta}}^i(\hat X^{\bar{\theta}}_s,s)\|^2,\\
    \quad
    a^{\bar{\theta}}_1
    &=
    -\lambda
    D_x\varphi(\hat X^{\bar{\theta}}_1)^\top
    \grad_y r(\hat Y^{\bar{\theta}}_1),
    \notag
    \end{aligned}
\end{equation}
where $D_x$ denotes the Jacobian with respect to $x$.
The full pathwise adjoint, its block-wise multi-agent form, and the corresponding adjoint matching surrogate are derived
in \Cref{apdx:adjoint_matching}.

\paragraph{Strategies for learning the coordination control.}
\method{} is defined by jointly controlling the reference product process under the SOC objective in \cref{eq:multi_agent_soc}, whereas AM is one possible strategy for learning the \colorsquare{controlleryellow}\, coordination control \(U_\theta\), rather than part of the framework.
When the reverse solver and assembly-level reward are differentiable, \(U_\theta\) may instead be learned by directly optimising the discretised SOC objective; we refer to reverse-mode differentiation through the solver transitions as \emph{discrete-adjoint} optimisation \citep{domingo2024taxonomy}.
Alternatively, path-space objectives such as relative trajectory balance \citep{venkatraman2024amortizing} could be adapted to the joint controlled reverse process to target the same reward-tilted law, provided that the corresponding controlled-to-reference trajectory-density ratio is tractable.
We adopt AM since it propagates gradient information without backpropagating through the complete computational graph of the sampling rollout.

\subsection{Path-Space Interpretation of Coordination}\label{theory_path_view_coord}
The SOC objective in \cref{eq:multi_agent_soc} admits a path-space interpretation.
Let \(\hat{\mathbb P}\) denote the reference path law induced by the independent pretrained reverse diffusion processes, and let \(\hat{\mathbb P}^{u}\) denote the controlled path law.
By Girsanov's theorem \citep{oksendal2003stochastic}, under the usual change-of-measure conditions and a shared initial law, the quadratic control cost equals the path-space relative entropy,
\(
D_{\mathrm{KL}}
\bigl(
\hat{\mathbb P}^{u}\,\|\,\hat{\mathbb P}
\bigr)
=
\frac{1}{2}
\mathbb E_{\hat{\mathbb P}^{u}}
[
\int_0^1
\|
u_s(\hat X_s^u)
\|^2
\,\dd s
].
\)
The control problem admits the path-space form,
\begin{eqbox}
\begin{equation}
    u^\star
    \in
    \arg\min_{u\in \mathcal U}
    \left\{
        \hlblue{D_{\mathrm{KL}}
        \bigl(
            \hat{\mathbb P}^{u}
            \,\Vert\,
            \hat{\mathbb P}
        \bigr)
        }
        -
        \hlpink{
        \lambda
        \mathbb E_{\hat{\mathbb P}^{u}}
        \left[
            r\!\left(\varphi(\hat X_1^{u})\right)
        \right]
        }
    \right\}.
    \label{eq:path_space_objective}
\end{equation}
\end{eqbox}
This formulation makes explicit that coordination balances the assembly-level reward of the assembled terminal object against deviation from the pretrained product process.
By the standard variational identity for KL-regularised stochastic control \citep{chetrite2015variational}, the optimal controlled path law is an exponential tilt of the reference law. 
Conditionally on the initial noise state,
\begin{equation}
    \dfrac{
        \mathrm d\hat{\mathbb P}^{u^\star}
        \bigl(\,\cdot\,\vert \hat X_0\bigr)
    }{
        \mathrm d\hat{\mathbb P}
        \bigl(\,\cdot\,\vert \hat X_0\bigr)
    }
    \bigl(\hat X_{\cdot}\bigr)
    \propto
    \exp\!\left(
        \lambda r\bigl(\varphi(\hat X_1)\bigr)
    \right).
    \label{eq:optimal_path_tilt}
\end{equation}
Thus, optimal coordination re-weights joint trajectories by the assembly-level reward; see \Cref{apdx:path_space_change_measure} for the change of measure and \Cref{rem:reward_induced_coordination} for the induced dependencies.
%

\section{Experimental Evaluation}
\begin{wrapfigure}[20]{l}{0.52\textwidth}
    \centering
    \includegraphics[
        width=\linewidth,
        trim=0 0 0 0,
        clip
    ]{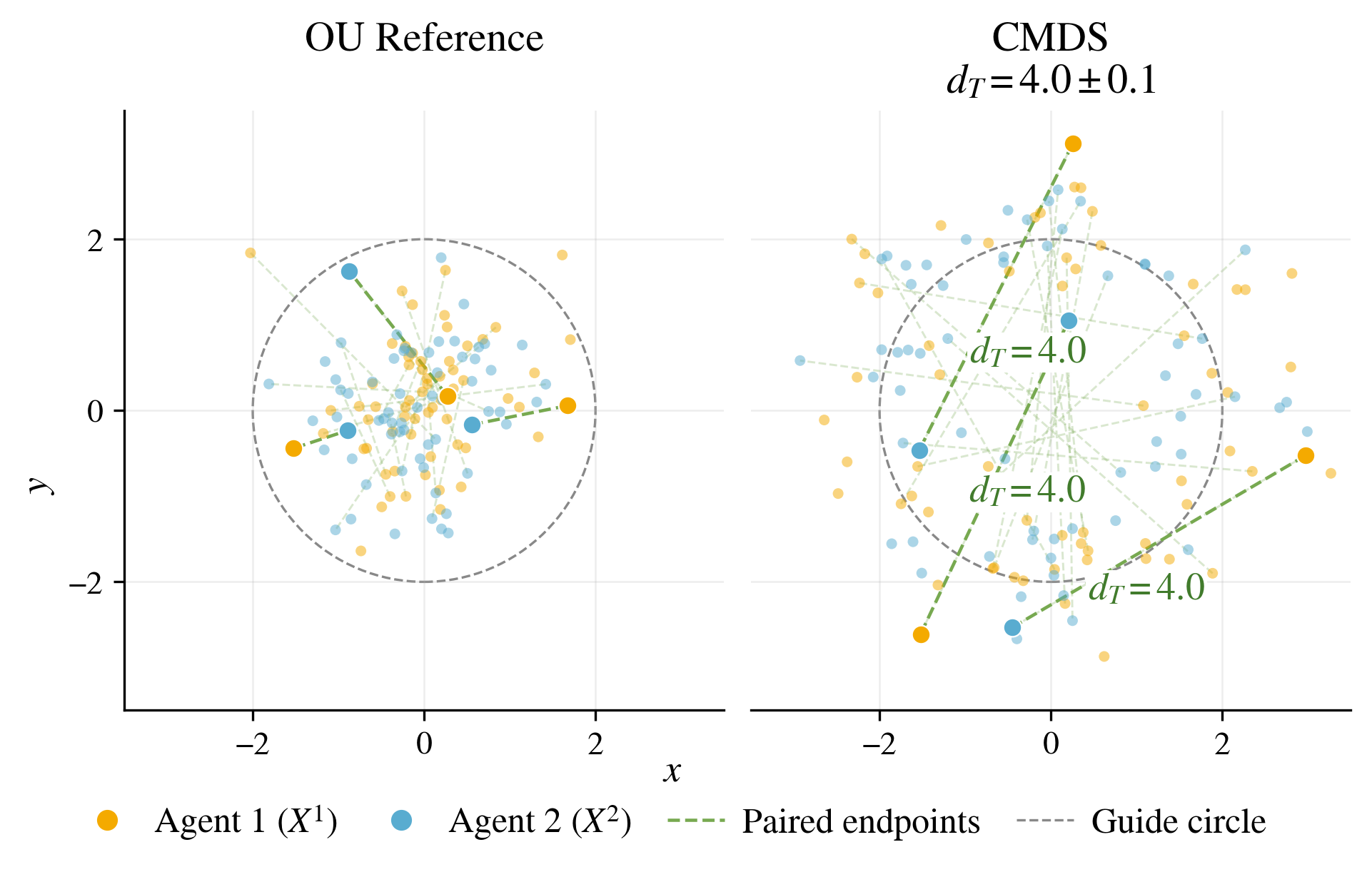}
    \caption{
    \textbf{Two-agent OU problem.}
    Terminal samples from the independent OU reference process and the rotation-equivariant \method{} control for a prescribed separation \(d_{T}=4\).
    Yellow and blue points denote the two agents, and each line joins the terminal states belonging to the same sample. 
    }
    \label{fig:ou-terminal}
\end{wrapfigure}
We evaluate \method{} along three axes:
\ingredient{i} a tractable stochastic-control problem with a known optimal solution, to assess recovery of the target distribution rather than constraint
satisfaction alone;
\ingredient{ii} source separation, to assess posterior inference from blurred mixtures when paired source--mixture training data are unavailable; and
\ingredient{iii} multi-agent motion, to assess scalability to larger models and more agents using off-the-shelf pretrained checkpoints, including
zero-shot transfer to changed geometry and additional agents without retraining the control.
Additionally, we study constraint satisfaction and amortised coordination across relational specifications in \Cref{apdx:square-assembly}.
These constrained visual-assembly experiments provide a controlled test-bed to isolate the effects of component factorisation, reward design,
amortisation, and optimisation.

\subsection{Two-agent OU coordination}
\paragraph{Setup.}
We test target-law recovery in a two-agent SOC problem with a computable value function, i.e., the minimum expected remaining cost, whose gradient determines the optimal control, providing a numerical oracle.
%
%
We consider two independent Ornstein--Uhlenbeck (OU) reference processes \(X_t^1,X_t^2\in\mathbb R^2\), with joint state
\(X_t=(X_t^1,X_t^2)\in\mathbb R^4\).
The agents are initialised and evolve according to
\[
    X_0^1,X_0^2
    \stackrel{\mathrm{iid}}{\sim}
    \mathcal N(0,I_2),
    \qquad
    dX_t^i
    =
    -X_t^i\,dt+dW_t^i,
    \qquad
    t\in[0,1],
    \quad
    i\in\{1,2\},
\]
where \(W^1\) and \(W^2\) are independent two-dimensional Brownian motions.
Consequently, the joint initial law is
\(
    X_0\sim\mathcal N(0,I_4)
\).
The reference components evolve independently, while coordination is induced by the terminal reward
    \(r_{\mathrm{OU}}(x^{1},x^{2})
    =
    -(\lVert x^{2}-x^{1}\rVert_{2}-d_{T})^{2}
\),
with \(d_{T}=4\), and realised through the resulting joint optimal control.
Because the problem reduces to the relative coordinate \(R_t=X_t^2-X_t^1\), we can compute a value-function oracle for the optimal stochastic control (refer to \Cref{apdx:ou_coordination}).
We also include importance sampling (IS) \citep{robert2004monte}, using samples from the (uncontrolled) OU reference process re-weighted according to the terminal reward.
\paragraph{Constraint satisfaction.} 
\begin{wrapfigure}[7]{r}{0.55\textwidth}
    \centering
    \vspace{-2.5em}
    \includegraphics[
        width=\linewidth,
    ]{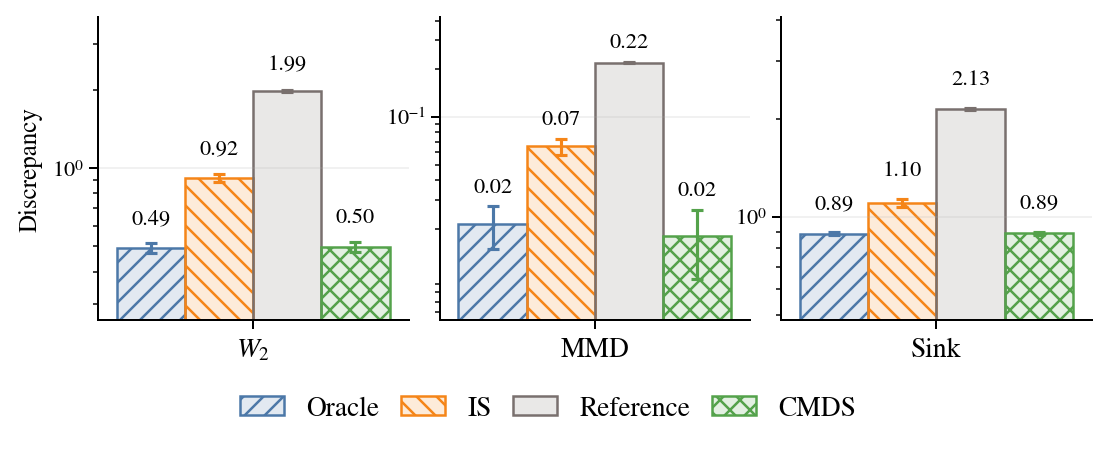}
    \vspace{-2em}
    \caption{
    \textbf{Recovery of the SOC target.}
    }
    \label{fig:ou-metrics}
\end{wrapfigure}
The reference OU process yields a terminal separation of \(1.3 \pm 0.7\), whereas the rotation-equivariant \method{} control achieves \(4.0 \pm 0.1\), see \cref{fig:ou-terminal}.
The equivariant control is used only to respect a known symmetry of the oracle target; it is not a requirement of \method{}.

\paragraph{Distributional recovery.} 
We compare terminal samples against the oracle's joint terminal law, i.e., the \(t=1\) marginal of the optimal path law characterised in \cref{eq:optimal_path_tilt}, using the 2-Wasserstein distance (\(W_2\)), maximum mean discrepancy (MMD), and Sinkhorn (Sink) transport cost over five seeds.
As reported in \cref{fig:ou-metrics}, \method{} achieves discrepancies comparable to oracle self-comparisons
\footnote{Oracle self-comparisons provide a reference level for each reported discrepancy, using two independent oracle sample sets, each matching the evaluation sample size.
}
and outperforms importance sampling (IS).
These results show that \method{} does more than reach the prescribed separation as it also recovers the distributional structure of the reward-tilted target law.
\Cref{apdx:ou_coordination} shows that an unconstrained control can satisfy the scalar distance constraint while failing to recover angular structure of the target distribution.

\subsection{Source separation}
\paragraph{Compositional posterior formulation.} We consider posterior inference in which \(N\) frozen pretrained diffusion samplers independently define component priors through their reverse processes, with reference terminal samples \(\hat X_1^i\sim\tilde{\pi}^i\).
Let \(X=(X^1,\ldots,X^N)\) denote the unknown component state, whose non-injective additive assembly is \(Y=\varphi(X):=\sum_{i=1}^N X^i\), and suppose that we observe a blurred mixture \(b={\rm{A}}_\tau(Y)+\varepsilon\), where \({\rm{A}}_\tau\) is a known Gaussian blurring operator with standard deviation \(\tau\), and \(\varepsilon\sim\mathcal N(0,\sigma_\varepsilon^2 I)\).
Since the additive assembly \(\varphi\) is non-injective, the observation does not uniquely identify its constituent sources; the independently pretrained component priors therefore regularise the resulting compositional posterior over valid decompositions.
Assuming Gaussian observation noise, we define the assembly-level reward \(r_b(y):=-\frac{1}{2\sigma_\varepsilon^2}\|{\rm{A}}_\tau(y)-b\|^2\).
The pretrained diffusion models define the reference product law \(\rho=\tilde{\pi}^1\otimes\cdots\otimes\tilde{\pi}^N\), while \method{} performs compositional posterior inference by coordinating their reverse dynamics, producing controlled terminal components \(\hat X_1^\theta=(\hat X_1^{\theta,1},\ldots,\hat X_1^{\theta,N})\) whose assembly \(\hat Y_1^\theta=\varphi(\hat X_1^\theta)\) jointly explains the observation \(b\), without requiring training samples from the joint source distribution.
For comparison, we adapt diffusion posterior sampling (DPS) \citep{chung2023diffusion} to the product prior, paralleling the posterior-sampling construction of \citet{luan2025projected}.
At each reverse step, we evaluate the observation likelihood on the Tweedie estimates \citep{efron2011tweedie} of both processes and apply the resulting component-wise gradients to their reverse updates; details are given in \Cref{apdx:source-separation}.
\begin{figure}[h]
\vspace{-.5em}
\centering
    \includegraphics[width=0.95\linewidth, 
    trim=0 0 10 0,
    clip
    ]{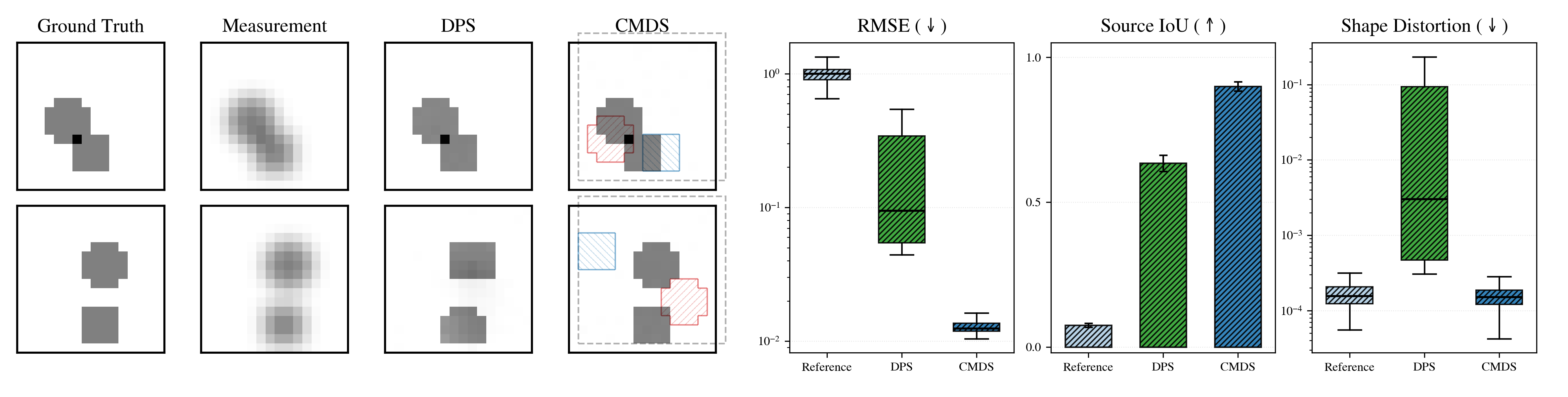}
    \vspace{-.5em}
    \caption{
        \textbf{Source separation from a blurred mixture (\(\tau=1\)).}
        Left: circle--square reconstructions for two held-out mixtures; red and blue hatching behind the \method{} reconstructions shows same-seed matched reference sources.
        Right: comparison of the reference, DPS, and \method{} over \(1{,}024\) held-out mixtures, measuring clean-mixture recovery (RMSE, \(\downarrow\)), labelled-source overlap (IoU, \(\uparrow\)), and shape distortion (\(\downarrow\)).
        Full metric definitions are in \Cref{apdx:source-separation}; aggregate results and representative reconstructions under milder blur (\(\tau=0.5\)) are shown in \cref{fig:source-separation-results,fig:source-separation-samples-0p5}.
    }
    \label{fig:source-separation}
\end{figure}
\paragraph{Recovery under ambiguity.}
As shown in \cref{fig:source-separation}, \method{} achieves the strongest overall recovery under the challenging \(\tau=1\) blur. 
It yields the lowest clean-mixture reconstruction error and the highest source intersection over union (IoU), while maintaining the low shape distortion induced by the frozen component priors.
DPS substantially improves source localisation over independent reference sampling, but markedly distorts the components and yields a larger reconstruction error than \method{}.
This localisation--distortion trade-off reflects the non-identifiability of the problem.
The likelihood depends only on the assembled mixture \(X^1+X^2\), through \({\rm A}_{\tau}(X^1+X^2)\), and therefore contains no direct information
about the differential mode \(X^1-X^2\) that distinguishes the individual sources.
Consequently, the step-wise likelihood corrections used by DPS can improve the reward by redistributing or deforming mass in ways that are poorly aligned with the pretrained component distributions.
By contrast, rather than applying step-wise likelihood guidance, \method{} learns joint feedback over the product of the pretrained reverse processes, while the quadratic path-action penalty favours small coordinated deviations from the frozen dynamics, helping preserve component structure.
Quantitative results under milder blur are provided in \Cref{apdx:source-separation}.

\subsection{From Single-Agent Priors to Coordinated Multi-Agent Motion}\label{sec:trajectory-coordination}
\paragraph{Coordination across motion domains.}
\begin{wrapfigure}[17]{r}{0.5\linewidth}
    \centering
    \vspace{-1.75em}
    \includegraphics[width=\linewidth]{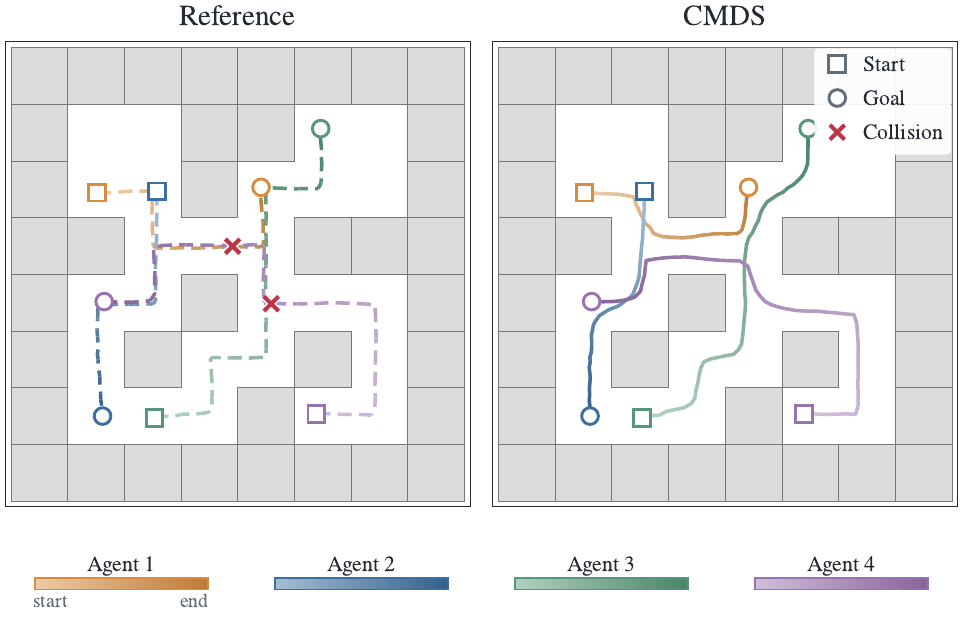}
    \vspace{-1.75em}
    \caption{
    \textbf{PointMaze (Medium) coordination.}
    Dashed (\uncontrolledpath) and solid (\controlledpath) paths use the same initial noise.
    Reference (uncontrolled) process trajectories collide, whereas \method{} steers to avoid collisions while preserving their start--end pairs.
    }
    \label{fig:maze-mosaic}
\end{wrapfigure}
We investigate multi-agent navigation problems where agents move through a shared environment, between prescribed start- and end-points, while satisfying task-specific constraints and avoiding collisions.
We consider planar navigation (PointMaze) \citep{fu2020d4rl}, articulated robot planning (KUKA) \citep{luo2024potential}, and text-conditioned human motion (MDM) \citep{tevet2023human}.
Pretrained single-agent priors generate component trajectories, the assembly operator combines them, and an assembly-level reward coordinates their movement.
Throughout, the component priors remain frozen and only the coordination control is learned.
PointMaze and KUKA condition the component models on prescribed start and end points.
For MDM, component models are text-conditioned and end points are supplied to the coordination control.
Success requires every agent to satisfy the task-specific goal, avoid collisions, and pass task-specific feasibility checks.
Further details are in \Cref{apdx:maze-collision,apdx:kuka-collision,apdx:mdm_multi_planning}.
\begin{table}[tb]
\centering
\small
\setlength{\tabcolsep}{4.5pt}
\renewcommand{\arraystretch}{1.10}
\providecommand{\mazeStd}[1]{\ensuremath{\,{\color{gray}\scriptstyle\pm #1}}}
\resizebox{\linewidth}{!}{%
\begin{tabular}{@{}lcccc@{\hspace{10pt}}cccc@{}}
\toprule
& \multicolumn{4}{c}{Medium} & \multicolumn{4}{c}{Large} \\
\cmidrule(lr){2-5}\cmidrule(lr){6-9}
& \multicolumn{2}{c}{\(N=2\)} & \multicolumn{2}{c}{\(N=4\)}
& \multicolumn{2}{c}{\(N=2\)} & \multicolumn{2}{c}{\(N=4\)} \\
\cmidrule(lr){2-3}\cmidrule(lr){4-5}\cmidrule(lr){6-7}\cmidrule(lr){8-9}
Method & Pool 1 \(\uparrow\) & Pool 2 \(\uparrow\) & Pool 1 \(\uparrow\) & Pool 2 \(\uparrow\)
& Pool 1 \(\uparrow\) & Pool 2 \(\uparrow\) & Pool 1 \(\uparrow\) & Pool 2 \(\uparrow\) \\
\midrule
Reference
& 0.174\mazeStd{0.005} & 0.195\mazeStd{0.008} & 0.068\mazeStd{0.005} & 0.010\mazeStd{0.005} & 0.195\mazeStd{0.016} & 0.219\mazeStd{0.000} & 0.039\mazeStd{0.000} & 0.049\mazeStd{0.005} \\
Best-of-256
& 0.216\mazeStd{0.005} & 0.263\mazeStd{0.016} & 0.076\mazeStd{0.005} & 0.044\mazeStd{0.005} & 0.232\mazeStd{0.009} & 0.328\mazeStd{0.021} & 0.076\mazeStd{0.005} & 0.070\mazeStd{0.000} \\
FK-Steering
& 0.797\mazeStd{0.028} & 0.820\mazeStd{0.034} & 0.398\mazeStd{0.036} & 0.331\mazeStd{0.016} & 0.924\mazeStd{0.030} & 0.953\mazeStd{0.008} & 0.622\mazeStd{0.009} & 0.729\mazeStd{0.023} \\
DPS
& 0.964\mazeStd{0.024} & 0.956\mazeStd{0.016} & 0.958\mazeStd{0.025} & 0.919\mazeStd{0.033} & \textbf{0.984}\mazeStd{0.016} & \textbf{0.977}\mazeStd{0.000} & 0.880\mazeStd{0.030} & 0.927\mazeStd{0.009} \\
\rowcolor{pastelblue!15}
\method{}
& 0.990\mazeStd{0.012} & 0.919\mazeStd{0.016} & 0.701\mazeStd{0.009} & 0.669\mazeStd{0.005} & 0.911\mazeStd{0.016} & 0.831\mazeStd{0.016} & 0.880\mazeStd{0.012} & 0.753\mazeStd{0.027} \\
\rowcolor{pastelblue!15}
\method{} + TRG
& \textbf{0.995}\mazeStd{0.005} & \textbf{0.992}\mazeStd{0.000} & \textbf{0.974}\mazeStd{0.012} & \textbf{0.935}\mazeStd{0.005} & 0.966\mazeStd{0.005} & 0.898\mazeStd{0.016} & \textbf{0.966}\mazeStd{0.012} & \textbf{0.977}\mazeStd{0.000} \\
\bottomrule
\end{tabular}%
}
\caption{\textbf{PointMaze joint success.}
Mean $\pm$ standard deviation (std) across sampling seeds. Training-seed variability is reported separately in \Cref{apdx:maze-training-variance}.
Each pool contains \(128\) instances per layout and agent count.
For \method{} and \method{}+TRG, we train five independent seeds and report the seed with median Pool~1 success for each setting, breaking ties using Pool~2.
}
\label{tab:maze-results}
\vspace{-1em}
\end{table}
\paragraph{PointMaze: amortised multi-agent navigation.}
We pretrain one conditional single-agent trajectory prior for each of the \textit{``Medium''} and \textit{``Large''} layouts \citep{fu2020d4rl}, then coordinate two or four copies.
The reward penalises insufficient wall clearance, inter-agent collisions, path length, and trajectory acceleration.
We evaluate on two sets of unseen start--end instances.
Pool~1 places starts and goals in navigable grid squares used during control training, whereas Pool~2 uses squares excluded from control training (\Cref{apdx:maze-data-splits}).
\Cref{tab:maze-results,fig:maze-mosaic} compare independent reference sampling, Best-of-256, FK-Steering \citep{pmlr-v267-singhal25b}, DPS \citep{chung2022diffusion}, and our learned control.
We evaluate two parameterisations of the coordination control: \method{} uses an attention-based control conditioned on the joint multi-agent state, while \method{}+TRG additionally incorporates the \emph{Tweedie reward gradient} (TRG), namely the reward gradient evaluated at the Tweedie estimate, inspired by prior work \citep{denker2024deft,venkatraman2024amortizing}.
\method{}+TRG improves or matches the reported base-\method{} rates, achieving Pool~2 success between \(89.8\%\) and \(99.2\%\) across the reported settings; DPS is the best performing on \textit{``Large''} with two agents.
Across five independent training seeds per setting, \method{}+TRG achieves mean Pool~2 success between \(90.6\%\) and \(98.3\%\), with training-seed standard deviations no larger than \(1.5\) percentage points (\Cref{apdx:maze-training-variance}).
To stabilise adjoint-matching training against large residuals, we use a pseudo-Huber regression loss in place of the squared loss; details are in \Cref{apdx:maze-robust-loss}.
\begin{figure}
    \centering
    \includegraphics[width=0.95\linewidth]{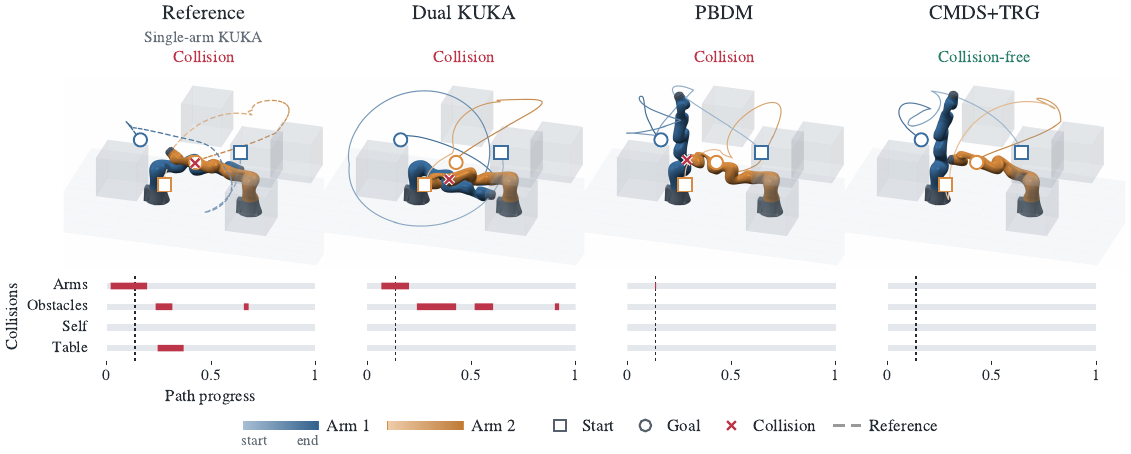}
    \caption{
        \textbf{A dual-arm coordination example.}
        Poses share normalised path progress \(s=0.135\); crosses mark arm--arm contact, and red intervals mark strict mesh collisions.
        PBDM uses \(n_{\mathrm{cand}}=6\) with up to five repair attempts per candidate, versus \(n_{\mathrm{cand}}=1\) for the other methods; its modified, native-accepted path fails strict checking between knots.
        This outcome-selected illustration is not a matched-budget comparison.
        Additional poses and clearance traces are provided in \Cref{apdx:kuka-collision}.
    }
    \label{fig:placeholder}
    \vspace{-1.5em}
\end{figure}
\paragraph{KUKA: multi-arm planning from single-arm priors.}
We coordinate two seven-joint robot arms in a shared environment containing obstacles.
Each task specifies the obstacle layout and the start and goal configurations of both arms.
A shared \method{}+TRG control coordinates frozen pretrained single-arm diffusion models to generate their paths jointly.
\begin{wraptable}[14]{r}{0.55\textwidth}
\vspace{-.75em}
\centering\small
\setlength{\tabcolsep}{4pt}
\renewcommand{\arraystretch}{1.1}
\providecommand{\std}[1]{\ensuremath{\,{\color{gray}\scriptstyle\pm #1}}}
\resizebox{\linewidth}{!}{%
    \begin{tabular}{@{}lcc@{}}
    \toprule
    Method & Strict success (\%, $\uparrow$) & Planning time (s, $\downarrow$) \\
    \midrule
    PBDM & 60.9\std{8.0} & 0.427\std{0.003} \\
    Dual KUKA + guidance & 74.3\std{5.7} & 0.356\std{0.002} \\
    Dual KUKA prior & 36.3\std{8.7} & 0.274\std{0.033} \\
    Single-arm priors + guidance & 51.0\std{0.6} & 0.355\std{0.001} \\
    Reference single-arm priors & 11.4\std{1.5} & \textbf{0.193}\std{0.002} \\
    \midrule
    \rowcolor{pastelblue!25}
    \method{}+TRG & \textbf{77.3}\std{0.7} & 0.368\std{0.002} \\
    \bottomrule
    \end{tabular}%
}
\caption{\textbf{Dual KUKA planning with one initial joint candidate.}
Mean $\pm$ sample std. strict successes on 413 held-out test tasks.
Planning time is the median value and includes scene setup, generation, verification and optional refinement.
}
\label{tab:kuka-success-time}\end{wraptable}
Publicly released checkpoints \citep{luo2024potential} predict noise as an energy gradient, \(\epsilon_\psi=\nabla_x E_\psi\).
Adjoint propagation therefore requires Hessian--vector products through the energy, while the checkpoint parameters remain frozen.
Strict success requires both arms to reach their goals, remain within the allowed angular range of every robot joint, and pass checks for inter-arm collisions, self-collisions, and collisions with the obstacles and worktable.
We evaluate on the test layouts released by \citet{luo2024potential}, unseen by both the priors and the controller (\Cref{apdx:kuka-data}), and denote by \(n_{\mathrm{cand}}\) the number of initial candidate multi-arm plans generated per task.
With one candidate plan, \method{}+TRG achieves \(77.3\%\) strict success, compared with \(11.4\%\) for the reference independent single-arm sampling and \(51.0\%\) when adding collision guidance (see \cref{tab:kuka-success-time}).
Using the separately pretrained two-arm diffusion prior, guided Dual KUKA achieves \(74.3\%\), while Potential Based Diffusion Motion Planning (PBDM; \citealp{luo2024potential}) achieves \(60.9\%\) using the same prior within its native candidate-selection and refinement procedure.
We also apply the learned control without retraining to shifted or enlarged obstacles, wider arm-base spacing, and additional arms (\cref{tab:kuka-generalization}).
On a separate set of tasks with five arms and four candidates per task (\(n_{\mathrm{cand}}=4\)), \method{} + TRG achieves \(68.3\%\) strict success, compared with \(48.3\%\) for guided single-arm priors.
Additional information is provided in \Cref{apdx:kuka-collision}.
%
%
%
\begin{table*}[t]
\centering
%
\begin{minipage}[t]{0.48\textwidth}
\vspace{0pt}
\centering

\begingroup
\fontsize{9}{10.5}\selectfont
\setlength{\tabcolsep}{1.5pt}
\renewcommand{\arraystretch}{1.0}
\providecommand{\std}[1]{}
\renewcommand{\std}[1]{\ensuremath{\,{\scriptstyle\pm #1}}}
\arrayrulecolor{black}%

\resizebox{\linewidth}{!}{%
\begin{tabular}{@{}lc*{3}{c}@{\hspace{5pt}}*{3}{c}@{}}
\toprule
\multirow{2}{*}{Method}
 & \multirow{2}{*}{\(n_{\mathrm{cand}}\)}
 & \multicolumn{3}{c}{Geometry transfer}
 & \multicolumn{3}{c}{Number of arms} \\
\cmidrule(lr){3-5}\cmidrule(lr){6-8}
 & & \shortstack{Shifted\\obstacles}
 & \shortstack{Larger\\obstacles}
 & \shortstack{Wider\\bases}
 & 3 & 4 & 5 \\
\midrule

\multirow{2}{*}{\shortstack[l]{Single-arm KUKA\\+ guidance}}
 & 1 & 53.4\std{4.0} & 14.3\std{1.3} & 49.2\std{2.4}
 & 33.8\std{0.7} & 28.4\std{1.7} & 19.3\std{2.0} \\
 & 4 & 83.2\std{2.8} & 42.1\std{4.1} & 84.0\std{0.7}
 & 69.1\std{1.5} & 55.7\std{3.5} & 48.3\std{2.4} \\

\addlinespace[1pt]

\multirow{2}{*}{\shortstack[l]{Dual KUKA\\+ guidance}}
 & 1 & 68.9\std{3.9} & 31.1\std{10.0} & 58.2\std{8.1}
 & -- & -- & -- \\
 & 4 & 89.7\std{0.8} & 66.9\std{3.8} & 87.8\std{2.8}
 & -- & -- & -- \\

\midrule

\rowcolor{pastelblue!25}
 & 1 & \textbf{74.0}\std{1.7} & \textbf{42.7}\std{3.7}
 & \textbf{72.1}\std{2.2} & \textbf{59.5}\std{0.3}
 & \textbf{52.3}\std{3.4} & \textbf{41.9}\std{1.7} \\

\rowcolor{pastelblue!25}
\multirow{-2}{*}{\method{} + TRG}
 & 4 & \textbf{91.6}\std{0.5} & \textbf{73.0}\std{1.0}
 & \textbf{91.8}\std{0.8} & \textbf{86.8}\std{1.0}
 & \textbf{76.2}\std{2.6} & \textbf{68.3}\std{1.2} \\

\bottomrule
\end{tabular}%
}
\endgroup

\captionof{table}{
\textbf{Zero-shot KUKA generalisation.}
Strict success (\%), mean \(\pm\) sample std. across three sampling seeds, for \(n_{\mathrm{cand}}=1,4\).
The left block reports geometry transfer; the right block reports transfer to larger numbers of arms.
}
\label{tab:kuka-generalization}
\label{tab:kuka-arm-count}

\end{minipage}%
%
\hspace{5pt}%
\begin{minipage}[t]{0.48\textwidth}
\vspace{5pt} 
\centering

\scriptsize
\setlength{\tabcolsep}{2.8pt}
\renewcommand{\arraystretch}{1.08}

\resizebox{\linewidth}{!}{%
\begin{tabular}{@{}lrrrr@{}}
\toprule
Method
& \shortstack{Success\\(\%) \(\uparrow\)}
& \shortstack{Collision\\free (\%) \(\uparrow\)}
& \shortstack{Goal err.\\(m) \(\downarrow\)}
& \shortstack{Time\\(s) \(\downarrow\)} \\
\midrule

Best-of-256
& \(0.0\,{\color{gray}\scriptstyle\pm0.0}\)
& \(16.7\,{\color{gray}\scriptstyle\pm5.9}\)
& \(2.456\,{\color{gray}\scriptstyle\pm0.146}\)
& \(69.90\,{\color{gray}\scriptstyle\pm0.29}\) \\

DPS
& \(0.0\,{\color{gray}\scriptstyle\pm0.0}\)
& \(61.7\,{\color{gray}\scriptstyle\pm10.8}\)
& \(3.751\,{\color{gray}\scriptstyle\pm0.284}\)
& \(8.57\,{\color{gray}\scriptstyle\pm0.02}\) \\

PCD
& \(4.2\,{\color{gray}\scriptstyle\pm2.9}\)
& \(8.3\,{\color{gray}\scriptstyle\pm2.9}\)
& \(0.500\,{\color{gray}\scriptstyle\pm0.000}\)
& \(9.99\,{\color{gray}\scriptstyle\pm0.06}\) \\

PCD++
& \(75.0\,{\color{gray}\scriptstyle\pm6.6}\)
& \(\mathbf{98.3}\,{\color{gray}\scriptstyle\pm2.3}\)
& \(0.487\,{\color{gray}\scriptstyle\pm0.001}\)
& \(145.34\,{\color{gray}\scriptstyle\pm4.46}\) \\

\midrule

\rowcolor{pastelblue!45}
\method{}
& \(\mathbf{90.0}\,{\color{gray}\scriptstyle\pm6.3}\)
& \(90.0\,{\color{gray}\scriptstyle\pm6.3}\)
& \(\mathbf{0.158}\,{\color{gray}\scriptstyle\pm0.008}\)
& \(\mathbf{0.54}\,{\color{gray}\scriptstyle\pm0.00}\) \\

\bottomrule
\end{tabular}%
}

\captionof{table}{
    \textbf{Four-person motion coordination.}
    Mean \(\pm\) sample std. across five sampling seeds on \(24\) scenes; \method{} uses one trained control.
    Failures count as zero successes.
    Full metrics appear in \cref{tab:n4_coordination}.
}
\label{tab:n4_coordination_main}
\vspace{-3em}
\end{minipage}
\end{table*}
\begin{figure}
    \centering
    \includegraphics[width=\linewidth]{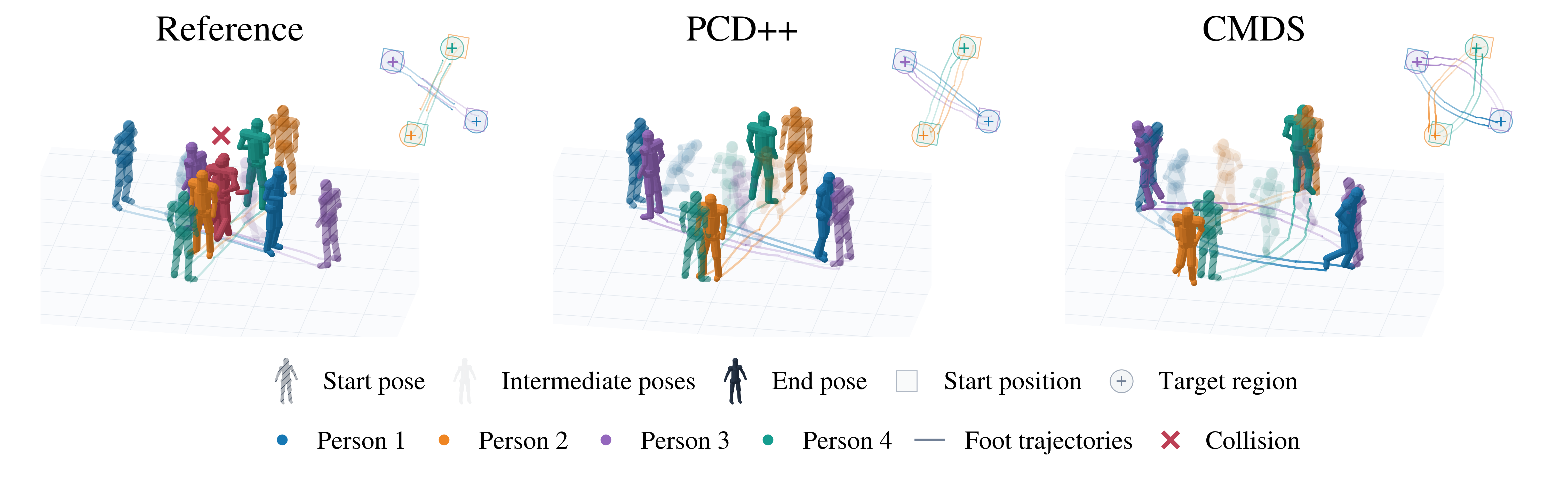}
    \caption{
    \textbf{Four-person motion coordination.}
        We compare reference MDM, PCD++, and CMDS on the four-person coordination task.
        Solid curves show left- and right-foot trajectories, with intermediate poses faded; the top-down insets summarise the trajectories and target completion.
        Here, the reference produces collisions and misses targets, whereas PCD++ and CMDS reach all targets without collisions.
        \Cref{fig:mdm_appedix} in \Cref{apdx:mdm_multi_planning} compares all methods.
    }
\label{fig:mdm_main}
\vspace{-1.5em}
\end{figure}
\vspace{-0.5em}
\paragraph{MDM: from single-person motion to coordinated crowds.}
We use the 50-step MDM checkpoint \citep{tevet2023human} to coordinate four independently text-conditioned people over six-second motions (see \Cref{apdx:mdm_multi_planning}).
A generated scene is successful when all four people move from their starting positions to their targets, shown in the top-down insets of \cref{fig:mdm_main}, and their motions pass both the inter-person collision and kinematic-validity checks.
Baselines include Best-of-\(256\), DPS, PCD \citep{luan2025projected}, and our PCD++ extension, which jointly optimises noisy motions through the frozen denoiser for goal attainment and clearance (see \cref{alg:motion-pcdpp}).
\Cref{tab:n4_coordination_main,fig:mdm_main} report \(90\%\) joint success for \method{} versus \(75\%\) for PCD++.
PCD++ has a higher collision-free rate, but lower joint success and longer sampling time.

\section{Conclusion}
We introduced \method{}, a modular post-training framework for coordinating frozen pretrained diffusion models through stochastic optimal control.
An assembly-level reward specifies joint requirements, while an amortised control couples the component reverse processes under a quadratic path-action penalty.
Experiments demonstrate target-law recovery in a tractable setting, amortised relational assembly, and source recovery, while released KUKA and MDM priors extend coordination to multi-arm planning and multi-person motion.
Transfer to additional robot arms further illustrates the reuse enabled by learning coordination separately from component generation.
Together, these results show that frozen component models can support structured tasks beyond their individual training settings through learned interactions.
Future work includes improving reward design, characterising how robust matching and control parameterisation affect target-law approximation, and extending coordination to larger, more heterogeneous generative models.

%% file: _related_work_apdx.tex
\section{Related Work}\label{app:related-work}
\paragraph{Modular and compositional generative modelling.}
Modular generative modelling has a long history in pattern theory and structured scene representations \citep{grenander1978pattern,bienenstock1996compositionality,williams2025structured}, and recent work advocates composing specialised generators rather than learning every combination within a single model \citep{pmlr-v235-du24d}.
For diffusion models, this perspective has motivated compositional visual generation through combinations of conditional scores
\citep{liu2022compositional} and sampling methods for composed energy-based models \citep{du2023reduce}.
Related approaches compose learned constraint factors \citep{yang2023compositional} or diffusion policies across skills, domains, and input modalities \citep{wang2024poco,cao2025modality}.
\method{} follows this modular perspective while explicitly distinguishing component states from their assembled output.
Each frozen generator supplies a component, and an assembly-level reward specifies the interactions introduced through learned joint control.

\paragraph{Joint generation with pretrained diffusion models.}
A related line of work coordinates multiple pretrained diffusion processes directly during generation.
MultiDiffusion binds multiple denoising paths through shared image-generation constraints \citep{bartal2023multidiffusion}, while SyncDiffusion synchronises image patches using perceptual-similarity gradients \citep{lee2023syncdiffusion}.
MMDisCo coordinates frozen single-modal diffusion models using a discriminator trained on paired audio--video examples 
\citep{hayakawa2025mmdisco}.
In contrast to paired-data supervision, \method{} learns coordination from an assembly-level reward under a path-space control objective, without requiring training examples of valid joint outputs.

\paragraph{Cost-based coordination of diffusion processes.}
Coordinated Diffusion (CoDi; \citealp{peters2026coordinated}) and Projected Coupled Diffusion (PCD; \citealp{luan2025projected}) are particularly close to our setting.
CoDi constructs a KL-regularised cooperative objective relative to single-agent base policies and guides their diffusion processes simultaneously.
Its sampling-based guidance estimator uses online cost evaluations without requiring cost gradients.
PCD likewise introduces interactions between pretrained diffusion processes at test time, combining cost-based guidance with projections onto prescribed constraint sets.
\method{} shares the use of separately pretrained generators and a joint criterion, but learns a reusable control across task instances under a quadratic path-action penalty.
Coordination is therefore amortised into a joint-state control, rather than reconstructed solely through inference-time guidance.
The approaches also differ in their constraint treatment: PCD explicitly projects onto supplied constraint sets, whereas the finite reward penalties used in our experiments do not guarantee hard constraint satisfaction.
Our current adjoint-based implementation additionally assumes differentiable rewards, unlike CoDi's cost estimator.
Since CoDi and PCD act at inference time, they are complementary to amortised coordination: their guidance or projection mechanisms could in principle be applied on top of a learned \method{} control when additional instance-specific constraints must be imposed at sampling time.

\paragraph{Diffusion-based multi-agent planning.}
Diffuser formulates planning as guided generation of trajectories
\citep{janner2022planning}.
Multi-agent extensions such as MADiff learn coordinated behaviour from offline multi-agent trajectories
\citep{zhu2024madiff}.
Closer to our data assumptions, Multi-robot Multi-model planning Diffusion (MMD) combines single-agent diffusion priors with constraint-based search and replanning
\citep{shaoul2025multi}.
Our motion experiments similarly begin from frozen component priors, but learn coordination separately from component generation.
In PointMaze, we coordinate copies of a single-agent trajectory prior across start--goal instances; in KUKA, we coordinate copies of a released single-arm energy-based prior for multi-arm planning; and in the MDM experiment, we coordinate copies of a released text-conditioned single-person motion diffusion prior to generate collision-aware multi-person motion.
Across these settings, the component models are not trained on the corresponding multi-agent task, while coordination is amortised into a reusable joint control rather than resolved solely through instance-specific search or guidance.

\paragraph{Inference-time steering and compositional posterior inference.}
Diffusion posterior sampling (DPS) conditions pretrained diffusion models on observations through approximate likelihood guidance evaluated using denoised estimates
\citep{chung2023diffusion}.
FK-Steering instead uses reward-dependent particle reweighting and resampling during generation
\citep{pmlr-v267-singhal25b};
its particles represent alternative complete candidates, rather than the components of one assembled output.
Our source-separation experiment places the observation model after an additive assembly and performs inference over the constituent sources.
We compare against DPS adapted to the same product prior, while learning observation-dependent joint feedback under a path-action penalty.
Amortisation is compatible with additional test-time information:
our Tweedie reward-gradient parameterisation retains explicit reward-gradient features within the learned control.
The test-time coordination mechanisms of PCD \citep{luan2025projected} and CoDi \citep{peters2026coordinated}
are closely related to the \emph{cooperative DPS} (CDPS) baseline introduced in earlier work \citep{barbano2026cmad}.
CDPS replaces a learned cooperative control with per-step guidance obtained from a joint cost evaluated on the agents' Tweedie estimates.
PCD explicitly derives an analogous posterior-sampling variant, in which a coupling cost evaluated on Tweedie-denoised component estimates is differentiated through each diffusion process.
CoDi follows the same general principle of augmenting the product of single-agent diffusion scores with an online joint cost-guidance term, but estimates this correction gradient-free.
These methods therefore provide non-amortised, inference-time alternatives to \method{}, whereas \method{} learns a reusable joint feedback control across task instances.

\paragraph{Stochastic control and amortised diffusion steering.}
Optimal-control perspectives connect diffusion generation and reward-based fine-tuning to controlled dynamics and relative-entropy regularisation \citep{berner2024optimal,uehara2024fine}.
Stochastic optimal control matching and adjoint matching develop regression-based approaches for learning such controls
\citep{domingo2024stochastic,domingoenrich2025adjoint}.
DEFT learns a generalised $h$-transform correction while keeping the pretrained diffusion model frozen
\citep{denker2024deft}, and relative trajectory balance amortises posterior sampling through a path-space balance objective \citep{venkatraman2024amortizing}.
Building on these foundations, \method{} coordinates a reference product process through an assembly-level reward and joint-state feedback, with the component generators remaining frozen.
Adjoint matching is one learning strategy within this framework; our discrete implementation uses the full controlled-transition adjoint, rather than the lean-adjoint variant of
\citet{domingoenrich2025adjoint}.

%% file: _theory_apdx.tex
\section{Diffusion and Path-Space Formulation of \method{}}
\begin{takehome}
\textbf{Key Takeaways:}
\noindent
\textbf{The pretrained models define an independent reference process.} The $N$ component diffusion models form a product reverse process, and because the reference density factorises across agents, its reverse drift acts agent-wise. Cross-agent dependence is introduced only by the coordination control (\cref{apdx:more_theory_DM}).
\textbf{The assembled process has well-defined marginal dynamics, but itself need not be Markov.}
\method{} controls the full component state $\hat X_s$, while the assembled state $\hat Y_s = \varphi(\hat X_s)$ is obtained through the assembly map. 
Its fixed-time law is the pushforward of the joint-component-state law, and its instantaneous marginals satisfy an exact projected Fokker-Planck equation, but $\hat Y$ need not be Markovian when $\varphi$ discards information about dynamics (\cref{apdx:controlled_time_reverse_SDE}).
\textbf{Coordination is a minimum-KL change of controls in path measure.}
The quadratic control action equals the relative entropy from the independent pretrained product process \cref{eq:path_kl_common_initial}, so \method{} trades assembly reward against the smallest path-space deviation from the component priors (\cref{apdx:path_space_change_measure}).
\textbf{The optimal law is reward tilting.}
The SOC optimum satisfies 
$$
\frac{\dd \mathbb{\hat P}^\star}{\dd \mathbb{\hat P}} (\hat{X}_\cdot) = \frac{\exp(\lambda r(\varphi(\hat{X}_1)))}{Z(\hat{X}_0)},
$$
where the $\hat X_0$-dependent normaliser preserves the prescribed initial marginal. 
In the complete-noising limit, $\hat{X}_1$ is independent of $\hat{X}_0$, and this reduces to the terminal reward tilt in \cref{eq: prop terminal marginal}.
The reward-tilting formulation holds for general assembly-level rewards.
In the particularly relevant and interesting case in which $r \circ \varphi$ is non-separable across components, the exponential tilt generally no longer factorises, so the resulting target law introduces intended dependencies between components that have no intended dependencies under the reference product law (\cref{apdx:proff}).
\textbf{The change of measure can be transferred to the assembled process.} If $\varphi$ is injective, the assembled path-density ratio can be written directly through $\varphi^{-1}$. 
If $\varphi$ is non-injective, the assembled density ratio is instead obtained by conditionally averaging the component-space likelihood ratio over all trajectories mapping to the same assembly (\cref{apdx:assembly_change_measure}).
\end{takehome}

We use the notation summarised in \cref{tab:notation-convention}.
Particularly, $s\in[0,1]$ denotes reverse, generative time; hatted variables denote reverse-time quantities; $i\in\{1,\ldots,N\}$ is an agent index; and $\hat X_s=(\hat X_s^1,\ldots,\hat X_s^N)\in\mathbb R^D$, with $D=Nd$. 
The fixed assembly map $\varphi:\mathbb R^D\to\mathbb R^m$ differs from experiment to experiment, and $\hat Y_s=\varphi(\hat X_s)$. 
The system-level reward is $r$ and $\lambda>0$ is its strength.

We use \textit{}{reference} to denote independent sampling from the component reference processes, without additional assembly-level steering.
In discrete-time algorithms and implementation descriptions, \(X_\ell\) denotes the reverse-rollout state with its hat and
control superscript suppressed; \(\hat X_{1\mid\ell}\) denotes the clean-state estimate (in discrete time).

\begin{table}[h]
    \centering
    \small
\begin{tabular}{ll}
    \toprule
    \textbf{Quantity} & \textbf{Notation} \\
    \midrule

    \multicolumn{2}{l}{\textbf{Variables and processes}} \\
    \midrule

    Abstract component \& assembly
    & \(X^i,\; X,\; Y=\varphi(X)\) \\

    (Uncontrolled) reference reverse process
    & \(\hat X_s^i,\; \hat X_s,\; \hat Y_s\) \\

    Generic controlled reverse process
    & \(\hat X_s^{u,i},\; \hat X_s^u,\; \hat Y_s^u\) \\

    Parametrised controlled reverse process
    & \(\hat X_s^{\theta,i},\; \hat X_s^\theta,\; \hat Y_s^\theta\) \\

    Stop-gradient controlled rollout
    & \(\hat X_s^{\bar\theta,i},\; \hat X_s^{\bar\theta},\; \hat Y_s^{\bar\theta}\) \\

    Forward-time OU experiment
    & \(X_t^i,\; X_t^{\theta,i}\) \\

    \midrule
    \multicolumn{2}{l}{\textbf{Densities and distributions}} \\
    \midrule

    Initial reverse-time noise density
    & \(\hat p_0\) \\

    Initial forward-time data density
    & \(p_0\) \\

    Terminal density of the reference reverse process
    & \(\hat p_1\) \\

    Terminal forward-time noise density
    & \(p_1\) \\

    Exact target law of a generic component
    & \(\pi\) \\

    Learned terminal law of a generic component
    & \(\tilde{\pi}\) \\

    Joint terminal law of the independent pretrained samplers
    & \(\rho = \tilde{\pi}^1 \otimes \cdots \otimes \tilde{\pi}^N\) \\

    \bottomrule
\end{tabular}
    \vspace{.5em}
    \caption{\textbf{Notation convention.}
    Hatted variables denote reverse-time quantities.
    The superscript \(u\) denotes a generic control, while \(\theta\) denotes the process induced by the parametrised control \(U_\theta\).
    For the continuous-time process notation, quantities without a control superscript refer to the uncontrolled reference process, unless stated otherwise. 
    In the discrete-time algorithms and implementation descriptions, $X_\ell$ denotes the rollout state, with the control superscript suppressed when the generating method is clear from the context.    
    }
    \label{tab:notation-convention}
\end{table}

\paragraph{Standing assumptions.}
Throughout the Appendix, we work under the following convenient sufficient assumptions.
For each \(i\in\{1,\ldots,N\}\), the drift
\(
    (t,x^i)\mapsto f_t^i(x^i)
\)
is jointly measurable, locally Lipschitz in \(x^i\), uniformly on compact time intervals, and has at most linear growth.
The diffusion scale
\(
    \sigma:[0,1]\to[0,\infty)
\)
is continuous and satisfies
\(
    \sigma_t>0
\)
for \(t\in(0,1)\).
The joint forward process admits strictly positive one-time densities \(p_t\), and the map \(p:(0,1)\times\mathbb R^D\to(0,\infty)
\), defined by \(p(t,x):=p_t(x)\), belongs to \(C^{1,2}((0,1)\times\mathbb R^D)\) and satisfies the integrability conditions required for the time-reversal identities below.
\subsection{Forward and Reverse Diffusion Processes}\label{apdx:more_theory_DM}


For clarity, we present the derivation for \(N\) component processes, each taking values in \(\mathbb R^d\) and driven by a common scalar
diffusion schedule, so that the full component state is their concatenation, 
\begin{align}
    X_t=(X_t^1,\ldots,X_t^N)\in\mathbb R^D,
    \qquad
    D=N d.
    \end{align}
Under the standing assumptions stated above, we define $\{X_t\}_{t\in[0,1]}$ by the time-forward SDE in a general form
\begin{align}
\dd X_t
=
f_t(X_t)\,\dd t + \sigma_t\,\dd \W_t,
\qquad
\W_t\text{ is }D\text{-dimensional},\quad X_t\in\mathbb R^D .
\end{align}

$X_t$ is the expanded process corresponding to the concatenation of the $N$ agents, with $X_t^i\in\Rd$.
Here \(\W_t=((\W_t^1)^\top,\ldots,(\W_t^N)^\top)^\top,\) where $\W^1,\ldots,\W^N$ are independent $d$-dimensional Wiener processes, and the initial component states are sampled independently.
We have $D=N\times d$ and
\begin{eqbox}
\begin{align}
    X_t=
    \begin{bmatrix}
    X^1_t\\
    \vdots\\
    X^N_t
    \end{bmatrix},
    \qquad
    f_t(X_t)=
    \begin{bmatrix}
    f^1_t(X^{1}_t)\\
    \vdots\\
    f_t^N(X^{N}_t)
\end{bmatrix}.
\end{align}
\end{eqbox}
Consequently, the reference process factorises across agents, and its one-time density satisfies \( p_t(x)=\prod_{i=1}^N p_t^i(x^i).\)
%
By the Fokker--Planck equation, the corresponding forward density $\pt$ satisfies
\begin{align}
    \partial_t \pt(x)
    &=
    -\,\grad_x\!\cdot\!\bigl(f_t(x)\,\pt(x)\bigr)
    +\frac12\,\sigma_t^{2}\,\Delta_x \pt(x).
\end{align}

Under the standard regularity assumptions ensuring time reversal \citep{nelson1967dynamical, anderson1982reverse, haussmann1986time}, define
\begin{align}
    \hat X_s := X_{1-s},
    \qquad
    \hat{p}_s := p_{1-s},
    \qquad
    \hat \sigma_s := \sigma_{1-s}
    \qquad s\in\TT .
    \end{align}
    Then $\hat X_s$ solves the reverse-time SDE
    \begin{align}\label{eq:uncontrolled-reverse-full-state}
    \dd \hat X_s
    &=
    \hat f_s(\hat X_s)\,\dd s
    + \hat\sigma_{s}\,\dd\bar{\W}_s,
    \qquad
    \hat f_s(x):=
    -\,f_{1-s}(x)+\hat\sigma_{s}^2\,\grad_x\log p_{1-s}(x),
\end{align}
where \(\bar{\W}_s\) is a standard \(D\)-dimensional Brownian motion with respect to the reverse-time filtration.
Since the reference density factorises across agents, the reverse drift also acts component-wise,
\[
\hat f_s(x)
=
\begin{bmatrix}
    \hat f_s^1(x^1)\\
    \vdots\\
    \hat f_s^N(x^N)
    \end{bmatrix},
    \qquad
    \hat f_s^i(x^i)
    =
    -f_{1-s}^i(x^i)
    +\hat\sigma_s^2
    \nabla_{x^i}\log p_{1-s}^i(x^i).
\]
Its density satisfies
\begin{align}
    \partial_s \hat{p}_s(x)
    &=
    -\,\grad_x\!\cdot\!\bigl(\hat f_s(x)\,\hat{p}_s(x)\bigr)
    +\frac12\,\hat\sigma_{s}^2\,\Delta_x \hat{p}_s(x).
\end{align}


Let \(\varphi:\mathbb R^D\to\mathbb R^m\) be the fixed assembly map.
We assume that
\(
    \varphi\in C^2(\mathbb R^D;\mathbb R^m)
\)
and that its first and second derivatives are bounded.
This condition includes the linear masking, stacking, and additive
assembly maps considered in our experiments.
We define
\begin{eqbox}
\begin{equation}
    \hat Y_s
    =
    \varphi(\hat X_s)
    \in\mathbb R^m.
\end{equation}
\end{eqbox}
Its marginal law is the push-forward measure
\begin{equation}
    \hat\mu_s^Y
    :=
    \varphi_{\#}
    \bigl(
        \hat p_s(x)\,\mathrm dx
    \bigr).
\end{equation}
Whenever \( \hat\mu_s^Y\) is absolutely continuous with respect to Lebesgue measure, we denote its density by \( \hat p_s^Y\).

\subsection{Controlled Reverse Diffusions and the Assembled Process}\label{apdx:controlled_time_reverse_SDE}

We control the joint component process, and hence the assembled process, through a normalised drift perturbation.
Let
\[
    u:[0,1]\times\mathbb R^D\to\mathbb R^D,
    \qquad
    (s,x)\mapsto u_s(x),
\]
be an admissible Markov control such that the controlled SDE below admits a weak solution and satisfies the integrability conditions.
We take the controlled and reference processes to share the same initial
law,
\(
    \hat X_0^u\sim\nu_0
\).
Then the controlled reverse diffusion is
\begin{align}
    \dd\hat X_s^u
    =
    \Bigl[
    \hat f_s(\hat X_s^u)+\hat \sigma_{s}u_s\bigl(\hat X_s^u\bigr)
    \Bigr]\,\dd s
    +
    \hat \sigma_{s}\,\dd\bar{\W}_s, 
    \label{eq:controlled-reverse-full-state}
\end{align}
where \(\bar{\W}\) is a standard \(D\)-dimensional Brownian motion under the controlled law \(\hat{\mathbb P}^u\).
When the law of $\hat X_s^u$ admits a density, denoted $\hat p_s^u$, this density satisfies
\begin{eqbox}
\begin{align}
    \partial_s \hat p_s^u(x)
    =
    -\,\grad_x\!\cdot\!\Bigl(
    \bigl[\hat f_s(x)+\hat\sigma_{s}u_s(x)\bigr]\hat p_s^u(x)
    \Bigr)
    +\frac12\,\hat \sigma_{s}^2\,\Delta_x \hat p_s^u(x).
\end{align}
\end{eqbox}

In practice, the drift perturbation may be parameterised using assembled
features, for example as
\(
    U_\theta(x,s)
    =
    \tilde U_\theta(x,\varphi(x),s),
\)
while remaining a full-state feedback map from \(\mathbb R^D\) to \(\mathbb R^D\).

Applying It\^o's formula to $\hat Y_s^u:=\varphi(\hat X_s^u)$ gives
\begin{align}
    \dd \hat Y_s^u
    &=
    \Bigl[
    D_x\varphi(\hat X_s^u)\,\hat f_s(\hat X_s^u)
    +\frac12\,\hat\sigma_{s}^2\,\Delta_x\varphi(\hat X_s^u)
    +
    \hat \sigma_{s}D_x\varphi(\hat X_s^u)\,u_s(\hat X_s^u)
    \Bigr]\,\dd s\\
    &\qquad
    +
    \hat \sigma_{s}\,D_x\varphi(\hat X_s^u)\,\dd\bar{\W}_s.
\end{align}
Here,
\(
    \Delta_x\varphi
    :=
    (\Delta_x\varphi_1,\ldots,\Delta_x\varphi_m)
\)
is understood component-wise.
For the linear masking, stacking, and additive assembly maps used in our
experiments, \(\Delta_x\varphi=0\).

Equivalently, introducing
\begin{align}
    \hat b_s^\varphi(x)
    :=
    D_x\varphi(x)\,\hat f_s(x)+\frac12\,\hat \sigma_{s}^2\,\Delta_x\varphi(x),
    \end{align}
    we may write
    \begin{align}\label{eq: Y_SDE}
    \dd \hat Y_s^u
    =
    \Bigl[
    \hat b_s^\varphi(\hat X_s^u)+\hat \sigma_{s} D_x\varphi(\hat X_s^u)\,u_s(\hat X_s^u)
    \Bigr] \dd s
    +
    \hat \sigma_{s}\,D_x\varphi(\hat X_s^u)\,\dd\bar{\W}_s.
\end{align}

We can also recover the evolution of marginals.
For \(\hat\mu_s^{u,Y}\)-almost every \(y\), define the conditional
projected coefficients,
\begin{align}\label{coefficients of Y}
    \overline b_s^u(y)
    &:={}
    \E_{\hat{\sP}^u}\!\left[
    \hat b_s^{\varphi}(\hat X_s^u)
    +
    \hat \sigma_{s} D_x\varphi(\hat X_s^u)u_s(\hat X_s^u)
    \,\middle|\,
    \hat Y_s^u=y
    \right], \\
    \overline{Q}_s^u(y)
    &:={}
    \E_{\hat{\sP}^u}\!\left[
    \hat \sigma_{s}^2D_x\varphi(\hat X_s^u)D_x\varphi(\hat X_s^u)^\top
    \,\middle|\,
    \hat Y_s^u=y
    \right].
\end{align}
Then, for test functions $\psi\in C_c^2(\mathbb R^m)$, the one-time marginals $\hat \mu_s^{u,Y}=\Law(\hat Y_s^u)$ satisfy
\begin{align}
    \frac{\dd}{\dd s}
    \int \psi(y)\,\hat \mu_s^{u,Y}(\dd y)
    =
    \int
    \left[
    \grad_y\psi(y)\cdot \overline b_s^u(y)
    +
    \frac12\operatorname{Tr}\!\left(
    \overline{Q}_s^u(y)D_y^2\psi(y)
    \right)
    \right]
    \hat \mu_s^{u,Y}(\dd y),
\end{align}
where $D_y^2\psi$ denotes the Hessian of $\psi$. 
If
\(
    \hat\mu_s^{u,Y}(\dd y)
    =
    \hat p_s^{u,Y}(y)\dd y
\),
then \(\hat p_s^{u,Y}\) satisfies, in the sense of distributions,
\begin{eqbox}
\begin{align}\label{FP_eqn_Y}
    \partial_s \hat p_s^{u,Y}(y)
    =
    -\grad_y\cdot\bigl(\overline b_s^u(y)\hat p_s^{u,Y}(y)\bigr)
    +
    \frac12
    \sum_{j,k=1}^m
    \partial_{y_j}\partial_{y_k}
    \Bigl(
    (\overline{Q}_s^u(y))_{jk}\hat p_s^{u,Y}(y)
    \Bigr).
\end{align}
\end{eqbox}
\begin{remark}[Projected marginal dynamics and the law of the assembly]
The \cref{FP_eqn_Y} characterises only the one-time marginals of \(\hat Y^u\), and it should be distinguished from the standard Fokker-Planck equation of a Markov diffusion on the assembled space.
Even though $\hat X_s^u$ is Markovian, the assembled process $\hat Y_s^u = \varphi (\hat X^u_s)$ need not be Markovian if $\varphi$ drops information about dynamics. 
The coefficients $\overline b_s^u$ and $\overline{Q}_s^u$ defined above average over unresolved component states consistent with the same assembly state $y$. 
Under the conditions of \citet[Corollary~3.7]{brunick2010mimicking}, there exists a weak solution of an SDE with drift \(\bar b_s^u\) and covariance
\(\bar Q_s^u\) whose one-time marginals coincide with those of \(\hat Y^u\).
If, moreover, the controlled projected drift
\(
    \hat b_s^\varphi(x)
    +
    \hat\sigma_s D_x\varphi(x)u_s(x)
\)
and the projected covariance
\(
    \hat\sigma_s^2
    D_x\varphi(x)D_x\varphi(x)^\top
\)
depend on \(x\) only through \(y=\varphi(x)\), and the associated martingale problem on the assembled space is well posed, then \(\hat Y^u\) is Markov
\citep[Corollary~3.5 and Remark~3.6]{kurtz1998martingale}.
At each fixed $s$, independently of this Markov question, the law of the assembled state is simply the pushforward
\begin{align}\label{eq: Y push-forward}
\hat \mu^{u, Y}_s = \varphi_{\#} (\hat p^u_s (x) \dd x) .
\end{align}

For $m = D$, if $\varphi$ is injective and $\mathrm{det} D\varphi (x) \neq 0$ for almost every $x \in \mathbb{R}^D$ in Lebesgue measure, the law of $\hat Y^u_s$ admits a density given by
\begin{align}
    \hat p_s^{u, Y} (y) = 
\frac{\hat p_s^u(\varphi^{-1}(y))}{|\mathrm{det } D \varphi(\varphi^{-1}(y))|},
\end{align}
for almost every $y \in \varphi(\mathbb{R}^D)$ with zero density outside $\varphi(\mathbb{R}^D)$. 
More generally, for $m\leq D$, if the rank of $D \varphi$ equals $m$ for almost every $x$, an explicit expression for the density of $\hat Y$ can be given by a generalised coarea formula \citep{negro2024sample} under appropriate conditions but without injectivity.
Without these assumptions, \cref{eq: Y push-forward} still holds, but an explicit expression of the density (with respect to Lebesgue measure) may not exist.
\end{remark}


\subsection{Path-Space Change of Measure and KL Cost}\label{apdx:path_space_change_measure}

Let \(\hat{\mathbb P}:=\hat{\mathbb P}^{u=0}\) and \(\hat{\mathbb P}^{u}\) denote the path-space laws on \(C([0,1],\mathbb R^D)\) induced by the reference and controlled reverse diffusions, respectively.
Throughout this subsection, \(\hat X\) denotes the canonical coordinate process on path space.
For completeness, we allow the controlled and reference processes to have potentially different initial marginals, although \method{} uses the common-initial-law setting.
Let
\begin{align}
    e_0:
    C([0,1],\mathbb R^D)
    &\longrightarrow
    \mathbb R^D,
    \qquad
    e_0(\omega):=\omega_0,
\end{align}
and define
\[
    \nu_0:=(e_0)_{\#}\hat{\mathbb P},
    \qquad
    \nu_0^u:=(e_0)_{\#}\hat{\mathbb P}^{u}.
\]
Assume that
\(
    \nu_0^u\ll\nu_0
\),
that
\(
    D_{\mathrm{KL}}(\nu_0^u\Vert\nu_0)<\infty
\),
and that Novikov's condition holds under the reference law:
\begin{align}
    \mathbb E_{\hat{\mathbb P}}
    \left[
        \exp\!\left(
            \frac{1}{2}
            \int_0^1
            \left\|
                u_s(\hat X_s)
            \right\|^2
            \dd s
        \right)
    \right]
    <\infty.
\end{align}
Girsanov's theorem \citep{oksendal2003stochastic} then gives
\begin{align}
    \frac{
        \dd\hat{\mathbb P}^{u}
    }{
        \dd\hat{\mathbb P}
    }
    (\hat X_{\cdot})
    &=
    \frac{
        \dd\nu_0^u
    }{
        \dd\nu_0
    }
    (\hat X_0)
    \exp\!\left(
        \int_0^1
        u_s(\hat X_s)\cdot\dd\bar{\W}_s
        -
        \frac{1}{2}
        \int_0^1
        \left\|
            u_s(\hat X_s)
        \right\|^2
        \dd s
    \right),
    \label{eq:girsanov_path_change}
\end{align}
where \(\bar{\W}\) denotes the Brownian motion under the reference law
\(\hat{\mathbb P}\).
In the common case
\(
    \nu_0^u=\nu_0
\),
the initial-density factor disappears.

Taking the expectation of the log-density ratio under the controlled
law gives
\begin{align} 
    D_{\mathrm{KL}}
    \bigl(
        \hat{\mathbb P}^{u}
        \,\Vert\,
        \hat{\mathbb P}
    \bigr)
    &=
    D_{\mathrm{KL}}
    \bigl(
        \nu_0^u
        \,\Vert\,
        \nu_0
    \bigr)
    +
    \frac{1}{2}
    \mathbb E_{\hat{\mathbb P}^{u}}
    \left[
        \int_0^1
        \left\|
            u_s(\hat X_s^{u})
        \right\|^2
        \dd s
    \right].
    \label{eq:path_kl_general}
\end{align}
In particular, if
\(
    \nu_0^u=\nu_0
\),
then
\begin{align}
    D_{\mathrm{KL}}
    \bigl(
        \hat{\mathbb P}^{u}
        \,\Vert\,
        \hat{\mathbb P}
    \bigr)
    &=
    \frac{1}{2}
    \mathbb E_{\hat{\mathbb P}^{u}}
    \left[
        \int_0^1
        \left\|
            u_s(\hat X_s^{u})
        \right\|^2
        \dd s
    \right].
    \label{eq:path_kl_common_initial}
\end{align}
For a neural parameterisation using assembled features, the same
identities apply with
\[
    u_s(x)
    =
    U_\theta(x,s)
    =
    \tilde U_\theta(x,\varphi(x),s),
\]
provided that the corresponding controlled process satisfies the
admissibility and integrability conditions above.

Finally, define the pathwise assembly map
\begin{align}
    \Phi:
    C([0,1],\mathbb R^D)
    &\longrightarrow
    C([0,1],\mathbb R^m),
    \qquad
    \bigl(\Phi(\omega)\bigr)_s
    :=
    \varphi(\omega_s),
\end{align}
and the induced assembled path laws
\begin{align}
    \hat\Pi
    &:=
    \Phi_{\#}\hat{\mathbb P},
    \qquad
    \hat\Pi^u
    :=
    \Phi_{\#}\hat{\mathbb P}^{u}.
\end{align}
Since
\(
    \hat{\mathbb P}^{u}\ll\hat{\mathbb P}
\),
the data-processing inequality gives
\begin{eqbox}
\begin{align}\label{data-processing ineq}
    D_{\mathrm{KL}}
    \bigl(
        \hat\Pi^u
        \,\Vert\,
        \hat\Pi
    \bigr)
    \leq
    D_{\mathrm{KL}}
    \bigl(
        \hat{\mathbb P}^{u}
        \,\Vert\,
        \hat{\mathbb P}
    \bigr).
\end{align}
\end{eqbox}
Thus, the relative-entropy cost of controlling the assembled paths is
bounded above by the corresponding cost on the full component path
space.

\subsection{Optimal Coordinated Path and Terminal Laws}\label{apdx:proff}

The idea of the proof of the result in this section mainly follows \cite[Section 4]{domingoenrich2025adjoint}.
Recall that $\mathbb{\hat P}$ and $\mathbb{\hat P}^{u^\star}$ denote the uncontrolled reference path measure and the optimal controlled path measure, respectively.
In addition, we define an evaluation map $e_1$ at time $1$,
\begin{align}
    e_1: C(\TT,\mathbb R^D)\to\mathbb R^D,
    \qquad
    e_1(\omega)=\omega_1.
\end{align}
The marginal distributions at time $1$ are then defined as
\begin{align}
    \rho^{\star} := ({e_1})_{\#} \mathbb{\hat P}^{u^\star} \qquad \rho := ({e_1})_{\#} \mathbb{\hat P}.
\end{align}
Based on these, we can formulate the result mentioned in \cref{theory_path_view_coord} as the following result.

{\bf Proposition}: 
The optimal and reference distributions of the full path $\hat X$ satisfy
\begin{align}\label{eq: prop path}
    \mathbb{\hat P}^{u^\star}(\dd \hat X) =  \mathbb{\hat P}(\dd \hat X) \frac{\exp \left(\lambda r(\varphi(\hat X_1) )\right)}{Z(\hat X_0)},
\end{align}
where $Z(\hat X_0)$ is a constant only depending on $\hat X_0$.

Furthermore, if $\hat X_1$ is independent of $\hat X_0$, then their marginal distributions at time $1$ satisfy
\begin{align} \label{eq: prop terminal marginal}
    \rho^{\star}(\dd \hat X_1)\propto \rho(\dd \hat X_1)\exp \left(\lambda r(\varphi(\hat X_1))\right)
\end{align}
{\it Proof. } 
By the tower property and definition of KL divergence, we can rewrite our SOC objective, 
\begin{align}
    & \KL(\hat{\sP}^u\|\hat{\sP}) - \lambda \mathbb{E}_{\hat{\sP}^u}\left[r(\varphi(\hat X_1))\right] \\
    = & \mathbb{E}_{\hat X \sim \mathbb{\hat P}^u}\left[ \log\frac{\dd \hat{\sP}^u}{\dd \hat{\sP}}(\hat X) - \lambda r(\varphi(\hat X_1))\right]\\
    = & \mathbb{E}_{\hat X_0 \sim \nu_0} \left[\mathbb{E}_{\hat X \sim \hat{\sP}^u|_{\hat X_0}}\left(\log\frac{\dd \hat{\sP}^u}{\dd \hat{\sP}}(\hat X) - \lambda r(\varphi(\hat X_1))\right) \bigg|\hat X_0\right]\\
    = & \mathbb{E}_{\hat X_0 \sim \nu_0}\left[D_{\mathrm{KL}}(\mathbb{\hat P}^u|_{\hat X_0} \| \mathbb{\hat P}|_{\hat X_0}) - \lambda \mathbb{E}_{\hat X \sim \hat{\sP}^u|_{\hat X_0}} [r(\varphi(\hat X_1))]\right],
\end{align}
so the SOC objective can be rewritten as
\begin{align}
    \min_{u}\left\{\mathbb{E}_{\hat X_0 \sim \nu_0}\left[D_{\mathrm{KL}}(\mathbb{\hat P}^u|_{\hat X_0} \| \mathbb{\hat P}|_{\hat X_0}) - \lambda \mathbb{E}_{\hat X \sim \hat{\sP}^u|_{\hat X_0}} [r(\varphi(\hat X^u_1))]\right] \right\}
\end{align}

Let $\mathbb{\hat P}|_{\hat X_0}$ and $\mathbb{\hat P}^u|_{\hat X_0}$ denote path measures of the uncontrolled process in \cref{eq:uncontrolled-reverse-full-state} and the controlled processes in \cref{eq:controlled-reverse-full-state} starting from the same $\hat X_0$ respectively. 
By the Girsanov theorem, we have
\begin{align}
    \frac{\dd \mathbb{\hat P}|_{\hat X_0}}{\dd \mathbb{\hat P}^u|_{\hat X_0}} & = \exp\left( - \int _0^1 u (\hat X^u_s, s) \dd \bar{\W}_s - \frac{1}{2} \int^1_0 \|u (\hat X^u_s, s)\|^2 \dd s\right)
    \\
    \iff \log \left( \frac{\dd \mathbb{\hat P}|_{\hat X_0}}{\dd \mathbb{\hat P}^u|_{\hat X_0}} \right) & =  - \int _0^1 u (\hat X^u_s, s) \dd \bar{\W}_s - \frac{1}{2} \int^1_0 \|u (\hat X^u_s, s)\|^2 \dd s
    ,
\end{align}
then we get 
\begin{align}
    D_{\mathrm{KL}}(\mathbb{\hat P}^u|_{\hat X_0} \| \mathbb{\hat P}|_{\hat X_0}) = \mathbb{E}_{\hat X^u \sim \mathbb{\hat P}^u|_{\hat X_0}}\left[ \frac{1}{2} \int^1_0 \|u (\hat X^u_s, s)\|^2 ds \bigg| \hat X_0\right]
\end{align}
Define the path measure $\mathbb{\hat P}^{u^{\star}}$ by
\begin{align}\label{opt_control_by_def}
    \frac{\dd \hat \sP^{u^\star}|_{\hat X_0}}{\dd \hat \sP |_{\hat X_0}} := \frac{\exp(\lambda r(\varphi(\hat X_1)))}{Z(\hat X_0)}, \qquad Z(\hat X_0) := \mathbb{E}_{\hat X \sim \mathbb{\hat P}|_{\hat X_0}}\left[\exp(\lambda r(\varphi(\hat X_1)))\right],
\end{align}
Notice that
\begin{align}
\KL(\mathbb{\hat P}^{u}|_{\hat X_0} || \mathbb{\hat P}^{u^{\star}}|_{\hat X_0}) & = \mathbb{E}_{\hat X \sim \mathbb{\hat P}^u|_{\hat X_0}}\left[\log \frac{\dd \mathbb{\hat P}^u|_{\hat X_0}}{\dd \mathbb{\hat P}^{u^\star}|_{\hat X_0}}\right]\\
& = \mathbb{E}_{\hat X \sim \mathbb{\hat P}^u|_{\hat X_0}}\left[ \log \frac{\dd \mathbb{\hat P}^u|_{\hat X_0}}{\dd \mathbb{\hat P}|_{\hat X_0}} - \lambda r(\varphi(\hat X_1)) + \log(Z(\hat X_0))\right]\\
& = \KL(\mathbb{\hat P}^{u}|_{\hat X_0} || \mathbb{\hat P}|_{\hat X_0}) - \mathbb{E}_{\hat X \sim \mathbb{\hat P}^u|_{\hat X_0}}[\lambda r(\varphi(\hat X_1))] + \log(Z(\hat X_0))
\end{align}

Rearranging terms, we get
\begin{align}
    \KL(\mathbb{\hat P}^{u}|_{\hat X_0} || \mathbb{\hat P}|_{\hat X_0}) - \mathbb{E}_{\hat X \sim \mathbb{\hat P}^u|_{\hat X_0}}[\lambda r(\varphi(\hat X^u_1))] = \KL(\mathbb{\hat P}^{u}|_{\hat X_0} || \mathbb{\hat P}^{u^{\star}}|_{\hat X_0}) - \log(Z(\hat X_0)).
\end{align}

Since relative entropy is non-negative and vanishes if and only if its arguments coincide, the conditional measure-level objective is uniquely minimised when \( \hat{\mathbb P}^{u}_{\vert\hat X_0} = \hat{\mathbb P}^{\star}_{\vert\hat X_0}.\)
Hence, \(\hat{\mathbb P}^{u^\star}_{\vert \hat X_0}\) is the unique minimiser of the conditional path-space objective.

\Cref{opt_control_by_def} is equivalent to
\begin{align}
    \hat \sP^{u^\star}|_{\hat X_0}(\dd \hat X) 
    = \frac{\exp(\lambda r(\varphi(\hat X_1)))} {Z(\hat X_0)} \hat \sP |_{\hat X_0} (\dd \hat X)
\end{align}
Multiplying both sides of the equation by $\nu_0 (\hat X_0)$, we get
\begin{align}
\begin{aligned}
    \hat \sP^{u^\star}(\dd \hat X) 
    &= \hat \sP^{u^\star}|_{\hat X_0}(\dd \hat X) \, \nu_0(\dd \hat X_0) \\
    & = \frac{\exp(\lambda r(\varphi(\hat X_1)))}{Z(\hat X_0)} \hat \sP |_{\hat X_0} (\dd \hat X)\, \nu_0(\dd \hat X_0) \\
    & = \frac{\exp(\lambda r(\varphi(\hat X_1)))}{Z(\hat X_0)} \hat \sP (\dd \hat X),
\end{aligned}
\end{align}
which finishes the proof of \cref{eq: prop path}.

Applying the endpoint map
\[
    e_{0,1}:
    C([0,1],\mathbb R^D)
    \to
    \mathbb R^D\times\mathbb R^D,
    \qquad
    e_{0,1}(\omega):=(\omega_0,\omega_1),
\]
to \cref{eq: prop path} yields

\begin{align}
    (e_{0,1})_{\#}\hat{\mathbb P}^{u^\star}
    (\dd x_0,\dd x_1)
    &=
    \frac{
        \exp\!\left(
            \lambda r\!\left(\varphi(x_1)\right)
        \right)
    }{
        Z(x_0)
    }
    (e_{0,1})_{\#}\hat{\mathbb P}
    (\dd x_0,\dd x_1).
\end{align}

Under the assumed independence of \(\hat X_0\) and \(\hat X_1\) under the reference law, as in the idealised complete-noising limit, we can split $\hat \sP (\dd \hat X_0, \dd \hat X_1) = \nu_0(\dd \hat X_0) \rho(\dd \hat X_1)$, and then evaluate the expectation on $\hat X_0$ to get
\begin{eqbox}
\begin{align}
    \rho^{\star}(\dd \hat X_1) &= \mathbb{E}_{\hat X_0 \sim \nu_0}\left[\frac{\exp(\lambda r(\varphi(\hat X_1)))}{Z(\hat X_0)} \rho(\dd \hat X_1)\right]\\
    &=\rho(\dd \hat X_1)\exp\left(\lambda r(\varphi(\hat X_1))\right)\mathbb{E}_{\hat X_0 \sim \nu_0}\left[\frac{1}{Z(\hat X_0)} \right]\\
    & \propto \rho(\dd \hat X_1)\exp \left(\lambda r(\varphi(\hat X_1))\right),
\end{align}
\end{eqbox}
which finishes the proof of \cref{eq: prop terminal marginal}.

\begin{remark}[Reward-induced coordination]
\label{rem:reward_induced_coordination}
In the idealised complete-noising limit, where \(\hat X_0 \) and \(\hat X_1\) are statistically independent, marginalising the optimal controlled path law yields the terminal distribution,
\begin{equation}
    \rho^\star(x)
    \propto
    \rho(x)\,
    \exp\!\left(
        \lambda r\!\left(\varphi(x)\right)
    \right),
    \qquad x\in\mathbb R^D,
    \label{eq:terminal_reward_tilt}
\end{equation}
where
\(
    \rho
    =
    \tilde{\pi}^{1}
    \otimes \cdots \otimes
    \tilde{\pi}^{N}
\)
is the terminal law of the independently sampled pretrained agents.
When \(r \circ \varphi\) is non-separable across components, this reward tilt breaks the product structure of \(\rho\) and introduces the task-specific dependencies required by the assembled system.
The learned control provides a dynamic realisation of this change of measure. 
Rather than explicitly re-weighting complete pretrained trajectories, it steers the joint reverse dynamics towards the reward-tilted terminal law.
%
\end{remark}

\paragraph{Relation to probabilistic couplings.}
A coupling of \(\tilde{\pi}^{1},\ldots,\tilde{\pi}^{N}\)
is a joint law on the product space whose \(i\)-th marginal is
\(\tilde{\pi}^{i}\).
The reference product law
\(
\tilde{\pi}^{1}\otimes\cdots\otimes\tilde{\pi}^{N}
\)
is their independent coupling.
By contrast, the reward tilt above generally changes both the dependence structure and the component marginals, and the optimal terminal law is therefore not, in general, a coupling of the original pretrained laws.
We instead use {coupled dynamics} to describe the system constructed through a joint control whose component-wise actions may depend on the full multi-agent state.

\subsection{Pushforward to the Assembled Process}\label{apdx:assembly_change_measure}

Starting from the optimal path-law characterisation established in \cref{apdx:proff}, we now examine the corresponding change of measure after pushing the reference and controlled processes through the assembly map \(\varphi\).
This distinction is particularly relevant when \(\varphi\) is non-injective, since multiple component trajectories may correspond to the same assembled trajectory.

By similar results discussed in \cite{blessing2025trust}, and \cite{domingo2024stochastic}, we already have that, for the objective and terminal cost we considered
\begin{align}
    \mathcal{J}(u)&:= \mathbb E_{\hat{\mathbb P}^{u}}
    \left[
        \int_0^1
         C(\hat X^u_s, s)+ \frac{1}{2}\|u_{ s}(\hat X^u_s)\|^2 \, \dd s 
        +
        g(\hat X^u_1)
    \right]\\
    & = \mathbb E_{\hat{\mathbb P}^{u}}
    \left[
        \int_0^1
        \frac{1}{2}\|u_{ s}(\hat X^u_s)\|^2 \, \dd s
        -
        \lambda r\!\left(\varphi(\hat X^u_1)\right),
    \right]
\end{align}
where $C$ and $g$ denote general running cost and terminal cost, respectively, and we consider zero running cost and take the terminal cost to be negative reward $g = -\lambda r\circ\varphi$ in this paper. The parametrised version of the objective expression is \cref{eq:adjoint-objective}.
If the function $g(x)$ is regular enough, the Radon--Nikodym derivative of the optimal path measure $\mathbb{\hat P}^{u^\star}$ is given as
\begin{align}
    \frac{\mathrm d \mathbb{\hat P}^{u^\star}}{\mathrm d\mathbb{\hat P}}(\hat X)
    =
    \frac{\exp\big(-\mathcal{W}(\hat X)\big)}{Z(\hat X_0)},
\end{align}
where $\mathbb{\hat P}$ is the path measure of the uncontrolled process $\hat X$, $Z$ is a normalizing constant and 
\begin{align*}
    \mathcal{W}(\hat X) & := \int_0^1 C(\hat X_s,s) \,\dd s + g(\hat X_1)\\
    &= - \lambda r( \varphi(\hat X_1)),
\end{align*}

Equivalently, their densities give
$$
    p_{\hat{X}^{u^\star}} (x(\cdot)) =  \frac{\exp\left( \lambda r(\varphi(x_1))\right)}{Z(x_0)} p_{\hat{X}^{u=0}}(x(\cdot)) ,
$$
where $x(\cdot)=\{x(s)\}_{s\in [0,1]}$ is a whole trajectory rather than a single point.
This means that we can use the above relation to generate samples of $\hat X^{u^\star}$ and $\hat X$, and then apply $\varphi$ on those generated samples to get samples of $\hat Y^{u^\star}$ and $\hat Y$ respectively.

If $\varphi$ is injective, then we further get
\begin{align}
    \frac{\mathrm{d}(\varphi_\# \mathbb{\hat P}^{u^\star})}{\mathrm{d}(\varphi_\#\mathbb{\hat P})} (\hat Y) 
    &= \frac{1}{Z(\varphi^{-1}(\hat Y_0))} \exp(\lambda r(\hat Y_1)) \exp \left(-\int_0^1 C(\varphi^{-1}(\hat Y_s), s) \mathrm d s\right) \\
    \label{eq: inj path RND Y}
    &= \frac{\exp(\lambda r(\hat Y_1))}{Z(\varphi^{-1}(\hat Y_0))}
\end{align}
where $\varphi_\# \mathbb{\hat P}^{u^\star}$ and $\varphi_\#\mathbb{\hat P}$ denote path measures of optimal control $\hat Y^{u^\star}$ and uncontrolled $\hat Y$ respectively. 
They are actually just push-forward measures of $\mathbb{\hat P}^{u^\star}$ and $\mathbb{\hat P}$ under $\varphi$ respectively. 

More generally, without requiring $\varphi$ to be injective, we have
\begin{eqbox}
\begin{align}
   \frac{\mathrm d (\varphi_\# \mathbb{\hat P}^{u^\star})}{\mathrm d(\varphi_\#\mathbb{\hat P})} (\hat Y) 
   &= \mathbb{E_{\hat P}}\left[\frac{\exp\left(\lambda r(\varphi(\hat X_1))\right)}{Z(\hat X_0)} \exp \left(-\int_0^1  C(\hat X_s, s) \mathrm d s\right)\bigg|\varphi(\hat X) = \hat Y\right]\\
   & = \mathbb{E_{\hat P}}\left[\frac{\exp\left(\lambda r(\hat Y_1)\right)}{Z(\hat X_0)} \, \bigg| \, \varphi(\hat X) = \hat Y\right]\\
   \label{eq: general path RND Y}
   &=\exp\left(\lambda r(\hat Y_1)\right)\mathbb{E_{\hat P}}\left[\frac{1}{{Z(\hat X_0)}} \bigg|\varphi(\hat X) = \hat Y\right]
   ,
\end{align}
\end{eqbox}
where notations without subscripts denote the whole paths or processes. When $\varphi$ is injective, the assembled path determines the component path through $\varphi^{-1}(\hat{Y}_s)$. Consequently, the conditional expectation in \cref{eq: general path RND Y} reduces to $1/Z(\varphi^{-1}(\hat Y_0))$ and recovers \cref{eq: inj path RND Y}.

\section{Adjoint Matching Optimisation}\label{apdx:adjoint_matching}
\begin{takehome}
\textbf{Key Takeaways:}
\textbf{The adjoint turns the global assembly objective into local learning signals.}
Starting from the terminal cost gradient, it propagates feedback backwards along the reverse trajectory, accounting for the remaining control cost and cross-agent dependencies introduced by the joint controller (\cref{apdx: full pathwise adjoint}).
\textbf{We retain the full adjoint, rather than the lean variant.}
Backward replay differentiates both the controlled transition and the quadratic control cost, retaining all state dependencies not removed by the task-specific detach conventions (\cref{apdx:adj_matching}).
\textbf{Discrete matching uses noise-normalised controls and next-state adjoints.}
At each sampling step, the controller-induced state displacement is normalised by the transition noise scale and matched to the negative next-state adjoint multiplied by that same scale. Rollout states and adjoint targets are held fixed during regression 
(Algorithm~\ref{alg:cmds-discrete-am}).
\end{takehome}

We use reverse, generative time $s\in\TT$. 
In this subsection, $\hat f_s$ denotes the reference reverse drift and $\hat \sigma_s$ denotes the corresponding reverse-time diffusion scale. 
Let
\begin{align}
    \hat X_s^{\theta}=(\hat X_s^{\theta,1},\ldots,\hat X_s^{\theta,N}),
    \qquad
    \hat Y_s^{\theta}=\varphi(\hat X_s^{\theta},s ), 
\end{align}
where $\varphi: \mathbb{R}^D \times [0,1] \rightarrow \mathbb{R}^m$ can be a time-dependent assembly function in general cases.
The controlled \method{} process is
\begin{align}
    \dd \hat X_s^{\theta}
    =
    \underbrace{
        \left[
        \hat f_s(\hat X_s^\theta)
        + \hat \sigma_s U_{\theta}(\hat X_s^\theta,s)
        \right]
    }_{\eqqcolon\,\hat f_{\theta}(\hat X_s^\theta,s)}
    \,\dd s
    +
    \hat \sigma_s\,\dd \bar{\W}_s.
    \label{eq:cmad-controlled-vector}
\end{align}

Here,
\begin{align}
    U_{\theta}(x,s)
    =
    \bigl(U_{\theta}^1(x,s),\ldots,U_{\theta}^N(x,s)\bigr)
    \in\mathbb R^D
\end{align}
is the actual full-state normalised control field. The block form of \cref{eq:cmad-controlled-vector} is
\begin{align}
    \dd \hat X_s^{\theta,i}
    =
    \left[
    \hat f^i
    (\hat X_s^{\theta},s)
    +
    \hat \sigma_sU_{\theta}^i(\hat X_s^{\theta},s)
    \right]\dd s
    +
    \hat \sigma_s\,\dd \bar{\W}_s^i,
    \qquad
    i=1,\ldots,N.
\end{align}
If the reference component diffusions are independent, then $\hat f^i(\hat X_s^{\theta},s)=\hat f^i(\hat X_s^{\theta,i},s)$; the derivation below does not require this simplification.

The approximated objective in the neural network is
\begin{align}
    \mathcal J(\theta)
    &=
    \E_{\hat{\sP}^{\theta}}\left[
    \int_0^1 C(\hat X_s^{\theta},s) + \frac 12 \|U_\theta(\hat X^\theta_s, s)\|^2\,\dd s
    +
    g(\hat X_1^{\theta})
    \right],
    \label{eq:adjoint-objective}
\end{align}
where \(C\) is an additional running cost and \(g=-\lambda r\circ\varphi\) is the terminal cost. 
In this work, we consider the relatively simpler case $\varphi(x, s) = \varphi(x)$ which does not change with time, and treat $\varphi(x, s)$ as $\varphi(x)$ in the following derivations.
\subsection{Full Pathwise Adjoint} \label{apdx: full pathwise adjoint}
Fix a sampled controlled path $\hat X^{\theta}$. The full adjoint corresponds to the gradient of the cost-to-go from $\hat X_s$ 
\begin{align}
    a_s
    & :=
    \grad_{\hat X_s}
    \left[
    \int_s^1 C(\hat X_\tau^{\theta},\tau) + \frac 12 \|U_\theta(\hat X^\theta_\tau, \tau)\|^2\,\dd\tau
    +
    g(\hat X_1^{\theta})
    \right]
    .
\end{align}
With column-vector adjoints and the Jacobian convention, we have:
\begin{align}
    D_xF(x)=\left(\partial_{x_j}F_i(x)\right)_{i,j}.
\end{align}
The full adjoint solves
\begin{align}
   \label{eq:cmad-full-adjoint}
    \frac{\dd}{\dd s}a_s
    &=
    -D_x\hat f_{\theta}(\hat X_s^{\theta},s)^\top a_s
    -\grad_x\left(C(\hat X_s^{\theta},s)+ \frac12\|U_\theta(\hat X^\theta_s, s)\|^2\right), \\
    \label{eq:cmad-terminal-adjoint}
    a_1
    &=
    \grad_x g(\hat X_1^{\theta})
    =
    -\lambda
    D_x\varphi(\hat X_1^{\theta} )^\top
    \grad_y r(\hat Y_1^{\theta}).
\end{align}

We can further develop the expression of the Jacobian and the gradient:
\begin{align}
    D_x\hat f_{\theta}(x,s)
    =
    D_x\hat f_s(x)+\hat \sigma_sD_xU_{\theta}(x,s).
    \label{eq:jacobian-controlled-drift}
\end{align}
Moreover,
\begin{align}
    \grad_x\left(C(x,s)+ \frac12\|U_\theta(x, s)\|^2\right)
    =
    D_xU_{\theta}(x,s)^\top U_{\theta}(x,s)
    +
    \grad_x C(x,s),
    \label{eq:grad-C}
\end{align}
where $D_xU_{\theta}^\top U_{\theta}$ denotes the gradient of $\frac12\|U_{\theta}\|^2$.

Write $a_s=(a_s^1,\ldots,a_s^N)$ with $a_s^i\in\Rd$. 
The equivalent block-wise representation of \cref{eq:cmad-full-adjoint} is given as 
\begin{align}
    \frac{\dd}{\dd s}a_s^i
    &=-
    \sum_{j=1}^N
    \left[
    D_{x^i}\hat f^j_s(\hat X_s^{\theta})
    +
    \hat \sigma_sD_{x^i}U_{\theta}^j(\hat X_s^{\theta},s)
    \right]^\top
    a_s^j
    \\
    &\quad-
    \sum_{j=1}^N
    \left[
    D_{x^i}U_{\theta}^j(\hat X_s^{\theta},s)
    \right]^\top
    U_{\theta}^j(\hat X_s^{\theta},s)
    \\
    &\quad-
    \partial_{x^i} C(\hat X^\theta_s,s),
    \label{eq:block-adjoint-full}
\end{align}
with terminal condition
\begin{align}
    a_1^i
    =
    -\lambda
    D_{x^i}\varphi(\hat X_1^{\theta} )^\top
    \grad_y r(\hat Y_1^{\theta})
    \label{eq:block-terminal-general}
\end{align}
for each $i$.

\subsection{Adjoint Matching Instantiation of \method{}}\label{apdx:adj_matching}
Under the assumptions of \citet{domingoenrich2025adjoint} (Proposition~2 and Appendices~E.2 and~E.3), the adjoint matching
loss in \cref{eq:cmad-basic-am} and the original objective in \cref{eq:adjoint-objective} share the same gradient with respect
to $\theta$.
Moreover, in the corresponding control-space formulation, the unique critical point of the adjoint matching loss in \cref{eq:cmad-basic-am} is the optimal control for the original objective in \cref{eq:adjoint-objective}.

Let $\bar{\theta}\coloneqq\operatorname{stopgrad}(\theta)$, and sample paths from the frozen controlled process
\begin{align}
    \hat X^{\bar{\theta}}\sim \hat{\sP}^{\bar{\theta}}.
\end{align}
Let $a^{\bar{\theta}}$ solve \cref{eq:cmad-full-adjoint}--\cref{eq:cmad-terminal-adjoint} along $\hat X^{\bar{\theta}}$, with all dependence on $\bar{\theta}$ stopped. 
The basic \method{} adjoint matching objective used in our work and shown in the main part of this paper is
\begin{align}
    \mathcal L_{\rm \method{}\text{-}AM}(\theta)
    =
    \frac12
    \E_{\hat{\sP}^{\bar{\theta}}}\left[
    \int_0^1
    \left\|
    U_{\theta}(\hat X_s^{\bar{\theta}}, s)
    +
    \hat \sigma_s a_s^{\bar{\theta}}
    \right\|^2
    \,\dd s
    \right].
    \tag{\ref{eq:cmad-basic-am}}
\end{align}

Block-wise,
\begin{align}
    \mathcal L_{\rm \method{}\text{-}AM}(\theta)
    =
    \frac12
    \E_{\hat{\sP}^{\bar{\theta}}}\left[
    \int_0^1
    \sum_{i=1}^N
    \left\|
    U_\theta^i(\hat X_s^{\bar\theta},s)
    +
    \hat\sigma_s a_s^{\bar\theta,i}
    \right\|^2
    \,\dd s
    \right].
\end{align}
The gradients of $\mathcal J$ can be recovered using the adjoint \citep{domingoenrich2025adjoint}.
\begin{align}
    \grad_{\theta}\mathcal J(\theta)
    =
    \E\left[
    \int_0^1
    D_{\theta} U_{\theta}(\hat X_s^{\theta}, s)^\top
    \Bigl(
    U_{\theta}(\hat X_s^{\theta}, s)+\hat \sigma_s a_s
    \Bigr)
    \,\dd s
    \right].
\end{align}

\subsubsection{Relation to Lean Adjoint Matching and Agent-Wise Structure}
The adjoint matching formulation above uses the full adjoint, so it keeps the same gradient with respect to the parameter \(\theta\) as direct optimisation over the original stochastic control objective. 
\cite{domingoenrich2025adjoint} also proposed a lean adjoint, obtained by removing terms depending on control from the adjoint dynamics, since their expectation equals $0$ at the optimum. 
Consequently, the resulting matching loss has a different gradient from the original SOC objective in general away from the optimum, but the unique optimum is proven to be the same as the optimal control under their continuous control-affine assumptions. 
In \method{}, it is formulated as a control-affine SOC problem, and thus the same lean adjoint construction can be applied to the continuous time \method{} formulation.
Specifically, the lean version of the original full adjoint dynamics (\cref{eq:cmad-full-adjoint}) is
\begin{align}
    \frac{\dd \tilde{a}_s}{\dd s} = - \left [D_x \hat f_s(\hat X_s^\theta)\right]^\top \tilde{a}_s
    -\grad_x C(\hat X_s^{\theta},s),
\end{align}
while $\tilde a_1$ is defined to be the same as $a_1$ in \cref{eq:cmad-terminal-adjoint}. 
The corresponding adjoint matching objective is
\begin{align}
    \mathcal L_{{\rm Lean AM}}(\theta)
    =
    \frac12
    \E_{\hat{\sP}^{\bar{\theta}}}\left[
    \int_0^1
    \left\|
    U_{\theta}(\hat X_s^{\bar{\theta}}, s)
    +
    \hat \sigma_s \tilde{a}_s^{\bar{\theta}}
    \right\|^2
    \,\dd s
    \right].
\end{align}

\paragraph{Agent-wise decomposition} The simplification is natural for the reference product process. As pretrained agents evolve independently, the state Jacobian of $\hat f_s(x)$ is in the block diagonal form
$$
D_x \hat f_s(x) = \mathrm{diag}(D_{x^1} \hat f^1_s(x^1), D_{x^2} \hat f^2_s(x^2), ..., D_{x^N}\hat f^N_s(x^N)),
$$
and thus the corresponding lean adjoint in agent-wise form satisfies
\begin{align}
    \frac{\dd \tilde{a}_s^i}{\dd s} = - \left [D_{x^i} \hat f_s^{\,i}(\hat X_s^{\theta,i})\right]^\top \tilde{a}^i_s
    -\grad_{x^i} C(\hat X_s^{\theta},s).
\end{align}
In the main \method{} setting considered in this work, $C = 0$, and thus the backward propagation of $\tilde a$ separates among agents after the terminal adjoint has been initialised. Importantly, this does not remove the coordination as 
\begin{align}
    \tilde{a}_1^i =  -\lambda
    D_{x^i}\varphi(\hat X_1^{\theta} )^\top
    \grad_y r(\hat Y_1^{\theta}).
\end{align}
Therefore, a precise interpretation is that the assembly reward performs joint credit assignment across agents at the terminal time, and the lean adjoint dynamics subsequently propagate the information assigned to each agent backward through each agent’s own frozen dynamics.

\paragraph{Computational Interpretation}
For factorised reference dynamics, the lean-adjoint propagation only needs agent-wise vectors.
If the cost of each component model is fixed, the computational cost of this lean-adjoint backward propagation grows linearly with respect to the number of agents, and can be parallelly computed. 
On the other hand, the full adjoint additionally differentiates through the joint control $U_\theta$, and its block form contains cross-agent terms $D_{x^i}U^j_{\theta}$ representing dependencies introduced by the coordination network.
The lean adjoint matching removes these interactions from the adjoint propagation through the reference dynamics.
Note that this does not imply that the overall \method{} computation is necessarily linear in the number of agents $N$, as the evaluation of the assembly reward and joint control could potentially involve higher-order interactions among different agents. 

\paragraph{Relation to our implementation}
For the reasons discussed above, in this work we retain the full adjoint rather than the lean variant.
In particular, the discrete DDIM recursion below differentiates the controlled transition and includes the quadratic control-cost derivative, thereby preserving the full cross-agent dependencies that are not detached by the task-specific implementation. 
A discrete lean \method{} variant could instead propagate adjoints through the frozen base transitions only, in analogy with the lean DDIM construction of \cite{domingoenrich2025adjoint}.

\subsubsection{Discrete-time adjoint matching for DDIM sampling}\label{apdx:ddim-adjoint-matching}
We use the rollout convention of \cref{alg:cmds-discrete-am}: \(K\) is the number of DDIM transitions, \(\ell=0,\ldots,K-1\) indexes transitions, and \(\zeta_0>\cdots>\zeta_{K-1}\) are their diffusion time-steps.
The rollout runs from the initial noise state \(X_0\) to the terminal sample \(X_K\).
Let
\(
X_\ell=(X_\ell^1,\ldots,X_\ell^N)\in\mathbb R^D
\)
denote the current realised joint state along the reverse DDIM trajectory, with the control superscript suppressed, and let
\[
    Y_\ell:=\varphi(X_\ell),
    \qquad
    \Xi_\ell=(\Xi_\ell^1,\ldots,\Xi_\ell^N)
    \sim\N(0, I_D),
\]
with
\(
    \Xi_\ell^i\sim\N(0, I_d)
\).
For agent \(i\), parameterise the fine-tuned denoiser as
\begin{align}
    \epsilon_{\theta,\ell}^i(x,y)
    &:=
    \epsilon^i_{{\rm b},\ell}(x^i)
    +
    \delta\epsilon_{\theta,\ell}^i(x,y),
    \qquad
    x=(x^1,\ldots,x^N),
    \qquad
    y=\varphi(x),
    \label{eq:cmds-ddim-denoiser}
\end{align}
where
\(
    \epsilon^i_{{\rm b},\ell}(x^i)
    :=
    \epsilon^i_{\rm b}(x^i,\zeta_\ell)
\).
The control parameters are shared across DDIM transitions; the native diffusion timestep \(\zeta_\ell\) is supplied through the time input rather than indexing a separate parameter vector.
In the two-agent implementation,
\[
    \delta\epsilon_{\theta,\ell}^1
    =
    C_{\theta_1}(x^1,y,x^2,\zeta_\ell),
    \qquad
    \delta\epsilon_{\theta,\ell}^2
    =
    C_{\theta_2}(x^2,y,x^1,\zeta_\ell).
\]

For each DDIM transition, let \(\bar\alpha_{\mathrm{src},\ell}\) and \(\bar\alpha_{\mathrm{dst},\ell}\) denote its source and destination cumulative alphas, respectively.
Here \(\bar\alpha_{\mathrm{src},\ell}=\bar\alpha_{\zeta_\ell}\), and \(\bar\alpha_{\mathrm{dst},\ell}=\bar\alpha_{\zeta_{\ell+1}}\) for \(\ell<K-1\); the terminal destination value is modified as specified below.
For \(\eta>0\), define
\begin{align}
    \varsigma_\ell
    &:=
    \eta
    \left[
        \frac{1-\bar\alpha_{\mathrm{dst},\ell}}{1-\bar\alpha_{\mathrm{src},\ell}}
        \left(
            1-\frac{\bar\alpha_{\mathrm{src},\ell}}{\bar\alpha_{\mathrm{dst},\ell}}
        \right)
    \right]^{1/2},
    \\
    \gamma_\ell
    &:=
    \left[
        1-\bar\alpha_{\mathrm{dst},\ell}-\varsigma_\ell^2
    \right]_+^{1/2}
    -
    \left(
        \frac{\bar\alpha_{\mathrm{dst},\ell}}{\bar\alpha_{\mathrm{src},\ell}}
    \right)^{1/2}
    \left(
        1-\bar\alpha_{\mathrm{src},\ell}
    \right)^{1/2}.
    \label{eq:cmds-ddim-transition-coefficients}
\end{align}
The component DDIM map is
\begin{align}
    F_\ell(z,\epsilon,\xi)
    &:=
    \left(
        \frac{\bar\alpha_{\mathrm{dst},\ell}}{\bar\alpha_{\mathrm{src},\ell}}
    \right)^{1/2}
    z
    +
    \gamma_\ell\epsilon
    +
    \varsigma_\ell\xi.
    \label{eq:cmds-ddim-transition}
\end{align}
At the terminal reverse transition \(\ell=K-1\), we use
\(
    \bar\alpha_{\mathrm{dst},\ell}=(1+\bar\alpha_{\mathrm{src},\ell})/2
\),
rather than \(\bar\alpha_{\mathrm{dst},\ell}=1\).  Hence \(\varsigma_\ell>0\) at that transition as well, and its control cost is included.

The base and controlled component updates, evaluated from the same current state and using the same noise, are
\begin{align}
    X_{\ell+1}^{{\rm b},i}
    &:=
    F_\ell\!\left(
        X_\ell^i,
        \epsilon^i_{{\rm b},\ell}(X_\ell^i),
        \Xi_\ell^i
    \right),
    \\
    X_{\ell+1}^{\theta,i}
    &:=
    F_\ell\!\left(
        X_\ell^i,
        \epsilon_{\theta,\ell}^i(X_\ell,Y_\ell),
        \Xi_\ell^i
    \right).
\end{align}

Using the same noise
\(
    \Xi_\ell^i\sim\N(0,I_d)
\)
in the base and controlled DDIM steps, the one-step displacement induced by the epsilon residual is
\begin{align}
    X_{\ell+1}^{\theta,i}
    -
    X_{\ell+1}^{{\rm b},i}
    &=
    \gamma_\ell
    \delta\epsilon_{\theta,\ell}^i(X_\ell,Y_\ell).
    \label{eq:cmds-ddim-displacement}
\end{align}
The normalised discrete control is therefore
\begin{align}
    w_{\theta,\ell}^i(X_\ell)
    &:=
    \frac{1}{\varsigma_\ell}
    \left(
        X_{\ell+1}^{\theta,i}
        -
        X_{\ell+1}^{{\rm b},i}
    \right)
    \nonumber\\
    &=
    \frac{\gamma_\ell}{\varsigma_\ell}
    \delta\epsilon_{\theta,\ell}^i
    \bigl(X_\ell,\varphi(X_\ell)\bigr)
    \approx
    \sqrt{\Delta s_\ell}\,
    U_\theta^i(X_\ell,s_\ell),
    \label{eq:cmds-ddim-control}
\end{align}
where \(\varsigma_\ell>0\) denotes the stochastic scale of DDIM transition \(\ell\).
For the continuous-time interpretation, \(s_\ell\) belongs to a generative-time grid \(0=s_0<\cdots<s_K=1\), with \(\Delta s_\ell:=s_{\ell+1}-s_\ell>0\).
The equality preceding the approximation is exact; the final relation is the continuous-time interpretation of the discrete control.
Consequently,
\[
    \frac{1}{2}
    \sum_{i=1}^N
    \left\|
        w_{\theta,\ell}^i(X_\ell)
    \right\|^2
\]
is the one-step discrete SOC control cost.

Define the joint base and controlled DDIM maps by
\begin{align}
    F_\ell^{\rm b}(X_\ell,\Xi_\ell)
    &:=
    \left(
        X_{\ell+1}^{{\rm b},1},
        \ldots,
        X_{\ell+1}^{{\rm b},N}
    \right),
    \\
    F_\ell^\theta(X_\ell,\Xi_\ell)
    &:=
    \left(
        X_{\ell+1}^{\theta,1},
        \ldots,
        X_{\ell+1}^{\theta,N}
    \right)
    =
    X_{\ell+1}^\theta.
\end{align}
The base map is a one-step counterfactual used to define the discrete control; only the controlled map advances the sampled trajectory.

As in the continuous-time formulation, let
\(
    \bar\theta=\operatorname{stopgrad}(\theta)
\),
and let
\(
    \{X_\ell^{\bar\theta}\}_{\ell=0}^{K}
\)
denote a trajectory generated by the frozen controlled DDIM maps
\(F_\ell^{\bar\theta}\).
Define the terminal cost
\[
    g(x):=-\lambda r\!\left(\varphi(x)\right).
\]
The full discrete adjoint used by the \method{} objective is initialised at \(K\) and propagated for \(\ell=K-1,\ldots,0\):
\begin{align}
    a_K^{\bar\theta}
    &=
    \grad_x g(X_K^{\bar\theta})
    \nonumber\\
    &=
    -\lambda
    D_x\varphi(X_K^{\bar\theta})^\top
    \grad_y r\!\left(\varphi(X_K^{\bar\theta})\right),
    \label{eq:cmds-ddim-terminal-adjoint}
    \\
    a_\ell^{\bar\theta}
    &=
    D_xF_\ell^{\bar\theta}
    \bigl(X_\ell^{\bar\theta},\Xi_\ell\bigr)^\top
    a_{\ell+1}^{\bar\theta}
    +
    \left.
    \grad_x
    \left[
        \frac{1}{2}
        \sum_{i=1}^N
        \left\|
            w_{\bar\theta,\ell}^i(x)
        \right\|^2
    \right]
    \right|_{x=X_\ell^{\bar\theta}}.
    \label{eq:cmds-discrete-full-adjoint}
\end{align}
Here \(a_{\ell+1}^{\bar\theta}\) is the costate of the post-transition state \(X_{\ell+1}^{\bar\theta}\); it is paired with the action at transition \(\ell\) before the recursion is propagated to \(a_\ell^{\bar\theta}\).

Equivalently, define
\begin{align}
    \Delta_\ell^{\bar\theta}(x,\xi)
    &:=
    F_\ell^{\bar\theta}(x,\xi)-x.
\end{align}
Then the adjoint recursion can be written as
\begin{align}
    a_\ell^{\bar\theta}
    &=
    a_{\ell+1}^{\bar\theta}
    +
    \left.
    \grad_x
    \left[
        \left\langle
            a_{\ell+1}^{\bar\theta},
            \Delta_\ell^{\bar\theta}(x,\Xi_\ell)
        \right\rangle
        +
        \frac{1}{2}
        \sum_{i=1}^N
        \left\|
            w_{\bar\theta,\ell}^i(x)
        \right\|^2
    \right]
    \right|_{x=X_\ell^{\bar\theta}}.
\end{align}
Because the DDIM stochastic term
\(
    \varsigma_\ell\Xi_\ell
\)
is additive and independent of the current state, the state derivative in the adjoint recursion is unchanged if \(\Xi_\ell\) is set to zero during adjoint re-computation.

The discrete \method{} adjoint matching loss is
\begin{align}
    \mathcal L_{\method{}}(\theta)
    &=
    \E
    \left[
        \sum_{\ell\in\mathcal K}
        \dfrac{1}{2}
        \sum_{i=1}^N
        \left\|
            w_{\theta,\ell}^i
            \bigl(
                \operatorname{stopgrad}(X_\ell^{\bar\theta})
            \bigr)
            +
            \varsigma_\ell
            \operatorname{stopgrad}
            \bigl(
                a_{\ell+1}^{\bar\theta,i}
            \bigr)
        \right\|^2
    \right],
    \label{eq:cmds-ddim-am-loss}
\end{align}
where \(\mathcal K\subseteq\{0,\ldots,K-1\}\) is the set of stochastic DDIM transitions selected for the matching loss.
The adjoint recursion itself is evaluated over all DDIM transitions; when \(\mathcal K\) is a strict subset, the matching terms are not importance re-weighted.
The implementation evaluates \cref{eq:cmds-discrete-full-adjoint} through vector--Jacobian products of the controlled joint map \(F_\ell^{\bar\theta}\), while including
\(
    \frac{1}{2}
    \sum_i
    \|w_{\bar\theta,\ell}^i\|^2
\)
in the adjoint recursion. 
The complete discrete-time adjoint-matching procedure, including the controlled rollout, backward adjoint replay, and control regression, is summarised in \Cref{alg:cmds-discrete-am}.
It therefore implements the full discrete adjoint of the controlled DDIM map under the task-specific detach conventions, including all non-detached cross-agent derivatives induced by \(\varphi\) and the controls.
Note that in our implementation, the matched-transition set \(\mathcal K\) is resampled at each optimisation iteration, while the adjoint recursion is always evaluated over all reverse transitions.
\input{_algo}
\noindent
Note that in \Cref{alg:cmds-discrete-am}, we report a generic context builder \(
    \mathcal C_\kappa:
    (
        X_\ell,\,
        \hat X_{1\mid\ell},\,
        Y_\ell,\,
        \hat Y_{1\mid\ell},\,
        \zeta_\ell
    )
    \mapsto
    \mathcal I_\ell
\), which is experiment-specific.
Thus, we use the basic, full-adjoint form of adjoint matching, discretised directly through the DDIM transitions.
Unlike the lean-adjoint variant of \citet{domingoenrich2025adjoint}, the backward replay differentiates the controlled transition and the quadratic control cost.

%% file: _algo.tex
\begin{algorithm}[t]
\caption{Adjoint matching for \method{} (discrete time)}
\label{alg:cmds-discrete-am}
\footnotesize
\begin{algorithmic}[1]

\Require frozen component denoisers
\(\epsilon_{\mathrm b}^i\), \(i=1,\ldots,N\);
parametrised component-wise residual
\(\delta\epsilon_\theta\);
assembly map \(\varphi\);
task-dependent context builder \(\mathcal C_\kappa\);
terminal reward \(r_\kappa\);
reward strength \(\lambda\);
reverse-time grid \(\zeta_0>\cdots>\zeta_{K-1}\);
joint controlled reverse maps \(F_\ell^\theta\),
signed epsilon-displacement coefficients \(\gamma_\ell\),
and stochastic scales \(\varsigma_\ell>0\);
matched transitions \(\mathcal K\subseteq\{0,\ldots,K-1\}\);
batch size \(B\)

\State Sample task conditions
\(\kappa_1,\ldots,\kappa_B\)
\bluecomment{e.g.,\ target \(d_{\Delta x}\), observation \(b\), or starts/goals \(\{(s_i,g_i)\}_{i=1}^N\)}

\State Sample
\(X_0^{i,(b)}\sim\mathcal N(0,I)\)
and independent rollout noise
\(\Xi_\ell^{i,(b)}\sim\mathcal N(0,I)\)

\State
\(\bar\theta\gets\operatorname{stopgrad}(\theta)\)

\algphase{1. Controlled rollout without gradient tracking}

\For{\(\ell=0,\ldots,K-1\)}

    \For{\(i=1,\ldots,N\)}

        \State
        \(
        \epsilon_{\mathrm b,\ell}^i
        \gets
        \epsilon_{\mathrm b}^i
        (X_\ell^i,\zeta_\ell)
        \)

        \State
        Compute
        \(
        \hat X_{1\mid\ell}^i
        \)
        \bluecomment{Tweedie:
        \(
        \hat X_{1\mid\ell}^i
        =
        \frac{
            X_\ell^i
            -
            \sqrt{1-\bar\alpha_{\zeta_\ell}}\,
            \epsilon_{\mathrm b,\ell}^i
        }{
            \sqrt{\bar\alpha_{\zeta_\ell}}
        }
        \)}

    \EndFor

    \State
    \(
    \hat X_{1\mid\ell}
    \gets
    \left(
        \hat X_{1\mid\ell}^1,
        \ldots,
        \hat X_{1\mid\ell}^N
    \right)
    \)

    \State
    \(
    Y_\ell
    \gets
    \varphi(X_\ell),
    \qquad
    \hat Y_{1\mid\ell}
    \gets
    \varphi(\hat X_{1\mid\ell})
    \)

    \State
    \(
    \mathcal I_\ell
    \gets
    \mathcal C_{\kappa}
    \left(
        X_\ell,\,
        \hat X_{1\mid\ell},\,
        Y_\ell,\,
        \hat Y_{1\mid\ell},\,
        \zeta_\ell
    \right)
    \)
    \bluecomment{The control for agent \(i\) depends on the full multi-agent state}

    \State
    \(
    \left(
        \delta\epsilon_{\bar\theta,\ell}^1,
        \ldots,
        \delta\epsilon_{\bar\theta,\ell}^N
    \right)
    \gets
    \delta\epsilon_{\bar\theta}(\mathcal I_\ell)
    \)

    \For{\(i=1,\ldots,N\)}

        \State
        \(
        w_{\bar\theta,\ell}^i
        \gets
        \dfrac{\gamma_\ell}{\varsigma_\ell}
        \delta\epsilon_{\bar\theta,\ell}^i
        \)

        \State
        \(
        X_{\ell+1}^i
        \gets
        F_\ell
        \left(
            X_\ell^i,\,
            \epsilon_{\mathrm b,\ell}^i
            +
            \delta\epsilon_{\bar\theta,\ell}^i,\,
            \Xi_\ell^i
        \right)
        \)
        \bluecomment{Controlled DDIM rollout; see \cref{eq:cmds-ddim-denoiser,eq:cmds-ddim-transition,eq:cmds-ddim-control}}

    \EndFor

    \State Store \(X_\ell\)

\EndFor

\Statex

\algphase{2. Backward discrete-adjoint replay}

\State
\(
a_K
\gets
\nabla_{X_K}
\left[
    -\lambda
    \sum_{b=1}^{B}
    r_{\kappa_b}
    \left(
        \varphi(X_K^{(b)})
    \right)
\right]
\)
\bluecomment{Terminal costate}

\For{\(\ell=K-1,\ldots,0\)}

    \State
    \(
    \tilde a_{\ell+1}
    \gets
    \operatorname{stopgrad}(a_{\ell+1})
    \)
    \bluecomment{Destination costate}

    \State
    Re-evaluate
    \(F_\ell^{\bar\theta}(x,0)\)
    and \(w_{\bar\theta,\ell}(x)\)
    at \(x=\operatorname{stopgrad}(X_\ell)\),
    using the task-specific detach conventions
    \bluecomment{Differentiate all non-detached state routes; keep \(\bar\theta\) fixed}

    \State
    \(
    \displaystyle
    a_\ell
    \gets
    \operatorname{stopgrad}
    \left[
        \tilde a_{\ell+1}
        +
        \left.
        \nabla_x
        \left(
            \left\langle
                \tilde a_{\ell+1},
                F_\ell^{\bar\theta}(x,0)-x
            \right\rangle
            +
            \dfrac12
            \sum_{b=1}^{B}
            \sum_{i=1}^{N}
            \left\|
                w_{\bar\theta,\ell}^{i,(b)}(x)
            \right\|^2
        \right)
        \right|_{x=X_\ell}
    \right].
    \)

\EndFor

\algphase{3. Control regression}

\For{\(\ell\in\mathcal K\)}

    \State Build
    \(
    \mathcal I_\ell
    =
    \mathcal C_\kappa
    \left(
        \operatorname{stopgrad}(X_\ell),
        \hat X_{1\mid\ell},
        Y_\ell,\,
        \hat Y_{1\mid\ell},\,
        \zeta_\ell
    \right)
    \)
    using domain-specific detach conventions

    \State
    \(
    \left(
        \delta\epsilon_{\theta,\ell}^1,
        \ldots,
        \delta\epsilon_{\theta,\ell}^N
    \right)
    \gets
    \delta\epsilon_\theta(\mathcal I_\ell)
    \)

    \For{\(i=1,\ldots,N\)}

        \State
        \(
        w_{\theta,\ell}^i
        \gets
        \dfrac{\gamma_\ell}{\varsigma_\ell}
        \delta\epsilon_{\theta,\ell}^i
        \)

        \State
        \(
        e_{\theta,\ell}^i
        \gets
        w_{\theta,\ell}^i
        +
        \varsigma_\ell
        \operatorname{stopgrad}
        \left(
            a_{\ell+1}^i
        \right)
        \)
        \bluecomment{Adjoint-matching regression residual}

    \EndFor

\EndFor

\State
\(
\displaystyle
\mathcal L_{\mathrm{AM}}(\theta)
\gets
\dfrac{1}{2B}
\sum_{b=1}^{B}
\sum_{\ell\in\mathcal K}
\sum_{i=1}^{N}
\left\|
    e_{\theta,\ell}^{i,(b)}
\right\|^2
\)
\bluecomment{PointMaze uses the self-calibrated pseudo-Huber variant; see \Cref{apdx:maze-robust-loss}}

\State
\(
\theta
\gets
\Call{OptimizerStep}{
    \theta,\,
    \nabla_\theta\mathcal L_{\mathrm{AM}}(\theta)
}
\)

\end{algorithmic}
\end{algorithm}

%% file: _exp_apdx.tex
\section{Experiments}
\subsection{Two-Agent OU Coordination: A Stochastic-Control Sanity Check}
\label{apdx:ou_coordination}
\begin{figure}
    \centering
    \includegraphics[width=0.95\textwidth]{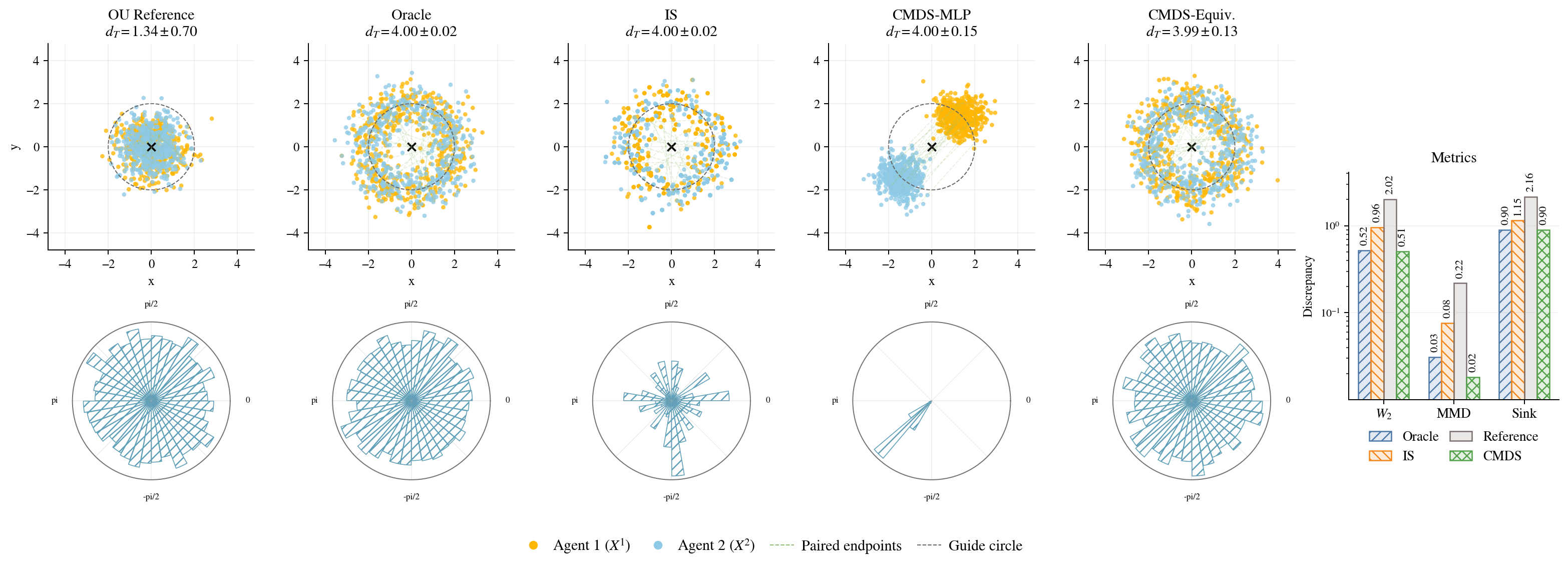}
    \caption{\textbf{Two-agent SOC toy: discrete adjoint.}
    Two agents start from an OU reference and are rewarded for reaching terminal configurations with a prescribed separation.
    We compare the (uncontrolled) OU reference, a value-function oracle, an importance-sampled reference (IS), and the learned control.
    }
    \label{fig:soc-toy-diffopt}
\end{figure}

\begin{figure}
    \centering
    \includegraphics[width=0.95\textwidth]{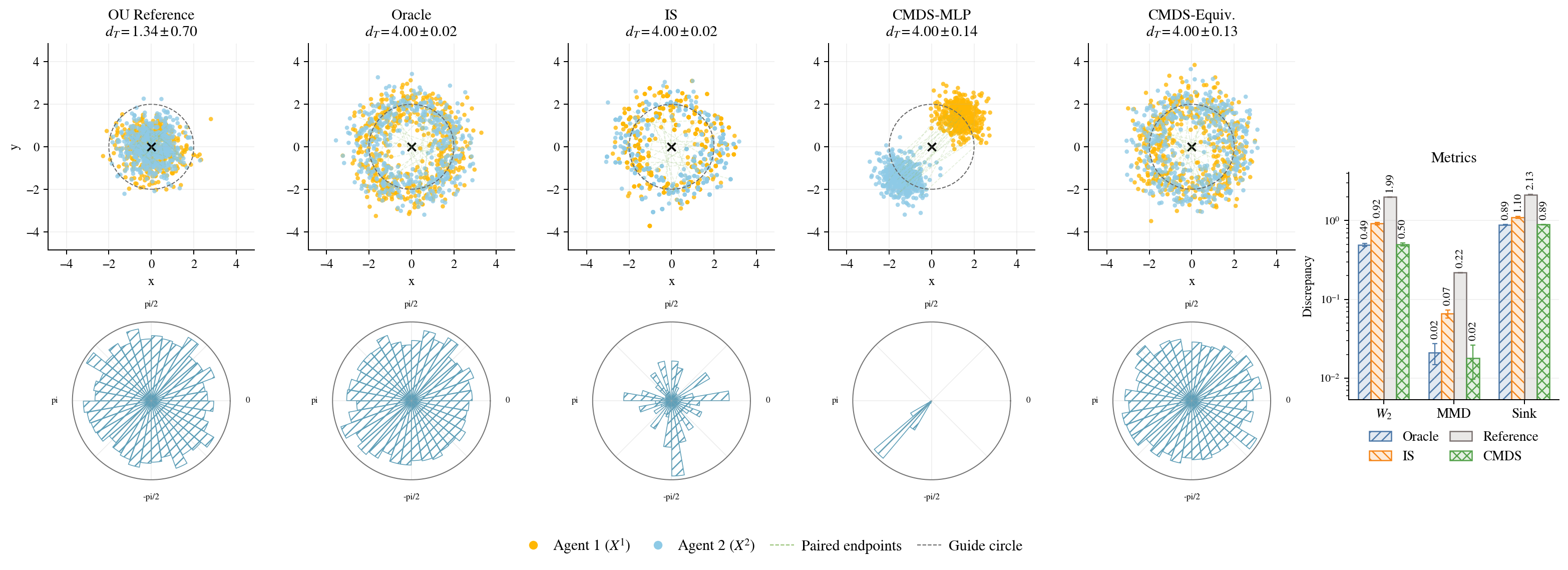}
    \caption{\textbf{Two-agent SOC toy: adjoint matching.}
    Two agents start from an OU reference and are rewarded for reaching terminal configurations with a prescribed separation.
    We compare the (uncontrolled) OU reference, a value-function oracle, an importance-sampled reference (IS), and the learned control.
    }
    \label{fig:soc-toy-adjoint}
\end{figure}

\begin{takehome}
\textbf{Key Takeaways:}
\noindent
\textbf{Least-action coordination recovers the target law.}
Across \cref{fig:soc-toy-diffopt,fig:soc-toy-adjoint}, we compare the learned terminal distribution against the stochastic-control oracle using the \(2\)-Wasserstein distance (\(W_2\)), maximum mean discrepancy (MMD), and Sinkhorn transport cost (Sink.).
Together with the terminal-separation and angular diagnostics, these metrics show whether the control recovers the full reward-tilted distribution, rather than merely satisfying the prescribed distance constraint.
\noindent
\textbf{Relational structure matters.}
A rotation-equivariant control preserves the symmetry of the target law, whereas an unconstrained MLP can satisfy the distance constraint while collapsing onto a preferred direction.
\noindent
\textbf{Constraint satisfaction alone may not be sufficient.}
A control may achieve the correct terminal separation while remaining far from the oracle distribution.
In particular, the angular histograms expose symmetry collapse that is not visible from the distance constraint alone, while \(W_2\), MMD, and Sinkhorn transport cost quantify this distributional mismatch.
\noindent
\textbf{The experiment isolates the control mechanism.}
The reference dynamics are known and an exact reduced value-function oracle is available, providing a controlled sanity check that the coordination objective learns the intended low-action path-space perturbation. 
\end{takehome}

To isolate the stochastic-control mechanism underlying our coordination framework, we consider two agents, each evolving in \(\mathbb R^2\) according to an Ornstein--Uhlenbeck (OU) reference process.
This experiment is a low-dimensional sanity check in which the reference dynamics are known and a high-accuracy value-function oracle can be computed in a reduced coordinate.
We use ordinary forward control time \(t\in[0,1]\) in this experiment, and therefore omit the reverse-time hats used for diffusion sampling elsewhere in the paper.
Let
\(
    X_t=(X^1_t,X^2_t)\in\mathbb R^4
\)
denote the joint state.
The controlled dynamics are:
\[
    \dd X^i_t
    =
    \left[
        -X^i_t
        +
        U^i_\theta(X_t,t)
    \right]\dd t
    +
    \dd W^i_t,
    \qquad
    i\in\{1,2\},
\]
where \(W^1\) and \(W^2\) are independent two-dimensional Brownian motions and the uncontrolled drift \(-X^i_t\) pulls each agent toward the origin.
The two agents are coupled only through the terminal relational cost,
\[
    g(X_1)
    = \beta
    \left(
        \|X^2_1-X^1_1\|-d_T
    \right)^2,
\]
which encourages their terminal separation to match a prescribed distance \(d_{T}\).
The corresponding stochastic optimal-control objective is
\[
    J(\theta)
    =
    \mathbb E_{\mathbb P^\theta}
    \left[
        g(X_1)
        +
        \frac{\alpha}{2}
        \int_0^1
        \sum_{i=1}^{2}
        \left\|
            U^i_\theta(X_t,t)
        \right\|^2
        \dd t
    \right],
\]
where \(\mathbb P^\theta\) denotes the path law induced by the controlled dynamics, \(\alpha>0\) controls the path-action penalty, and \(\beta>0\) controls the strength of the terminal constraint.
In the uncontrolled process, both agents concentrate near the origin, and their terminal separation is typically substantially smaller than \(d_T\).
A successful control should therefore separate the agents sufficiently to satisfy the relational constraint while avoiding unnecessary translation of their centre of mass.

\paragraph{Reduced value-function oracle.}
Since the two agents have the same component-wise OU dynamics, their common and relative modes can be decoupled. 
Since the terminal cost only depends on the relative displacement and their quadratic action is minimised with zero common mode control (i.e., $q_t = \dfrac{u^1_t+u^2_t}{2}$), the control problem can be reduced to the relative coordinate
\[
    R_t
    \coloneqq
    X^2_t-X^1_t
    \in\mathbb R^2.
\]
For controls \(u^1_t,u^2_t\), define the relative control
\(
    v_t
    \coloneqq
    u^2_t-u^1_t.
\)
For any fixed \(v_t\), the minimum-action decomposition across the two agents is
\[
    u^1_t=-\frac12v_t,
    \qquad
    u^2_t=\frac12v_t,
\]
and hence
\[
    \frac{\alpha}{2}
    \left(
        \|u^1_t\|^2+\|u^2_t\|^2
    \right)
    =
    \frac{\alpha}{4}\|v_t\|^2.
\]
The relative controlled process therefore satisfies
\[
    \dd R_t
    =
    \left(
        -R_t+v_t
    \right)\dd t
    +
    \sqrt{2}\,\dd B_t,
\]
where
\(
    B_t=(W^2_t-W^1_t)/\sqrt{2}
\)
is a standard two-dimensional Brownian motion.
Writing
\[
    g_R(r)
    \coloneqq
    \beta
    \left(
        \|r\|-d_T
    \right)^2,
\]
the reduced value function is
\[
    V(t,r)
    \coloneqq
    \inf_v
    \mathbb E
    \left[
        g_R(R_1)
        +
        \frac{\alpha}{4}
        \int_t^1
        \|v_s\|^2
        \dd s
        \,\middle|\,
        R_t=r
    \right].
\]
Its Hamilton--Jacobi--Bellman equation is
\[
    \partial_t V
    -
    r^\top\nabla_r V
    +
    \Delta_r V
    -
    \frac{1}{\alpha}
    \|\nabla_r V\|^2
    =
    0,
    \qquad
    V(1,r)=g_R(r).
\]
Introducing the desirability function
\[
    \psi(t,r)
    \coloneqq
    \exp\!\left(
        -\frac{1}{\alpha}V(t,r)
    \right),
\]
linearises the HJB equation.
Equivalently, by the Feynman--Kac representation,
\[
    \psi(t,r)
    =
    \mathbb E
    \left[
        \exp
        \left(
            -\frac{1}{\alpha}
            g_R(R_1)
        \right)
        \,\middle|\,
        R_t=r
    \right]
    =
    \mathbb E
    \left[
        \exp
        \left(
            -\frac{\beta}{\alpha}
            \left(
                \|R_1\|-d_T
            \right)^2
        \right)
        \,\middle|\,
        R_t=r
    \right],
\]
where $\beta/\alpha$ matches the $\lambda$ in the SOC formulation, and
\(
    V(t,r)
    =
    -\alpha\log\psi(t,r).
\)
Under the uncontrolled OU reference process,
\[
    \dd R_t
    =
    -R_t\,\dd t
    +
    \sqrt{2}\,\dd B_t,
\]
so that
\[
    R_1
    \mid
    R_t=r
    \sim
    \mathcal N
    \left(
        e^{-(1-t)}r,
        \left(
            1-e^{-2(1-t)}
        \right)I_2
    \right).
\]
Since the terminal cost depends only on \(\|R_1\|\), the value function is radial, and the conditional distribution of \(\|R_1\|\) is Rice distributed.
Consequently, \(\psi(t,r)\) can be evaluated accurately by one-dimensional quadrature over the terminal radius.
Minimising the reduced Hamiltonian gives
\[
    v^\star(t,r)
    =
    -\frac{2}{\alpha}
    \nabla_r V(t,r)
    =
    2\nabla_r\log\psi(t,r),
\]
with the corresponding minimum-action split
\[
    u^{1,\star}_t
    =
    -\frac12v^\star(t,R_t),
    \qquad
    u^{2,\star}_t
    =
    \frac12v^\star(t,R_t).
\]
This defines the radial stochastic-control oracle used in \cref{fig:soc-toy-diffopt,fig:soc-toy-adjoint}.
The HJB construction above directly introduces the optimal change of path measure through $\psi(t, r) = \exp(-V(t, r)/\alpha)$, which solves the backward Kolmogorov equation of the uncontrolled relative OU process 
$$
\dd R_t = - R_t \dd t + \sqrt{2} \dd B_t,
$$
with terminal condition $\psi(1, r) = \exp(-g_R(r)/\alpha)$. Then, by It\^{o}'s formula, under the uncontrolled law, 
$$
    \dd \log \psi(t, R_t) = \sqrt{2} \nabla_r \log \psi(t, R_t) \cdot \dd B_t - \|\nabla_r \log \psi(t, R_t)\|^2 \dd t,
$$
and thus
$$
    \frac{\psi(t, R_t)}{\psi(0, R_0)} = \exp \left(\int_0^t \frac{ v^\star(s,R_s)}{\sqrt{2}} \cdot \dd B_s -\frac{1}{2} \int_0^t \bigg\|\frac{v^\star(s,R_s)}{\sqrt{2}}\bigg\|^2 \dd s \right).
$$
By Girsanov's theorem, the RHS of the above equation is exactly the Radon-Nikodym derivative of the optimally controlled path measure of $R_t$ with respect to its uncontrolled counterpart up to time $t$. Evaluating at $t=1$, it gives
$$
\frac{\dd \mathbb{P}_R^\star}{\dd \mathbb{P}_R}(R_\cdot) = \frac{\exp(-g_R(R_1)/\alpha)}{\psi(0, R_0)}.
$$
Since our above formulation has shown that controls are only applied to $R_t$ and the common mode is uncontrolled and since $g(X_1) = g_R(R_1)$ in this OU example, substituting these relations back to the two-agent process gives
\[
    \frac{
        \dd\mathbb P^\star
    }{
        \dd\mathbb P
    }
    (X_\cdot)
    = \frac{ \exp
    (- g(X_1)/\alpha
    )}{\psi(0, X^2_0 - X^1_0)}
   .
\]
The importance-sampling baseline approximates this same change of measure by drawing trajectories from the uncontrolled OU process and weighting each sample proportionally to
\(
    \dfrac{\exp(-g(X_1)/\alpha)}{\psi(0, R_0)}
\).

For the oracle samples, we numerically evaluate the Feynman--Kac representation of the reduced HJB solution using radial quadrature, and sample directly from its implied joint terminal marginal. 
This avoids the time-discretisation error of an Euler rollout of the optimal control.

\paragraph{Learning the control.}
We train the same feedback-control parameterisation using either direct discrete adjoint (a.k.a., differentiable optimisation of \(J(\theta)\) or differentiable simulations) or an adjoint-matching surrogate \citep{domingo2024taxonomy}.
For the latter, let
\(
    \bar\theta=\operatorname{stopgrad}(\theta)
\)
and sample a trajectory
\(
    X^{\bar\theta}\sim\mathbb P^{\bar\theta}
\)
without gradient tracking.
%
%
For this experiment, we train \method{} using the full discrete adjoint of the controlled Euler--Maruyama dynamics.
Let \(h=1/K\), \(t_\ell=\ell h\), and
\(X_\ell=X_{t_\ell}^{\bar\theta}\).
Define
\[
    F_\ell(x)
    =
    x+h\bigl[-x+U_{\bar\theta}(x,t_\ell)\bigr],
    \qquad
    c_\ell(x)
    =
    \frac{\alpha}{2}
    \left\|U_{\bar\theta}(x,t_\ell)\right\|^2.
\]
The joint adjoint satisfies
\[
    a_K^{\bar\theta}
    =
    \nabla_x g(X_K),
    \qquad
    a_\ell^{\bar\theta}
    =
    D_xF_\ell(X_\ell)^\top a_{\ell+1}^{\bar\theta}
    +
    h\,\nabla_x c_\ell(X_\ell),
    \quad \ell=K-1,\ldots,0.
\]
The parameters \(\bar\theta\) are held fixed during this backward recursion, while the state derivatives of the feedback control
and its quadratic running cost are retained.
The adjoint-matching update at step \(\ell\) uses \(-a_{\ell+1}^{\bar\theta}/\alpha\) as its control target, with both the sampled states and the computed adjoints detached during the matching update.
The resulting discrete adjoint-matching surrogate is
\[
    \mathcal L_{\mathrm{AM}}^{\mathrm{OU}}(\theta;\bar\theta)
    =
    h\,\mathbb E\!\left[
        \sum_{\ell=0}^{K-1}\sum_{i=1}^{2}
        \left\|
            U_\theta^i(X_\ell,t_\ell)
            + \frac{1}{\alpha}a_{\ell+1}^{i,\bar\theta}
        \right\|^2
    \right].
\]
The expectation is over frozen Euler--Maruyama rollouts.
Both the sampled states and the computed adjoints are detached during the matching update.

\paragraph{Control parameterisation.}
We consider two feedback-control parameterisations.
The first is an unconstrained multilayer perceptron that receives the complete joint state and time,
\(
    U^i_\theta
    =
    U^i_\theta(X_t,t).
\)
The second explicitly encodes the rotational symmetry of the problem,
\[
U_\theta^i(X_t,t)
=
c_{\theta_i}\!\left(\|X_t^j-X_t^i\|^2,t\right)
(X_t^j-X_t^i),
\qquad j\ne i.
\]
The two scalar coefficient networks have separate parameters.
This equivariant parameterisation is a natural inductive bias because both the reference dynamics and the terminal cost are invariant to a common rotation of the two-agent system.
The optimal correction therefore acts along the relative displacement between the agents, pushing them apart when their predicted terminal separation is too small and pulling them together when it is too large.
The comparisons in \cref{fig:soc-toy-diffopt,fig:soc-toy-adjoint} highlight the importance of this inductive bias.
The uncontrolled OU process concentrates both agents near the origin and therefore produces terminal separations substantially below the target.
The value-function oracle achieves the prescribed separation while preserving rotational symmetry.
The importance-sampling baseline approximates the same terminal tilt but suffers from finite-sample variance.
The unconstrained MLP control can substantially reduce the terminal-distance error, but it may select a preferred direction in the plane and thereby collapse the terminal-angle distribution.
In contrast, the rotation-equivariant control achieves a terminal separation close to the target while maintaining an approximately uniform angular distribution, substantially closer to the oracle.
Thus, terminal-distance accuracy alone is insufficient: a control may satisfy the relational constraint while distorting symmetries and diversity of the target law.
This toy problem provides a controlled SOC sanity check in which the component dynamics are known, the global constraint is purely relational, and a reduced value-function oracle is available.
The experiment tests whether a learned control can balance satisfaction of a terminal system-level constraint against quadratic path action.
It therefore provides a low-dimensional analogue of the least-action principle underlying \method{}: instead of arbitrarily moving samples toward high reward, the control seeks the smallest path-space perturbation required to make independently evolving components satisfy a consistency criterion.

\paragraph{A remark on the IS baseline.}
Despite using \(2^{20}\) proposals per evaluation seed, conditionally normalised IS achieves a pre-resampling effective sample size (ESS) of only \(99.6\pm15.3\), approximately \(0.0095\%\) of the proposal budget.
The \(4{,}096\) terminal pairs obtained by multinomial resampling contain only \(343.4\pm20.6\) distinct proposals, indicating severe weight concentration.
Metrics are evaluated on \(1{,}024\) equally weighted terminal pairs per distribution, subsampled without replacement from the endpoint banks, retaining repeated IS endpoints.
Against the numerical SOC terminal oracle, IS yields \(0.917\pm0.035\), \(0.0656\pm0.0075\), and \(1.103\pm0.030\)
for \(W_2\), MMD, and Sinkhorn transport cost, respectively, versus \(0.497\pm0.022\), \(0.0180\pm0.0082\), and \(0.888\pm0.011\) for \method{}.
Reported uncertainties are sample standard deviations across five independent evaluation banks, with the trained controller held fixed.
Thus, at the reported proposal budget, conditionally normalised IS exhibits severe weight concentration, while \method{} provides a close approximation to the SOC target terminal law, with empirical joint-terminal discrepancies comparable to oracle self-comparisons at the reported sample size.

\subsection{Relational Visual Assembly}\label{apdx:square-assembly}
\begin{takehome}
\textbf{Key Takeaways:}
\noindent
\textbf{Compositionality helps.}
When the sampler is factorised according to the object structure, coordination is substantially easier than forcing a single process to learn the full constraint through control alone due to positive inductive bias of the decomposition.
\noindent
\textbf{Rewards are fragile.}
Reward-based coordination is only as reliable as the system-level signal used to define it. 
Poorly specified rewards can be over-optimised or hacked.
\noindent
\textbf{Least-work coordination.} 
The SOC objective favours assemblies that satisfy the global design while requiring the smallest deviation from the independent pretrained dynamics.
Thus \method{} is not only searching for valid component combinations; it is learning a low-action path to valid assemblies. 
\method{} is sampling high-reward assemblies while paying a quadratic control/action cost that penalises deviation from the pretrained reverse diffusion dynamics.
\end{takehome}
This experiment is not reported in the main paper, but serves as a controlled testbed in which the coordination network can be trained cheaply, allowing us to study the method in detail from both theoretical and computational perspectives.
In particular, it lets us isolate the effects of component factorisation, optimisation strategy, and reward design, and examine how different rewards trade off constraint satisfaction, diversity, and the amount of control required to steer the pretrained processes.
\paragraph{Setup.}
\begin{wraptable}[15]{r}{0.5\linewidth}
    \centering
    \setlength{\tabcolsep}{3pt}
    \renewcommand{\arraystretch}{1.06}
    \resizebox{\linewidth}{!}{%
    \begin{tabular}{llrrr}
    \toprule\toprule
    Method & Optim. scheme & ELBO $(\uparrow)$ & $\log Z_{\mathrm{IS}}$ & Valid $(\uparrow)$ \\
    \midrule
    
    \multirow{2}{*}{$N=1$ (1-square)}
    & \adjmatch & -1303.1 & -7.2 & 0.0\% \\
    & \discadj & -1309.5 & -7.2 & 0.0\% \\
    \midrule
    
    \multirow{2}{*}{$N=1$ (2-square)}
    & \adjmatch & -289.1 & -0.7 & 18.9\% \\
    & \discadj & -173.4 & -0.7 & 25.4\% \\
    \midrule
    
    \multirow{2}{*}{\method{}}
    & \cellcolor{pastelblue!45}\adjmatch
    & \cellcolor{pastelblue!45}-74.3
    & \cellcolor{pastelblue!45}-1.6
    & \cellcolor{pastelblue!45}\textbf{99.2\%} \\
    & \cellcolor{pastelblue!45}\discadj
    & \cellcolor{pastelblue!45}\textbf{-61.2}
    & \cellcolor{pastelblue!45}-1.6
    & \cellcolor{pastelblue!45}98.5\% \\
    \bottomrule
\end{tabular}%
}
\caption{\textbf{Classifier-only steering diagnostics.}
    We report $\log Z_{\mathrm{IS}}$ as an importance-sampling estimate of the log normalising constant of the reward-tilted target, providing a calibration diagnostic for the strength of the induced tilt.
}
\label{tab:square_classifier_only}
\end{wraptable}
We test two claims central to \method{}: whether a single amortised control can satisfy a family of relational constraints, and whether explicitly factorising the sampler by component simplifies relational steering.
We consider one-channel \(10\times10\) canvases with a white background and dark \(2\times2\) squares.
The target distribution consists of valid two-square canvases satisfying the prescribed constraints.
We pretrain a single diffusion model \(G\) on canvases containing one square, and instantiate two independent frozen copies at sampling time.
No samples from the valid two-square distribution are provided to \method{} during either pretraining or coordination.
Their controlled terminal states are combined through the fixed left/right assembly
\[
    \hat Y_1^\theta
    =
    \varphi(\hat X_1^{\theta,1},\hat X_1^{\theta,2})
    =
    M_L\odot\hat X_1^{\theta,1}
    +
    M_R\odot\hat X_1^{\theta,2},
\]
where \(M_L,M_R\in\{0,1\}^{10\times10}\) mask the left and right halves and satisfy \(M_L+M_R=1\).
Each component contributes only within its assigned half; its contribution outside that region is masked out.
The masks therefore define the components' regions of influence, but do not ensure that a complete square appears in each visible half or that the two squares satisfy the requested relative separation.
Independent sampling can consequently produce plausible single-square components but invalid assemblies.

\paragraph{Left-Right Steering.}
We first evaluate whether the agents can learn an additional drift on top of the pretrained dynamics that guides each square toward its prescribed region of interest.
For this initial experiment, we train a simple multilayer perceptron to classify whether a square lies in the target half of the domain.
The classifier-only assembly reward is
\[
    r_{\mathrm{clf}}(y)
    =
    \log\sigma(f_\psi(M_L\odot y))
    +
    \log\sigma(f_\psi(M_R\odot y)),
\]
where \(f_\psi\) denotes the trained classifier, \(\sigma(z)=(1+e^{-z})^{-1}\) is the logistic sigmoid, and each term scores square presence in its corresponding masked half.
\begin{figure}[t]
    \centering
    \includegraphics[width=0.95\linewidth]{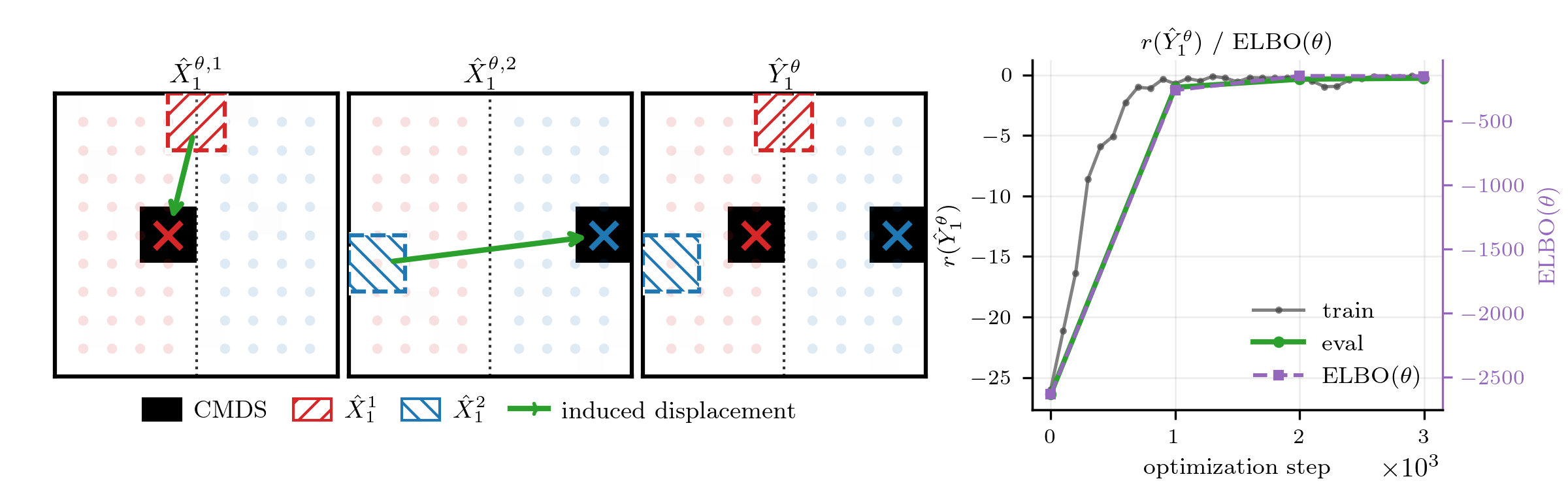}
    \vspace{-1.25em}
    \caption{\textbf{Left-right steering overview.}
    Two copies of a pretrained single-square diffusion generator are coordinated through a controlled reverse process and assembled by the fixed masking operator
    \(
        \hat Y^{\theta}_{1}
        =
        \varphi(\hat X^{\theta,1}_{1},\hat X^{\theta,2}_{1}).
    \)
    Black filled squares show the controlled terminal assembly \(\hat Y^{\theta}_{1}\). Red and blue dashed outlines show the paired counterfactual uncontrolled terminal proposals \(\hat X^{1}_{1}\) and \(\hat X^{2}_{1}\), obtained from the same initial noise draw but with the learned control removed. Thus, each dashed outline indicates where the corresponding generator copy would have ended up under the reference product law \(\hat{\mathbb P}\), before reward-based steering of the assembled canvas.
    The training curves report the terminal reward \(r(\hat Y^{\theta}_{1})\) and the path-space objective
    \(
        \mathrm{ELBO}(\theta)
        =
        \mathbb E_{\hat{\mathbb P}^{\theta}}
        [
        \lambda r(\hat Y^{\theta}_{1}) - A_{\theta}
        ],
    \)
    where \(A_{\theta}\) denotes the accumulated control/action cost. The learned controls steer the two locally valid component generators into their assigned left/right regions while remaining close to the pretrained dynamics.
    The path-action cost remains low because the uncontrolled generator already places substantial mass near valid square configurations, so the control only needs a moderate path-space perturbation to steer samples into high-reward assemblies.
    \label{fig:left_right_overview}
}
\end{figure}
\begin{figure}[t]
    \centering
    \includegraphics[width=0.95\linewidth, 
    trim=5 10 0 535,
    clip
    ]{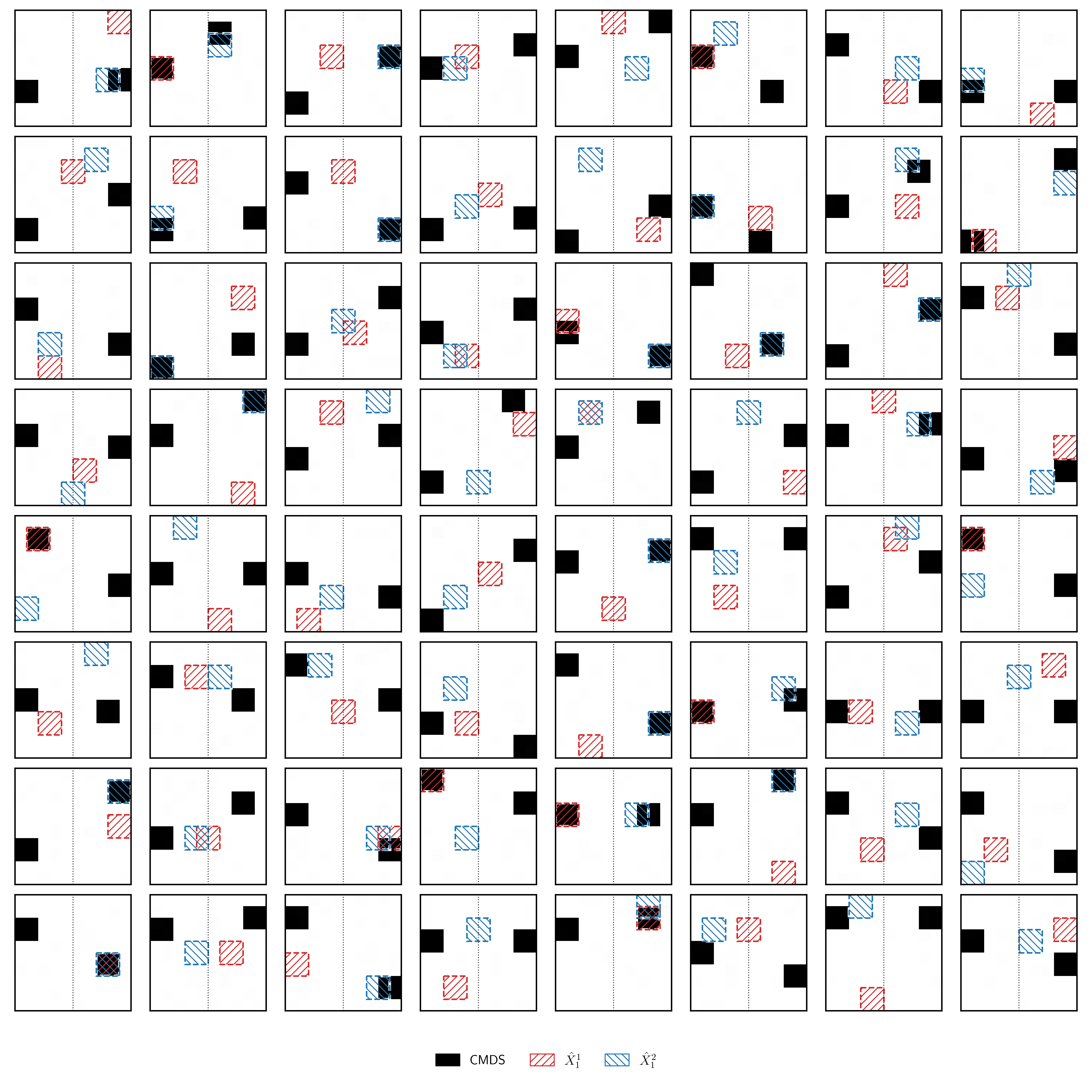}
    \caption{\textbf{Left-right steering (Samples).}
    Samples from the aggregated canvas after training.
    Black filled squares denote the \method{} assembled output. 
    Red and blue dashed outlines denote the paired uncontrolled terminal proposals from the two generator copies, obtained from the same initial/noise draw but with the learned control removed.
    Thus, for each sample, the outlines show where that instance would have landed without reward-based steering of the assembled canvas.
    The learned controls steer each square into its assigned half of the canvas, yielding valid stitched assemblies under the fixed left/right masking operator.
    }
    \label{fig:left_right_samples}
\end{figure}
In \cref{fig:left_right_samples}, we show samples of the aggregated canvas, illustrating that fine-tuning produces samples that satisfy the prescribed constraints.
In both cases, each component model generates a square in its assigned region, so that the final stitched assembly is valid.
The output of the first model is masked on the right half of the canvas, while the output of the second model is masked on the left half.
We optimise the controls using both the discrete adjoint method (\discadj) and adjoint matching (\adjmatch), as described in \cref{method:multi_agent_adjoint}.
We compare our approach against two single-process baselines: (i) tilting the same one-square base model used as the component generator in \method{}; and (ii) tilting a generator trained directly to sample two \(2\times 2\) squares within the \(10\times 10\) canvas.

\textbf{One-square baseline.} The failure of the one-square baseline is expected as its pretrained support is misaligned with the target object class. 
Since the base model represents canvases with a single \(2\times 2\) square, producing a valid two-square configuration requires the control to create (or hallucinate) an additional square through tilting alone, rather than merely steering existing component structure.

\textbf{Two-square baseline.} The two-square baseline has the correct object count, but not the right factorisation.
It can generate two-square canvases, yet the control must still solve the assignment problem of which square should satisfy which part of the left/right and distance reward (below).
This creates an ambiguous credit-assignment problem. 
The reward gradients act on the whole image rather than on separately addressable components.
In contrast, \method{} builds the assignment into the sampler by using one component process per square and applying the reward after assembly, so the same global signal induces more targeted, lower-action coordination.

\begin{figure}[htbp]
    \centering
    \includegraphics[width=0.95\linewidth]{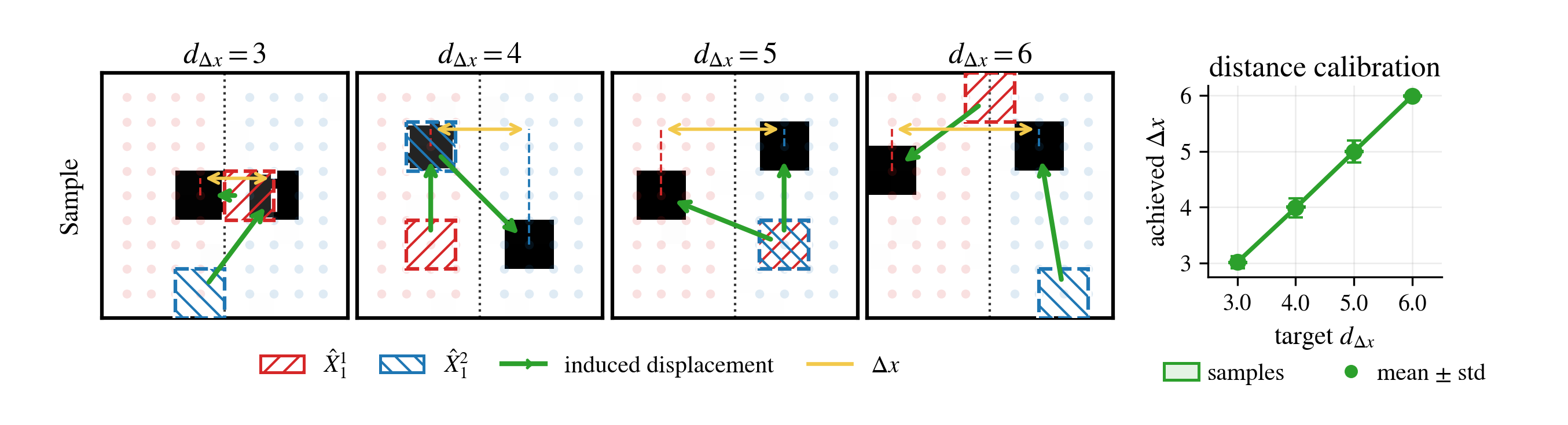}
    \caption{
        \textbf{Amortised assembly from copies of a single-square prior.}
        Two frozen copies of a generator pretrained only on one-square canvases are coordinated by a control and combined through a fixed left/right mask.
        Left: for each requested separation \(d_{\Delta x}\in\{3,4,5,6\}\), red and blue dashed outlines show paired uncontrolled terminal proposals obtained from the same noise, black squares show the controlled assembly, and green arrows show the induced corrections.
        Dots indicate feasible barycentre locations within the two masked halves.
        Right: the separation closely matches the requested target \(d_{\Delta x}\), with points and error bars reporting the sample mean and standard deviation.
        Yellow arrows indicate the achieved horizontal separation \(\Delta x\).
    }
    \label{fig:distance-amortisation}
\end{figure}

\textbf{Metrics.} 
In \cref{tab:square_classifier_only,tab:classifier_distance_diagnostics,tab:amortized_classifier_distance}, we report the path-space evidence lower bound (ELBO)
\[
    \mathrm{ELBO}(u)
    =
    \mathbb{E}_{\hat{\mathbb P}^{u}}\!\left[
    \lambda r\!\left(\varphi(\hat X^u_1)\right) - A_{u}
    \right],
\]
where \(\hat{\mathbb P}^{u}\) is the controlled path measure and \(A_u\) is the accumulated control path-action. The normalizing constant of the reward-tilted target,
\[
Z = \mathbb{E}_{\hat{\mathbb P}}\!\left[
\exp\!\left(\lambda r\!\left(\varphi(\hat X_1)\right)\right)
\right],
\]
is estimated independently from uncontrolled samples \(\hat X_1^{(j)}\) which follow the terminal time marginal of $\hat{\mathbb P}$ via
\[
    \log Z_{\mathrm{IS}}
    =
    \log\left(
    \frac{1}{M}\sum_{j=1}^M
    \exp\!\left(\lambda r\!\left(\varphi(\hat X_1^{(j)})\right)\right)
    \right).
\]
Thus, \(\log Z-\mathrm{ELBO}(u)\) is the path-space KL divergence from the controlled sampler to the reference path distribution re-weighted by the exponential of the scaled terminal reward and normalised by \(Z\).
This unconditional reward tilt differs from the fixed-initial-law SOC optimum unless the conditional normaliser \(Z(\hat X_0)\) is constant.
In the numerical diagnostic, \(\log Z\) is replaced by its estimate \(\log Z_{\mathrm{IS}}\).
Here $A_\theta$ denotes the quadratic path-action, i.e., the accumulated control cost along a parametrised controlled trajectory.
Following \cref{alg:cmds-discrete-am}, we write $X_\ell$ for the controlled rollout state, suppressing its parameter superscript, and $\zeta_\ell$ for the native diffusion time-step at transition $\ell$.
\[
    A_\theta
    =
    \frac{1}{2}
    \sum_{\ell=0}^{K-1}
    \sum_{i=1}^{N}
    \left\|
        w_{\theta,\ell}^{i}(X_\ell)
    \right\|_2^2,
\]
where $w_{\theta,\ell}^{i}$ is the normalised discrete control associated with component $i$ at DDIM transition $\ell$.
For multi-component samplers, the norm includes the sum over all controlled processes and pixels.
This is the discrete Girsanov control cost appearing in the path-space change of measure from the uncontrolled process \(\hat{\mathbb P}\) to the controlled process \(\hat{\mathbb P}^{u}\).
\paragraph{Coupled Steering.}
We impose a constraint that can only be satisfied through coordination.
\begin{wraptable}[20]{r}{0.58\linewidth}
\centering
\setlength{\tabcolsep}{3pt}
\renewcommand{\arraystretch}{1.06}
\resizebox{\linewidth}{!}{%
\begin{tabular}{llrrr}
\toprule\toprule
Method & Optim. scheme & ELBO $(\uparrow)$ & $\log Z_{\mathrm{IS}}$ & Valid $(\uparrow)$ \\
\midrule
\multirow{2}{*}{$N=1$ (1-square)}
& \adjmatch & -1706.8 & -22.8 & 0.0\% \\
& \discadj  & -1565.0 & -22.8 & 0.0\% \\
\midrule
\multirow{2}{*}{$N=1$ (2-square)}
& \adjmatch & -1561.7 & -4.5 & 8.3\% \\
& \discadj  & -381.4  & -4.5 & 9.6\% \\
\midrule
\multirow{2}{*}{\method{}}
& \cellcolor{pastelblue!45}\adjmatch
& \cellcolor{pastelblue!45}\textbf{-156.7}
& \cellcolor{pastelblue!45}-5.4
& \cellcolor{pastelblue!45}\textbf{97.9\%} \\
& \cellcolor{pastelblue!45}\discadj
& \cellcolor{pastelblue!45}-190.9
& \cellcolor{pastelblue!45}-5.4
& \cellcolor{pastelblue!45}96.8\% \\
\bottomrule
\end{tabular}%
}

\caption{\textbf{Coupled steering diagnostics.}
We report $\log Z_{\mathrm{IS}}$ as an importance-sampling estimate of the log normalising constant of the reward-tilted target, providing a diagnostic for the strength of the induced tilt.
Single-process baselines struggle to satisfy the coupled validity constraint, whereas \method{} coordinates two pretrained one-square generators through the assembled canvas and achieves near-perfect validity with improved ELBO.
}
\label{tab:classifier_distance_diagnostics}
\end{wraptable}
The assembly-level reward combines square presence with the requested horizontal separation:
\[
    r_{d_{\Delta x}}(y)
    =
    r_{\mathrm{clf}}(y)
    -
    \alpha\lvert\Delta_x(y)-d_{\Delta x}\rvert,
\]
where \(d_{\Delta x}\) is the target separation and \(\alpha\) is the distance penalty weight.
The achieved separation is
\[
    \Delta_x(y)
    =
    \mu_x(M_R\odot y)
    -
    \mu_x(M_L\odot y),
\]
with \(\mu_x\) denoting the horizontal foreground barycentre.
This introduces coordination through an analytic distance term computed between the barycentres of the two squares.
This term makes each agent's reward depend on the relative configuration of both squares, encouraging the learned drifts to coordinate rather than solve two independent localisation tasks.
\begin{table}
\centering
\setlength{\tabcolsep}{4pt}
\renewcommand{\arraystretch}{1.06}

\resizebox{0.75\linewidth}{!}{%
\begin{tabular}{lrrccccc}
\toprule\toprule
Training scheme
& ELBO$_{d_{\Delta x}=5}$ $(\uparrow)$
& $\left.\log Z_{\mathrm{IS}}\right|_{d_{\Delta x}=5}$
& \multicolumn{5}{c}{Valid $(\uparrow)$} \\
\cmidrule(lr){4-8}
& & & $d_{\Delta x}=3$ & $d_{\Delta x}=4$ & $d_{\Delta x}=5$ & $d_{\Delta x}=6$ & Avg. \\
\midrule
\cellcolor{pastelblue!25}\adjmatch
& \cellcolor{pastelblue!25}-172.2
& \cellcolor{pastelblue!25}-5.4
& \cellcolor{pastelblue!25}\textbf{95.8\%}
& \cellcolor{pastelblue!25}95.4\%
& \cellcolor{pastelblue!25}\textbf{97.3\%}
& \cellcolor{pastelblue!25}\textbf{97.1\%}
& \cellcolor{pastelblue!25}\textbf{96.4\%} \\
\cellcolor{pastelblue!25}\discadj
& \cellcolor{pastelblue!25}\textbf{-158.5}
& \cellcolor{pastelblue!25}-5.4
& \cellcolor{pastelblue!25}95.2\%
& \cellcolor{pastelblue!25}\textbf{95.7\%}
& \cellcolor{pastelblue!25}97.2\%
& \cellcolor{pastelblue!25}95.5\%
& \cellcolor{pastelblue!25}95.9\% \\
\bottomrule
\end{tabular}%
}
\vspace{.5em}
\caption{\textbf{Amortized steering diagnostics.}
    ELBO and $\log Z_{\mathrm{IS}}$ are evaluated at $d_{\Delta x}=5$. 
    Validity is evaluated at every supported target using 4096 samples per target with matched random seeds across target conditions.
    Avg. is the macro-average over $d_{\Delta x}\in\{3,4,5,6\}$.
}
\label{tab:amortized_classifier_distance}
\end{table}
In this setting, we also compare against the two single-process baselines.
The failure for the two-square baseline becomes more pronounced for the distance constraint, where validity depends on the relative barycentres of the two squares.
A unique control can improve the reward by moving mass in reward-favourable ways, but it does not inherit the component-level separation that makes the desired relation easy to express and control.
This failure is therefore an inductive-bias failure, not just a capacity failure. 
The two-square generator represents valid objects, but its representation does not expose the component assignments needed for least-action steering.
\begin{figure}[t]
    \centering
    \includegraphics[width=0.75\linewidth, 
    trim=5 10 0 0,
    clip
    ]{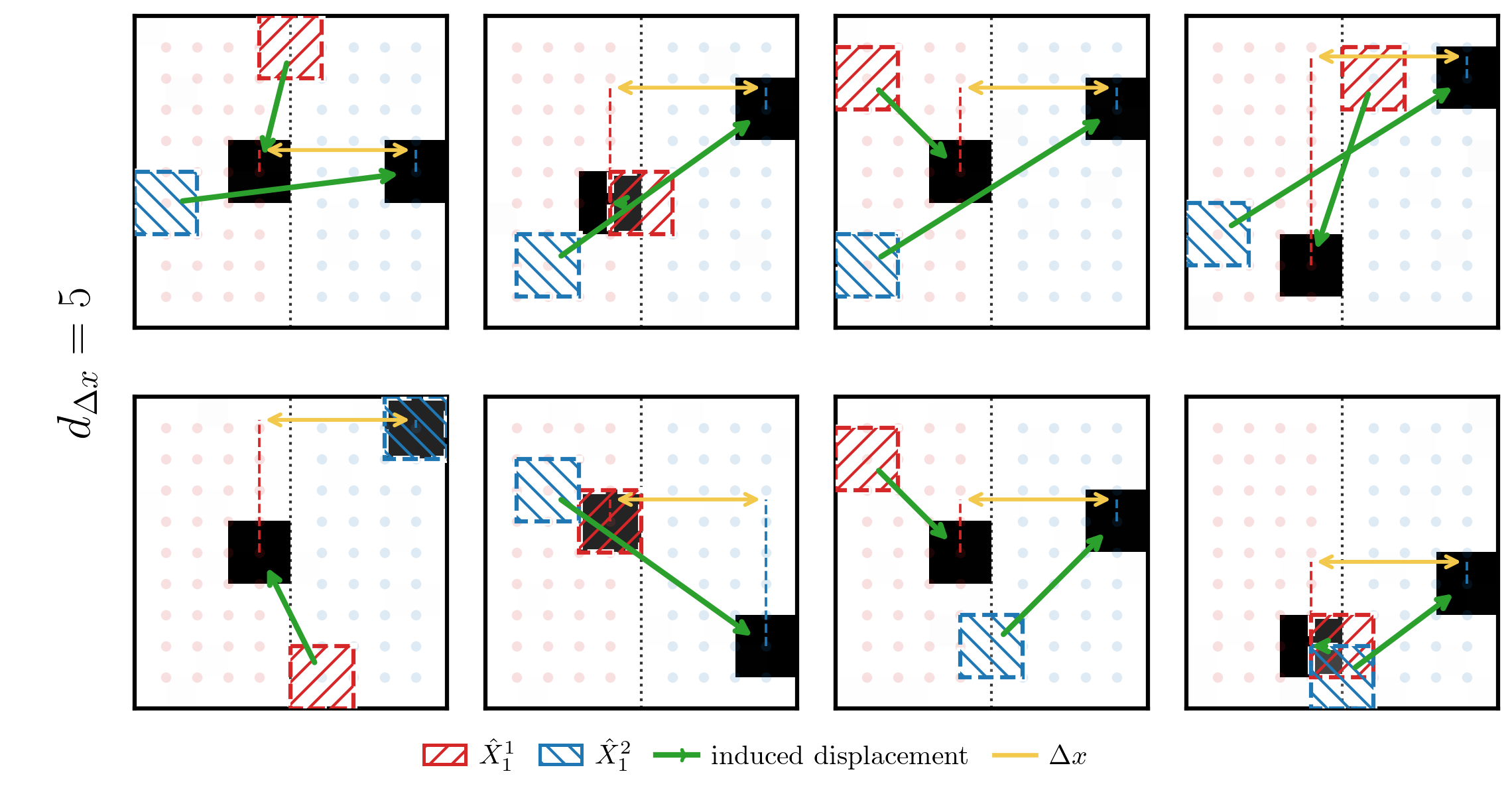}
    \caption{
    \textbf{Fixed-target coupled steering (\(d_{\Delta x}=5\)).}
    Black squares show the controlled assembly, red and blue hatched squares the matched reference terminal proposals, green arrows the induced corrections.
    Yellow arrows indicate the achieved horizontal separation \(\Delta x\).
    }
    \label{fig:x_distance_fix}
\end{figure}
\begin{figure}[t]
    \centering
    \includegraphics[width=0.85\linewidth, 
    trim=5 10 0 0,
    clip
    ]{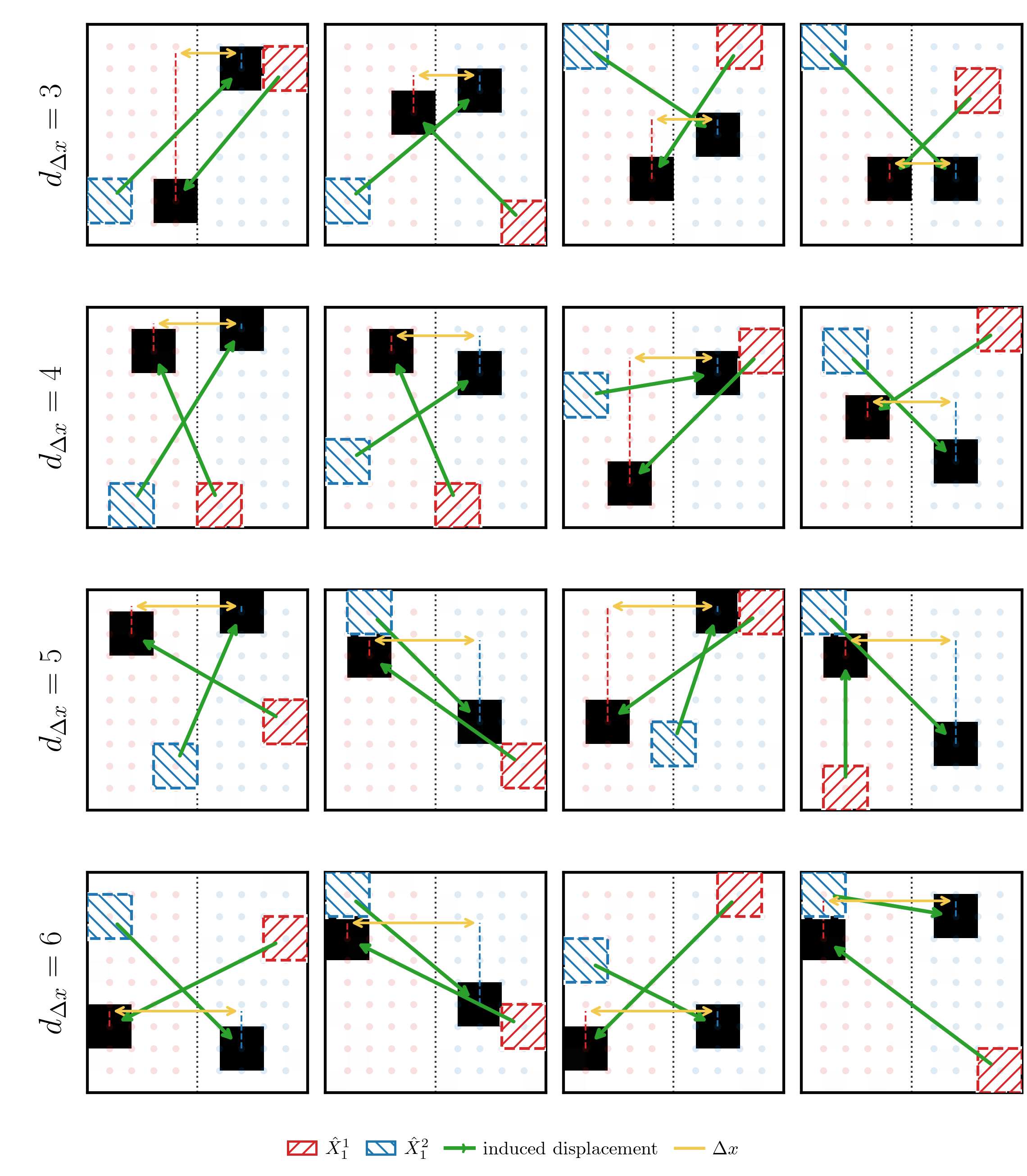}
    \caption{
    \textbf{Amortised coupled steering.}
    Samples from a single pair of agent-specific controls trained jointly over
    \(d_{\Delta x}\in\{3,4,5,6\}\).
    Black squares show the controlled assembly, red and blue hatched squares the matched reference terminal proposals, and green arrows the induced corrections.
    Yellow arrows indicate the achieved horizontal separation \(\Delta x\).
    }
    \label{fig:amortised_samples}
\end{figure}
\paragraph{Amortised coordination across design specifications.}
To test amortisation, we jointly train two agent-specific controllers with separate parameters, one acting on each frozen copy of the one-square reverse process, over \(d_{\Delta x}\in\{3,4,5,6\}\).
The requested separation is represented by a four-way one-hot vector, broadcast over the spatial grid and concatenated to each controller input at every DDIM transition.
Each controller outputs an additive residual to its corresponding frozen denoiser's noise prediction, thereby steering the two component processes jointly under the shared assembly-level reward.
As shown in \cref{fig:distance-amortisation}, the amortised control steers the matched reference terminal proposals into valid assemblies, while the achieved separations closely match the requested targets.
The discrete adjoint-matching formulation is provided in \Cref{apdx:ddim-adjoint-matching}, and the control implementation is described below.

\paragraph{Control implementation.}
For the squares experiments, we attach two agent-specific convolutional control heads to two frozen copies of the same pretrained one-square denoiser.
At each DDIM transition $\ell$, the frozen denoiser provides Tweedie estimates $\hat X_{1\mid\ell}^{\,i}$ of the terminal component states, which are assembled as
\(
    \hat Y_{1\mid\ell}
    =
    \varphi(
        \hat X_{1\mid\ell}^{\,1},
        \hat X_{1\mid\ell}^{\,2}
    ).
\)
The control head for component $i$ receives the full two-agent configuration,
\[
    \operatorname{cat}_c\!\left[
        X_\ell^i,\,
        \hat X_{1\mid\ell}^{\,i},\,
        Y_\ell,\,
        \hat Y_{1\mid\ell},\,
        X_\ell^j,\,
        \hat X_{1\mid\ell}^{\,j}
    \right],
    \qquad
    Y_\ell=\varphi(X_\ell^1,X_\ell^2),
    \qquad j\ne i,
\]
together with the embedding of the native diffusion timestep $\zeta_\ell$.
Coordination therefore enters through joint conditioning: although the two heads have separate parameters and produce component-specific corrections, each control depends on both reverse processes and on their current and predicted assembly.
Each head outputs an unbounded additive epsilon residual $\delta\epsilon_{\theta,\ell}^{\,i}$, which is added to the frozen denoiser
prediction and induces the normalised DDIM control $w_{\theta,\ell}^{i}$.
For the amortised variant, both heads are additionally conditioned on the requested separation $d_{\Delta x}$.
The Tweedie features are recomputed at every transition and remain differentiable with respect to the component states, while the  denoiser parameters remain frozen.

\subsection{Source Separation}\label{apdx:source-separation}
\begin{takehome}
\textbf{Key Takeaways:}
\noindent
\textbf{Joint control improves reconstruction while preserving component structure.}
In the evaluated circle--square source-separation task, \method{} achieves lower clean-mixture reconstruction error and more accurate labelled source placement than DPS, while retaining the structural coherence of the frozen component priors.
DPS has no equivalent control-cost regularisation and tends to enforce the measurement reward more aggressively, often at the expense of the individual components.
\noindent
\textbf{The control resolves component-specific corrections.}
For additive mixtures, the measurement-reward gradient is identical with respect to either clean source.
Providing this Tweedie-evaluated gradient to the joint controller is important: by combining it with both source states and their denoised estimates, the controller can turn shared measurement feedback into component-specific corrections.
\noindent
\textbf{Under ambiguity, plausible recovery need not equal ground-truth recovery.}
For sufficiently ill-posed problems, \method{} can recover \emph{a} plausible decomposition rather than necessarily \emph{the} ground-truth one.
Under stronger blur, components may preserve valid shapes while swapping their labelled locations and still explain the observation well; DPS tends to exhibit such swaps together with substantially greater component deformation.
\end{takehome}

Source separation seeks to recover two or more latent signals from a combined observation.
In many cases, datasets of individual source types and datasets of mixtures may each be available, while paired examples containing a mixture together with its constituent sources are unavailable or difficult to obtain.
We investigate whether \method{} can coordinate samplers trained independently on the individual source distributions to explain an observed mixture without requiring paired source--mixture training data.
The objective is to recover components whose assembly matches the observation while each component remains consistent with its respective pretrained distribution.

\paragraph{Task formalization.}
We consider \(N=2\) frozen pretrained samplers, \(G^1\) and \(G^2\), defining circle and square priors \(\tilde{\pi}^1\) and
\(\tilde{\pi}^2\), respectively.
The ground-truth component images are single-channel \(16\times16\) images, \(X_{\mathrm{gt}}^1,X_{\mathrm{gt}}^2
\in\{-1,+1\}^{1\times16\times16}\), with background value \(+1\) and foreground value \(-1\).
Their clean additive assembly is
\(
Y_{\mathrm{gt}}
=
\varphi(X_{\mathrm{gt}})
:=
X_{\mathrm{gt}}^1+X_{\mathrm{gt}}^2.
\)

\begin{figure}[h]
\centering
\includegraphics[
    width=0.95\linewidth,
    trim=0 25 10 0,
    clip
]{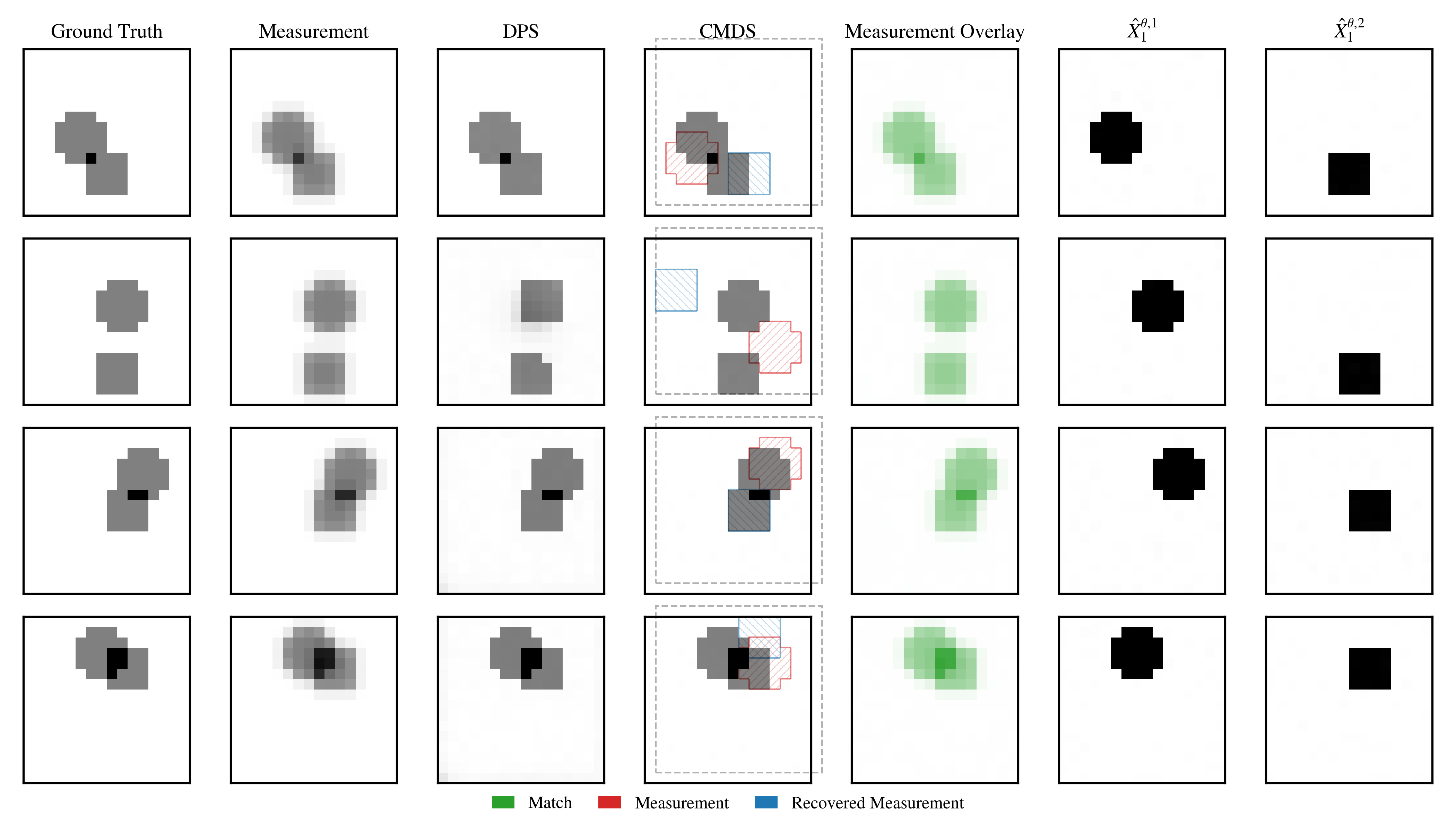}
\vspace{-.5em}
\caption{
\textbf{Representative samples (\(\tau=0.5\)).}
    Each row represents a different instance of the problem.
    The columns are: (i) the ground-truth mixture; (ii) the observation \(b\); (iii) the DPS mixture reconstruction; (iv) the \method{} mixture reconstruction, with the matched reference circle and square shown using red and blue hatching, respectively; (v) the blurred \method{} reconstruction overlaid with the observation \(b\); and (vi--vii) the \method{} estimates of the circle and square sources, respectively.
    In the measurement overlay, green indicates agreement between the predicted and observed mixtures, red indicates structure present only in the observation, and blue indicates structure present only in the predicted mixture.
}
\label{fig:source-separation-samples-0p5}
\vspace{-.5em}
\end{figure}

\begin{figure}[h]
\centering
\includegraphics[
    width=0.95\linewidth,
    trim=0 25 10 0,
    clip
]{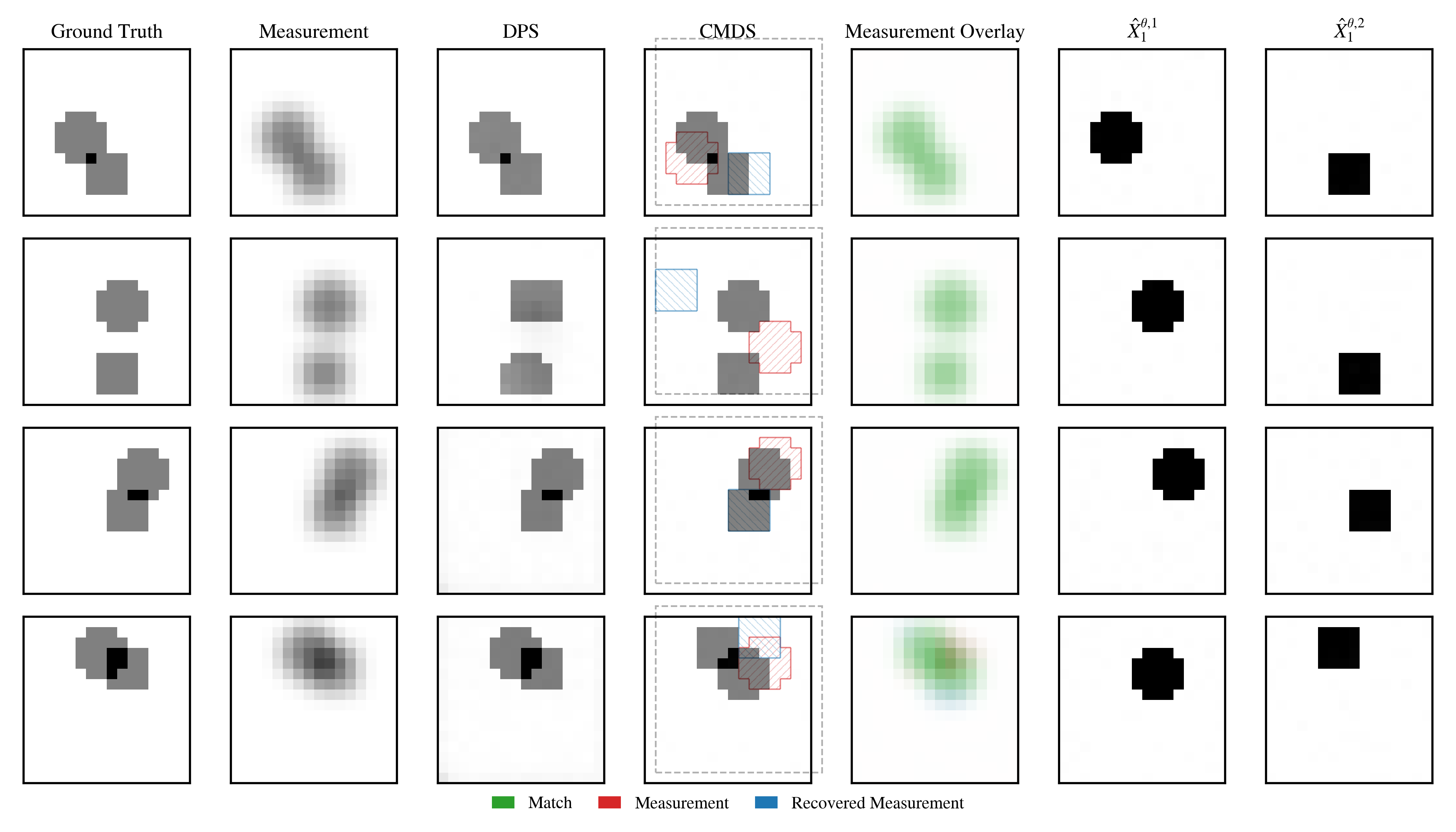}
\caption{
    \textbf{Representative samples (\(\tau=1.0\)).}
    Each row represents a different instance of the problem.
    The columns are: (i) the ground-truth mixture; (ii) the observation \(b\); (iii) the DPS mixture reconstruction; (iv) the \method{} mixture reconstruction, with the matched reference circle and square shown using red and blue hatching, respectively; (v) the blurred \method{} reconstruction overlaid with the observation \(b\); and (vi--vii) the \method{} estimates of the circle and square sources, respectively.
    In the measurement overlay, green indicates agreement between the predicted and observed mixtures, red indicates structure present only in the observation, and blue indicates structure present only in the predicted mixture.
}
\label{fig:source-separation-samples-1}
\end{figure}

\paragraph{Observation model.}
We observe
\(
    b=A_\tau(Y_{\mathrm{gt}}),
\)
where \({\rm A}_{\tau}\) denotes Gaussian convolution with a \(5\times5\) kernel and blur standard deviation
\(
    \tau\in\{0.5,1.0\}.
\)
No additional observation noise is applied (\(\varepsilon=0\)).
Although the ground-truth sources are binary-valued, the blurred observations and generated sources are continuous-valued.
Representative reconstructions under the two blur operators are shown in \cref{fig:source-separation-samples-0p5,fig:source-separation-samples-1}.

\paragraph{Component priors.}
The circle and square priors are trained independently using \(4096\) sampled examples per component and epoch, defining the reference product law
\(
    \rho=\tilde{\pi}^1\otimes\tilde{\pi}^2.
\)
Each noise-prediction network projects a \(32\)-dimensional sinusoidal time-step embedding and concatenates it spatially with the noisy image.
The resulting tensor is processed by two \(3\times3\), \(64\)-channel convolutional layers with SiLU activations, a residual four-head spatial self-attention layer, and two further \(3\times3\) convolutional layers producing a single-channel epsilon prediction.
Both samplers use a linear \(100\)-step diffusion schedule,
\(
    \beta_1=10^{-4},
    \beta_{100}=0.08.
\)
They are trained for \(1000\) epochs using Adam, batch size \(128\), and a learning rate cosine-annealed from \(10^{-3}\) to \(10^{-5}\), and are frozen.

\paragraph{Jointly controlled source processes.}
The reference process samples the circle and square independently.
\method{} coordinates their reverse processes using a joint feedback
control, producing
\[
    \hat X_1^\theta
    =
    \left(
    \hat X_1^{\theta,1},
    \hat X_1^{\theta,2}
    \right),
    \qquad
    \hat Y_1^\theta
    =
    \varphi(\hat X_1^\theta)
    =
    \hat X_1^{\theta,1}+\hat X_1^{\theta,2}.
\]
At each reverse step, the control outputs a separate additive epsilon-space control for each component process.

\paragraph{\method{} training.}
Following \cref{alg:cmds-discrete-am}, we use $\ell$ for rollout indices and $\zeta_\ell$ for the corresponding native diffusion timesteps.
At reverse transition $\ell$, the control receives the current noisy states $X_\ell^1,X_\ell^2$, their Tweedie estimates $\hat X_{1\mid\ell}^{\,1},\hat X_{1\mid\ell}^{\,2}$, the observation $b$, and the predicted observation
\[
    \hat b_\ell
    =
    A_\tau\!\left(
        \hat X_{1\mid\ell}^{\,1}
        +
        \hat X_{1\mid\ell}^{\,2}
    \right).
\]
It additionally receives the residual $b-\hat b_\ell$, the component-wise reward gradients, i.e., $\nabla_{\hat X_{1\mid\ell}^{\,1}}r_b$ and $\nabla_{\hat X_{1\mid\ell}^{\,2}}r_b$, and a sinusoidal embedding of $\zeta_\ell$.
It outputs two additive epsilon-space controls, one for each reverse process.
The control uses joint multi-scale processing with hidden width \(32\) and one output head per component.
We train the control by adjoint matching for \(10{,}000\) optimisation steps.
Each iteration samples \(512\) circle--square pairs, initial diffusion states, and rollout noise.
Rollouts use \(20\) DDIM transitions with \(\eta=1\), and adjoint matching uses \(15\) selected matching transitions.
We use Adam with cosine learning-rate decay from \(10^{-4}\) to \(10^{-6}\), gradient clipping at norm \(1.0\), and rejection of non-finite updates.
An exponential moving average of the control parameters is maintained with decay \(0.995\).

\paragraph{Reward and control cost.}
The system-level reward is
\[
    r_b(y)
    =
    -\frac{1}{d}
    \left\|
    {\rm A}_{\tau}(y)-b
    \right\|_2^2,
    \qquad
    d=16^2.
\]
Training uses reward multiplier \(1000\) and the uniform quadratic control cost from the \method{} objective.
The component-wise reward-gradient inputs are computed exactly through the known observation operator \({\rm A}_{\tau}\).

\paragraph{Source-separation control implementation.}
For source separation, we use a centralised joint control with a shared representation and two source-specific output heads, one for each frozen
component diffusion model.
At each DDIM transition \(\ell\), the shared control receives the current noisy states \(X_\ell^1,X_\ell^2\), their Tweedie estimates \(\hat X_{1\mid \ell}^1,\hat X_{1\mid \ell}^2\), the observation \(b\), and the predicted observation
\[
    \hat b_\ell    =
    {\rm A}_{\tau}
    \left(
    \hat X_{1\mid \ell}^1+\hat X_{1\mid \ell}^2
    \right).
\]
It is additionally conditioned on the measurement residual \(b-\hat b_\ell\), the component-wise reward gradients, and the diffusion-time embedding \(\zeta_\ell\).
Coordination arises through the shared representation as both output heads have access to both reverse processes, their predicted clean sources, and their joint measurement mismatch before producing source-specific corrections.
The two heads output additive epsilon residuals \(\delta\epsilon_{\theta,\ell}^1\) and \(\delta\epsilon_{\theta,\ell}^2\), which modify their respective frozen denoisers and induce the normalised DDIM controls \(w_{\theta,\ell}^1\) and \(w_{\theta,\ell}^2\).
The reward-gradient features are detached, while the Tweedie features remain differentiable.

\paragraph{Baselines.}
The reference baseline samples the frozen component priors independently,
\[
    \hat X_1^1\sim\tilde{\pi}^1,
    \qquad
    \hat X_1^2\sim\tilde{\pi}^2,
\]
and returns their assembly
\[
    \hat Y_1=\hat X_1^1+\hat X_1^2
\]
without conditioning on \(b\).
We additionally adapt diffusion posterior sampling (DPS) \citep{chung2023diffusion} to the product of the two frozen component priors.
At each reverse step, DPS evaluates the observation residual through the Tweedie estimates of both sources and applies the resulting component-wise data-fidelity gradients to their reverse updates.
Writing $\hat X_{1\mid\ell}^{\,i}$ for the Tweedie estimate of component $i$ at rollout index $\ell$, its guidance is computed from
\[
    \nabla_{X_\ell^i}
    \left\|
        {\rm A}_\tau\!\left(
            \hat X_{1\mid\ell}^{\,1}
            +
            \hat X_{1\mid\ell}^{\,2}
        \right)
        -b
    \right\|_2,
    \qquad i\in\{1,2\}.
\]
In this sense, product-DPS provides a non-amortised sampling-time analogue of coordination.
The observation-dependent correction is recomputed for every mixture, whereas \method{} amortises this dependence into a learned joint feedback control.

\paragraph{Evaluation protocol.}
All methods are evaluated on the same \(1024\) procedurally generated circle--square instances with batch size \(128\).
All methods are evaluated against the same ground-truth components \(X_{\mathrm{gt}}^1,X_{\mathrm{gt}}^2\) and use matched initial diffusion states; observation-conditioned methods use the same \(b\).
The \method{} and matched reference rollouts share identical per-step noise, while DPS uses the same rollout-noise seed under its selected DDIM discretisation.

\paragraph{Evaluation metrics.}
We report three complementary properties of the recovered components.
In the following, \(X_K^i\) denotes the recovered terminal component \(i\), with the method superscript suppressed, while \(X_{\mathrm{gt}}^i\) denotes the corresponding ground-truth component.
We write
\[
Y_K
=
\varphi(X_K)
=
X_K^1+X_K^2,
\qquad
Y_{\mathrm{gt}}
=
X_{\mathrm{gt}}^1+X_{\mathrm{gt}}^2.
\]
\textbf{Clean-mixture reconstruction RMSE} measures whether the recovered assembly matches the unblurred ground-truth mixture:
\[
\mathrm{RMSE}_{\mathrm{recon}}
=
\left[
\frac{1}{d}
\left\|
Y_K-Y_{\mathrm{gt}}
\right\|_2^2
\right]^{1/2}.
\]
This metric evaluates recovery of the assembled mixture but does not determine whether each recovered structure has been assigned to the correct source.
\textbf{Source IoU} compares the foreground mask of each recovered component with that of its corresponding labelled ground-truth component:
\[
\operatorname{IoU}_{\mathrm{source}}
=
\frac{1}{2}
\sum_{i=1}^{2}
\frac{
\left|M(X_K^i)\cap M(X_{\mathrm{gt}}^i)\right|
}{
\left|M(X_K^i)\cup M(X_{\mathrm{gt}}^i)\right|
}.
\]
The darkness representation and foreground mask are
\[
D(x)
=
\frac{1-\operatorname{clip}(x,-1,1)}{2},
\qquad
M(x)
=
\left\{p:D(x)_p\geq0.5\right\}.
\]
Equivalently, a pixel is classified as foreground when its clipped model value is non-positive.
Source IoU equals \(1\) when both recovered components occupy exactly the correct pixels, and \(0\) when neither overlaps its corresponding ground-truth component.
\textbf{Shape distortion} measures the difference between each recovered source and the nearest valid instance of its labelled prior shape, independently of location.
Let \(\mathcal P_1\) contain all \(100\) generator-valid circle placements and \(\mathcal P_2\) all \(169\) generator-valid square placements.
For each source, we define
\[
\delta_i
=
\min_{T\in\mathcal P_i}
\frac{
\left\|D(X_K^i)-T\right\|_2^2
}{
\left\|T\right\|_2^2
},
\qquad
\operatorname{Dist}_{\mathrm{shape}}
=
\frac{1}{2}
\sum_{i=1}^{2}\delta_i.
\]
where \(T\in\mathcal P_i\) is a binary foreground template in the darkness representation. 
Shape distortion is zero when each recovered component is a valid instance of its labelled prior shape at any generator-valid location.
It increases when a component is blurred, deformed, missing, or has the wrong shape.
Together, reconstruction RMSE, source IoU, and shape distortion measure assembly recovery, labelled source placement, and source-shape validity, respectively.

\begin{figure}[h]
\centering
\includegraphics[
    width=0.95\linewidth,
    trim=0 0 10 0,
    clip
]{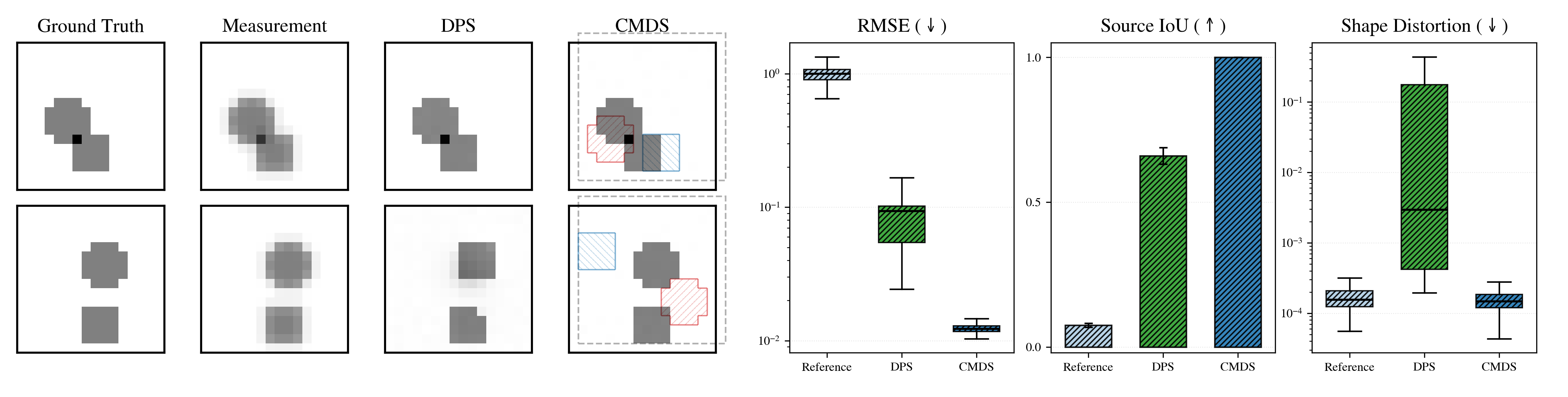}
\vspace{-.5em}
\caption{
    \textbf{Source separation from a blurred mixture (\(\tau=0.5\)).}
    Each task contains a ground-truth circle \(X_{\mathrm{gt}}^1\)
    and square \(X_{\mathrm{gt}}^2\).
    We observe the blurred mixture
    \(
    b={\rm A}_{0.5}(Y_{\mathrm{gt}})
    = {\rm A}_{0.5}(X_{\mathrm{gt}}^1+X_{\mathrm{gt}}^2)
    \)
    and recover the labelled sources.
    Left: two representative evaluation mixtures.
    The red and blue hatched shapes behind the \method{} reconstructions are the matched reference terminal sources.
    Right: results over \(1024\) evaluation mixtures for the reference process, DPS, and \method{}.
    Reconstruction RMSE measures whether the two recovered sources add up to the correct clean mixture.
    Source IoU measures whether each recovered object occupies the same pixels as its corresponding ground-truth object (\(\uparrow\)): it is \(1\) for an exact match and \(0\) when neither recovered source overlaps its corresponding ground-truth source.
    The bars show the evaluation-set mean with \(95\%\) bootstrap confidence intervals computed from \(10{,}000\) resamples.
    Shape distortion measures whether source \(1\) still looks like a circle and source \(2\) like a square, irrespective of their location (\(\downarrow\)).
    It is zero for an exact valid shape and increases when an object is blurred, deformed, missing, or has the wrong shape.
}
\label{fig:source-separation-results}
\vspace{-.5em}
\end{figure}

\paragraph{Additional evaluations.}
The aggregate comparisons over \(1024\) instances at \(\tau=0.5\) (low blur) and \(\tau=1.0\) (high blur) are shown in \cref{fig:source-separation-results} and \cref{fig:source-separation}, respectively.
\method{} achieves the lowest reconstruction RMSE and highest source IoU while retaining the low shape distortion of the frozen component priors.
Representative reconstructions for the low-blur setting are shown in \cref{fig:source-separation-samples-0p5}; corresponding results under the
stronger blur \(\tau=1.0\) are shown in \cref{fig:source-separation-samples-1}.
Under stronger blur, the observation contains less information about the locations of the individual components, and the recovered circle and square
can therefore exchange their ground-truth locations, as illustrated in the fourth row of \cref{fig:source-separation-samples-1}.
Such decompositions preserve plausible component shapes and can produce a similar blurred assembly.
The quadratic cost can consequently favour a nearby prior-consistent decomposition over the larger intervention required to recover the exact
ground-truth placement.


\subsection{PointMaze Collision Coordination}\label{apdx:maze-collision}
\begin{takehome}
\textbf{Key Takeaways:}
\noindent
\textbf{Multi-agent coordination can be amortised.}
A single control coordinates up to four frozen single-agent trajectory diffusion models, achieving high collision-free joint success and generalising to unseen start--goal regions.
\noindent
\textbf{Cross-agent communication must scale.}
Factoring attention into temporal and cross-agent passes preserves the timing of predicted conflicts while reducing complexity from $O(N^2H^2)$ to $O(NH(N+H))$.
\noindent
\textbf{Tweedie reward gradients improve control.}
Adding a learned time-dependent scaling of the reward gradient at the Tweedie estimate improves performance, with Pool~2 joint success ranging from \(89.8\%\) to \(99.2\%\).
\noindent
\textbf{Robust regression stabilises adjoint matching.}
Squared adjoint matching can be seed-sensitive in harder settings. 
The self-calibrated pseudo-Huber loss improves mean joint success in six of eight comparisons and reduces variance in the most unstable cases, but does not inherit the gradient identity or critical-point guarantee of squared adjoint matching.
\end{takehome}

This section provides detail on the protocol for the maze collision coordination experiment.

\subsubsection{Task, State, and Reward}
We first recall the setup of the task.
Agent $i$ generates a length-$H$ trajectory $\tau_i=(p_i(0),\ldots,p_i(H-1))\in\mathbb R^{H\times 2}$ using the same pretrained single-agent diffusion model $G$, instantiated independently for each agent.
We follow the discrete rollout convention in \cref{alg:cmds-discrete-am}.
Here, \(K\) is the number of reverse diffusion transitions, \(\ell\in\{0,\ldots,K\}\) indexes rollout states, and transitions
\(\ell=0,\ldots,K-1\) use native diffusion time-steps \(\zeta_\ell\).
Thus $X_0$ is the initial noise and $X_K$ is the terminal sample; the physical waypoint index $h\in\{0,\ldots,H-1\}$ is distinct from $\ell$.
The assembly operator stacks the $N$ component states,
\begin{align}
    Y_\ell
    =\varphi(X_\ell)
    =(X_\ell^1,\ldots,X_\ell^N),
\end{align}
where $\varphi$ is the identity on the concatenated state.
We write $\hat X_{1\mid\ell}$ for the frozen prior's terminal-state estimate and $\hat Y_{1\mid\ell}=\varphi(\hat X_{1\mid\ell})$ for its assembly.
Start and goal positions condition the reconstructed trajectory, which is learned as a residual modification to a straight-line base path interpolating between start and goal (\cref{apdx:maze-prior}).
Let $\kappa=\{(s_i,g_i)\}_{i=1}^N$ denote these fixed conditions and $\mathcal R_\kappa$ the residual-to-world reconstruction, so that
\[
    Z_\ell=\mathcal R_\kappa(Y_\ell),
    \qquad
    \hat Z_{1\mid\ell}=\mathcal R_\kappa(\hat Y_{1\mid\ell}).
\]
Coordination is introduced by a shared control conditioned on these reconstructed joint trajectories.
Goal reach is enforced structurally by the reconstruction and is not part of the reward.
\subsubsection{Reward}
For an assembled state $Y$, let
\[
    p_i(h;Y)
    \coloneqq
    \bigl[\mathcal R_\kappa(Y)\bigr]_{i,h}
    \in\mathbb R^2
\]
denote agent $i$'s reconstructed world-space position at waypoint $h$.
We suppress the dependence on $Y$ below.
The reward combines wall avoidance, inter-agent collision avoidance, and penalties on speed, acceleration, and path length.
The terminal reward is evaluated at $Y_K$.

The path penalties use the waypoint grid $\mathcal T=\{0,\ldots,H-1\}$.
Collision penalties use a denser linearly interpolated grid $\mathcal T_{\mathrm{dense}}$, with $|\mathcal T_{\mathrm{dense}}|=5(H-1)$ evaluation points.
Positions at fractional waypoint indices are obtained by linear interpolation.
We write $[a]_+=\max\{a,0\}$, so collision penalties vanish once the prescribed clearance is satisfied.
\paragraph{Wall penalty.}
Wall avoidance is enforced through the signed distance to the maze geometry.
We refer to \emph{world space} as the two-dimensional maze coordinate system in which the start and goal positions, maze geometry, and reconstructed agent trajectories are expressed.
For each agent, we penalise only positions whose clearance from the nearest wall is smaller than the agent radius:
\begin{align}
    r_{\mathrm{wall}}^{(i)}(Y)
    =
    -\lambda_{\mathrm{wall}}
    \sum_{h\in\mathcal T_{\mathrm{dense}}}
    \left[
        r_{\mathrm{agent}}
        -
        \mathrm{sdf}\!\left(p_i(h)\right)
    \right]_+^2.
\end{align}
For a world-space position ($p\in\mathbb R^2$), let ($\mathrm{sdf}(p)$) denote its signed Euclidean distance to the boundary of the collision-free maze region, taken to be positive in free space, zero on the boundary, and negative inside an obstacle or outside the maze boundary.
The hinge therefore vanishes whenever the trajectory maintains at least one agent radius of wall clearance.

\paragraph{Inter-agent collision penalty.}
Collisions between agents are handled analogously by penalising insufficient pairwise separation along the \textit{densified} trajectory.
We define
\begin{align}
    r_{\mathrm{pair}}(Y)
    =
    -\frac{\lambda_{\mathrm{pair}}}{\binom{N}{2}}
    \sum_{i<j}
    \sum_{h\in\mathcal T_{\mathrm{dense}}}
    \left[
        2r_{\mathrm{agent}}+m
        -
        \|p_i(h)-p_j(h)\|
    \right]_+^2,
\end{align}
where $m$ is an additional safety margin.
The term is zero once two agents are separated by at least $2r_{\mathrm{agent}}+m$, and the factor $\binom{N}{2}^{-1}$ averages the penalty over agent pairs.
\paragraph{Path regularisation.}
The collision terms constrain feasibility but do not by themselves favour short or smooth trajectories.
We therefore include three per-agent regularisers on the raw waypoint grid $\mathcal T$: a one-sided penalty on excessive waypoint displacement, a discrete acceleration penalty, and total path length.
These are
\begin{align}
    r_{\mathrm{speed}}^{(i)}(Y)
    &=
    -\lambda_{\mathrm{speed}}
    \sum_{h=0}^{H-2}
    \left[
        \|p_i(h{+}1)-p_i(h)\|-v_{\max}
    \right]_+^2,
    \\
    r_{\mathrm{accel}}^{(i)}(Y)
    &=
    -\lambda_{\mathrm{accel}}
    \sum_{h=1}^{H-2}
    \left\|
        p_i(h{+}1)-2p_i(h)+p_i(h{-}1)
    \right\|^2,
    \\
    r_{\mathrm{length}}^{(i)}(Y)
    &=
    -\lambda_{\mathrm{length}}
    \sum_{h=0}^{H-2}
    \|p_i(h{+}1)-p_i(h)\|.
\end{align}
The speed term is active only above $v_{\max}$, whereas acceleration and path length are penalised throughout the trajectory.

\paragraph{Total reward.}
Each agent receives its own wall and path penalties together with the shared inter-agent penalty:
\begin{align}
    r^{(i)}(Y)
    &=
    r_{\mathrm{wall}}^{(i)}(Y)
    +
    r_{\mathrm{speed}}^{(i)}(Y)
    +
    r_{\mathrm{accel}}^{(i)}(Y)
    \notag\\
    &\quad+
    r_{\mathrm{length}}^{(i)}(Y)
    +
    r_{\mathrm{pair}}(Y).
\end{align}
For a world-space trajectory $Z$, let $r_\kappa^{\mathrm{world}}(Z)$ denote this reward evaluated directly at positions $p_i(h)=Z_{i,h}$, so that
$r_\kappa(Y)=r_\kappa^{\mathrm{world}}(\mathcal R_\kappa(Y))$.
The subscripted penalty weights are distinct from the outer SOC reward strength $\lambda$.
We use
\[
    \bigl(
        \lambda_{\mathrm{wall}},
        \lambda_{\mathrm{pair}},
        \lambda_{\mathrm{speed}},
        \lambda_{\mathrm{accel}},
        \lambda_{\mathrm{length}}
    \bigr)
    =
    (200,400,0,1,1),
\]
with $m=0.05$ and $v_{\max}=0.15$.
The Tweedie reward-gradient extension (\cref{apdx:maze-axial-arch}) uses the same reward definition, except that $\lambda_{\mathrm{wall}}=300$.

\subsubsection{Data splits.}\label{apdx:maze-data-splits}
\begin{figure}
\centering
\begin{subfigure}[t]{0.44\textwidth}
    \centering
    \includegraphics[width=\linewidth]{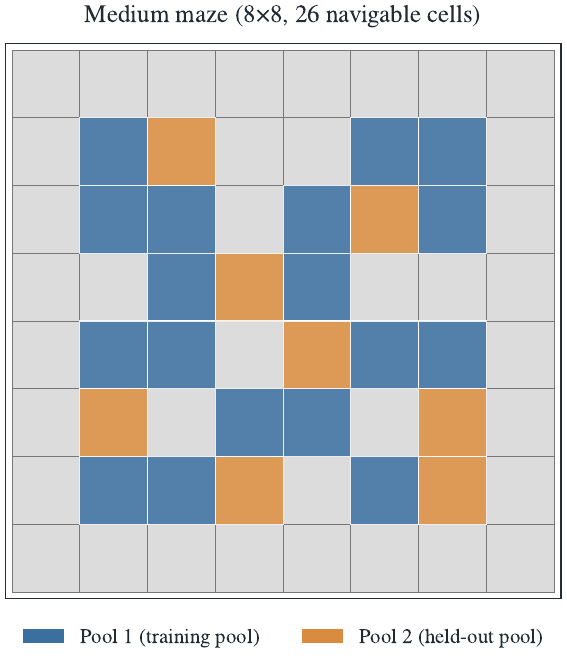}
    \caption{Medium ($8\times8$): 26 navigable cells, comprising 18 training and 8 held-out cells.}
\end{subfigure}
\hfill
\begin{subfigure}[t]{0.52\textwidth}
    \centering
    \includegraphics[width=\linewidth]{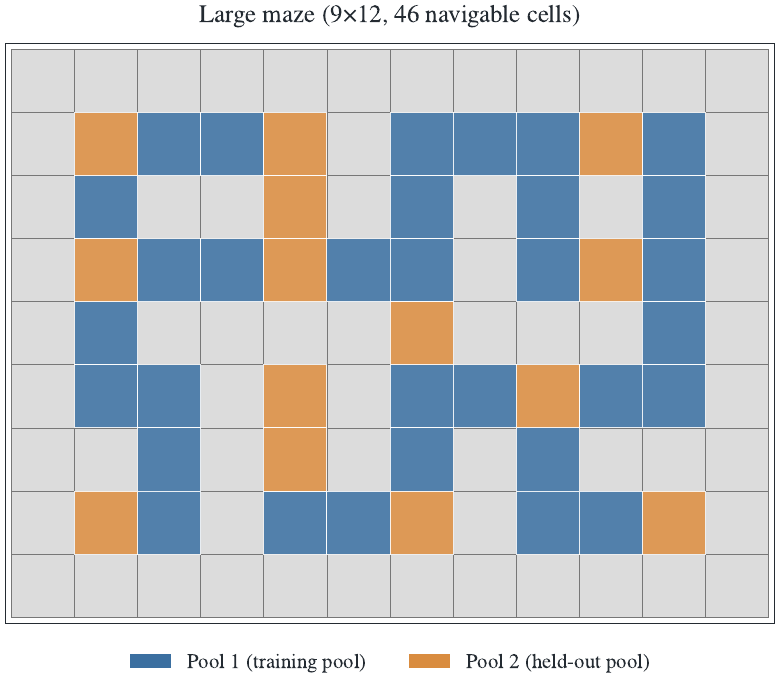}
    \caption{Large ($9\times12$): 46 navigable cells, comprising 32 training and 14 held-out cells.}
\end{subfigure}
\caption{\textbf{Fixed PointMaze cell partitions.}
Coordinates use zero-based $(\mathrm{row},\mathrm{column})$ matrix indexing.
The colours show the exact partition of cells between Pool 1 and Pool 2.}
\label{fig:maze-cell-splits}
\end{figure}

We use the medium and large PointMaze layouts \citep{fu2020d4rl}.
Each maze is mapped by a binary grid with 0 and 1 entries corresponding respectively to corridors and walls. 
The continuous PointMaze environment is obtained by fixing each grid entry to a physical size of $1.0\times1.0$.
Cells are partitioned once, in advance, into disjoint training and test pools:
\begin{center}
\small
\setlength{\tabcolsep}{7pt}
\renewcommand{\arraystretch}{1.08}
\begin{tabular}{lrrrr}
\toprule
\toprule
Layout & Grid & Navigable cells & Pool 1 cells & Pool 2 cells \\
\midrule
Medium & $8\times 8$  & 26 & 18 & 8  \\
Large  & $9\times 12$ & 46 & 32 & 14 \\
\bottomrule
\end{tabular}
\end{center}
The Pool 1 and Pool 2 cells are disjoint, spatially interspersed across each maze and fixed across runs. The exact partition of cells is presented in \cref{fig:maze-cell-splits}.

A training sample is obtained by sampling every start and goal cell independently and uniformly from the selected pool, adding continuous within-cell jitter, and retaining the draw only if:
\begin{enumerate}
    \item Every agent's uncontrolled path has a minimum length of 4.
    \item The independently planned paths contain collisions.
    \item Conflict-Based Search (CBS) \citep{sharon2015conflict} finds a joint solution within a specified compute budget.
\end{enumerate}

We report two held-out splits. Pool 1 contains unseen instances sampled from the training-cell pool; Pool 2 contains instances sampled from the disjoint test-cell pool. 
Every method and hyperparameter setting is evaluated on the same fixed set of $128$ instances per pool.

\subsubsection{Frozen Per-Agent Prior}\label{apdx:maze-prior}
Each layout uses a single-agent trajectory diffusion model, pretrained once and frozen throughout coordination training. 
We adopt the \emph{Diffusion Veteran} configuration \citep{lu2025makes}: a DiT-style denoiser with six adaLN-zero blocks, hidden width $256$, four attention heads, and $\approx 7.3$M parameters. 
The model is trained as a $100$-step discrete-time DDPM with a linear schedule and mean-squared $\epsilon$-prediction loss. 
This configuration reaches near-perfect ($\approx 0.99$) wall feasibility on both layouts.

\subsubsection{Control Architecture}\label{apdx:maze-axial-arch}
\paragraph{Base control.}
The pretrained diffusion model represents each trajectory as a residual correction to the straight-line path between the start and goal.
Before passing the trajectory to the control network, we reconstruct the corresponding absolute $(x,y)$ positions in the maze.
Cross-agent attention therefore operates on the agents' predicted physical positions rather than on the residual representation used by the diffusion model.
At reverse transition $\ell$, it receives the joint trajectory $Z_\ell$, the reconstructed Tweedie estimate $\hat Z_{1\mid\ell}$, the start--goal conditions $\kappa$, and the native diffusion time-step $\zeta_\ell$.
The trajectory and Tweedie estimate are concatenated along the channel axis and projected to width $344$, with sinusoidal embeddings of diffusion time and physical waypoint position.
An initial temporal self-attention block processes each agent's trajectory separately.
Two axial blocks then alternate temporal self-attention within each agent, cross-agent attention at each shared waypoint, and a feed-forward layer.
Both blocks retain the full waypoint resolution and use four attention heads, a feed-forward multiplier of $4$, residual connections, and layer normalisation.
Cross-agent attention includes a learned geometric bias.
For each agent pair, an eight-dimensional feature collects relative position ($2$), distance ($1$), forward-differenced relative velocity ($2$), closing speed ($1$), and relative goal offset ($2$).
An MLP with dimensions $8\!\to\!32\!\to\!4$ maps these features to an additive bias for each attention head, applied through a custom scaled-dot-product implementation.
Separate output heads read the local and communication representations to produce the epsilon residual $\delta\epsilon_{\theta,\ell}$.
Both heads and the final layer of the geometric-bias MLP are zero-initialised, so the base control initially leaves the pretrained sampler unchanged.
The network has \(\sim 4.3\)M parameters.
Rather than applying attention jointly over all $N\times H$ agent--waypoint tokens, we factor it along the two natural axes of the trajectory tensor \citep{ho2019axial}: temporal attention operates over waypoints within each agent, while cross-agent attention operates over agents at each shared waypoint.
This reduces the per-block attention complexity from $O(N^2H^2)$ to $O(NH(N+H))$.
For $H=50$ and $N=4$, this gives approximately $3.7$ times fewer attention-pair evaluations.
Temporal attention is batched over agents, and cross-agent attention is batched over waypoints.
Shared weights and attention along the agent axis make the control equivariant to a consistent permutation of agents and their conditions \citep{lee2019set}.
\paragraph{Tweedie reward gradient extension.}
Inspired by \citet{denker2024deft,venkatraman2024amortizing}, \method{}+TRG augments the base axial control with the joint reward gradient
\(
    \hat g_\ell^{\mathrm{raw}}
    =\lambda^\star
    \left.\nabla_Z
    r_\kappa^{\mathrm{world}}(Z)
    \right|_{Z=\hat Z_{1\mid\ell}},
\)
evaluated at the reconstructed Tweedie estimate $\hat Z_{1\mid\ell}$ where $\lambda^\star>0$ is a fixed scale, distinct from the outer SOC reward strength $\lambda$.
This is normalised to unit root-mean-square magnitude and clamped element-wise to $[-3,3]$ to give $\hat g_\ell$ ($\lambda^\star$ cancels under this normalisation, so its value does not affect $\hat g_\ell$).
This is combined additively with the base network:
\begin{align}
    \label{eq:cmds_trg}
    \delta\epsilon_{\theta,\ell}
    =
    \mathrm{NN}_\theta\!\left(Z_\ell,\hat Z_{1\mid\ell},\hat g_\ell,\kappa,\zeta_\ell\right)
    -
    s_\theta(\zeta_\ell)\,\hat g_\ell,
\end{align}
where $\delta\epsilon_{\theta,\ell}$ is the additive epsilon residual, inducing the normalised control $w_{\theta,\ell} =(\gamma_\ell/\varsigma_\ell)\delta\epsilon_{\theta,\ell}$ as in \cref{eq:cmds-ddim-control}.
Here $\mathrm{NN}_\theta$ is the same control described above, applied to a widened, three-channel input -- the reconstructed noisy state, the reconstructed Tweedie estimate, and $\hat g_\ell$, concatenated along the channel axis (so it can condition on the reward gradient rather than only react to its additive contribution) -- and $s_\theta(\zeta_\ell)=1+\tanh(\mathrm{MLP}_\theta(\zeta_\ell))$ is a single learned, time-dependent, agent-shared scalar gate (a function of the diffusion time-step only, not of $\hat Z_{1\mid\ell}$ or $\hat g_\ell$).
The gate MLP is zero-initialised so that $s_\theta(\zeta_\ell)=1$ at the start of training.
This adds only $\approx 1{,}000$ parameters over the base control (approximately $4{.}3$M total).
We use this full configuration (both the network conditioning and the gated additive term) throughout; restricting to either component alone, or removing the network's access to $\hat Z_{1\mid\ell}$ or $\hat g_\ell$, trades off differently across cells but does not change the qualitative picture and is left to future ablation work.

\subsubsection{Training Objectives}
We apply the adjoint matching coordination objective introduced in \cref{method:multi_agent_adjoint} to the stacked \(N\)-agent state. 
Adjoint matching rolls out the current controlled trajectory along the reverse grid with no gradient tracking.
Starting from the terminal cost gradient $a_K=\nabla_{X_K}[-\lambda r_\kappa(Y_K)]$, it propagates an adjoint variable backward one DDIM step at a time.
This backward pass replays individual controlled transitions rather than retaining the computational graph of the full multi-step sampler. 
The control is then trained to match the correction implied by the next-state adjoint $a_{\ell+1}$.
All headline results use the robust regression variant described in \cref{apdx:maze-robust-loss}.
At low signal-to-noise ratios, the DiT prior can also produce excessively large \(\hat X_{1\mid\ell}\) estimates; following \citet{janner2022planning}, we clip \(\hat X_{1\mid\ell}\) before converting it back to the equivalent noise prediction used in the control-penalty decomposition.

The adjoint matching configuration is summarised below.

\begin{center}
\small
\setlength{\tabcolsep}{6pt}
\renewcommand{\arraystretch}{1.08}
\begin{tabular}{@{}p{0.27\linewidth}p{0.65\linewidth}@{}}
\toprule\toprule
Setting & Value \\
\midrule
Optimiser steps
& \(15{,}000\) \\

Initial learning rate
& \(10^{-4}\) (Medium \(N{=}2\), Medium \(N{=}4\), Large \(N{=}2\)); \(3\times10^{-4}\) (Large \(N{=}4\))  \\

Learning-rate schedule
& Cosine decay to \(10^{-5}\) \\

EMA decay
& \(0.9995\) \\

Clipping
& Gradient norm \(0.5\), elementwise \(\hat X_{1\mid\ell}\) magnitude \(4\) \\

Batch size
& \(256\) for \(N=2\), \(128\) for \(N=4\) \\

Training instances
& \(4096\) joint instances for \(N=2\);
  \(2048\) for \(N=4\) \\
\bottomrule
\end{tabular}
\end{center}

The table above is \method{}'s recipe. 
\method{}+TRG converges in fewer steps and uses a correspondingly slower schedule: \(7{,}000\) optimiser steps, initial learning rate \(10^{-5}\) with cosine decay to \(10^{-6}\), and reward wall-weight \(\lambda_{\mathrm{wall}}=300\).
Batch size, training instances, EMA decay, and gradient/\(\hat X_{1\mid\ell}\) clipping are shared between both variants.
\subsubsection{Robust Adjoint-Matching: pseudo-Huber versus squared \(\ell_2\)}\label{apdx:maze-robust-loss}

\paragraph{Motivation.} Adjoint matching is an on-policy stochastic regression problem whose inputs and targets depend on the current control.
Because adjoints are propagated through Jacobians evaluated along stochastic, control-dependent trajectories, finite-batch gradient estimates can have high variance despite being unbiased in expectation \citep{domingoenrich2025adjoint}.
The squared matching loss can amplify this variability because its gradient grows with the adjoint residual.
For collision rewards, early trajectories often contain persistent agent collisions or deep wall penetrations, which can dominate minibatch adjoints and make optimisation unstable and strongly seed-dependent.
This motivates a modification of adjoint matching's regression objective.
\paragraph{Objective.} Using the rollout convention of \cref{alg:cmds-discrete-am}, with the frozen-rollout superscript suppressed, we write the joint matching residual as
\[
    e_{\theta,\ell} \coloneqq \operatorname{col}_{i=1}^{N} \left[ w_{\theta,\ell}^i\!\left( \operatorname{stopgrad}(X_\ell) \right) + \varsigma_\ell \operatorname{stopgrad}\!\left( a_{\ell+1}^{i} \right) \right] \in\mathbb R^{D}.
\]
Here $D$ is the dimension of the stacked residual after vectorisation ($D=2NH$ for the full $H\times2$ representation).
This is the residual appearing in \cref{eq:cmds-ddim-am-loss}.
The normalised control is defined in \cref{eq:cmds-ddim-control}, and the adjoint target is obtained from the terminal condition \cref{eq:cmds-ddim-terminal-adjoint} and recursion \cref{eq:cmds-discrete-full-adjoint}.
To reduce the influence of rare high-norm residuals induced by these recursively propagated targets, we use a pseudo-Huber, or Charbonnier loss \citep{huber1992robust,charbonnier1994two,barron2019general}.
For a residual vector \(e\in\mathbb R^{D}\), let \(\operatorname{rms}(e)=\|e\|/\sqrt{D}\) be its root-mean-square magnitude and define
\begin{align}
    \rho_{\delta_\ell}(e)
    \coloneqq
    D\delta_\ell^2
    \left(
    \sqrt{1+\frac{\|e\|^2}{D\delta_\ell^2}}
    -1
    \right).
    \label{eq:pseudo-huber}
    \end{align}
    The resulting robust matching objective is
    \begin{align}
    \mathcal L_{\method{}\text{-}\mathrm{PH}}(\theta)
    \coloneqq
    \E\!\left[
    \sum_{\ell\in\mathcal K}
    \rho_{\delta_\ell}(e_{\theta,\ell})
    \right].
    \label{eq:maze-am-pseudo-huber-loss}
    \end{align}
    For \(\|e\|\ll\sqrt{D}\delta_\ell\),
    \[
    \rho_{\delta_\ell}(e)
    =
    \frac12\|e\|^2
    +
    O\!\left(
    \frac{\|e\|^4}{D\delta_\ell^2}
    \right),
    \]
    so \(\nabla_e^2\rho_{\delta_\ell}(0)=I\) and the loss retains the local curvature of squared matching.
    For \(\|e\|/(\sqrt{D}\delta_\ell)\to\infty\), it instead satisfies
    \(\rho_{\delta_\ell}(e)=\sqrt{D}\delta_\ell\|e\|+O(1)\), and
    \begin{align}
    \nabla_e\rho_{\delta_\ell}(e)
    &=
    \frac{e}{
    \sqrt{1+\|e\|^2/(D\delta_\ell^2)}
    },
    \qquad
    \|\nabla_e\rho_{\delta_\ell}(e)\|
    \le
    \sqrt{D}\,\delta_\ell.
    \label{eq:pseudo-huber-influence}
\end{align}
Because both the rollout and adjoint target are stopped in \cref{eq:cmds-ddim-am-loss},
\[
    \nabla_\theta\rho_{\delta_\ell}(e_{\theta,\ell})
    =
    D_\theta w_{\theta,\ell}^{\top}
    \nabla_e\rho_{\delta_\ell}(e_{\theta,\ell}).
\]
Thus, pseudo-Huber bounds the influence of an individual target in residual space, although the parameter gradient can still be amplified by the control Jacobian \(D_\theta w_{\theta,\ell}\).
\paragraph{Self-calibration of the pseudo-Huber loss.}
Adjoint residual scales differ substantially across DDIM transitions, so a single global threshold would constrain some steps too aggressively and others too weakly.
We first run a short calibration phase with the squared DDIM matching objective \cref{eq:cmds-ddim-am-loss} and record the per-instance RMS residual \(\operatorname{rms}(e_{\theta,\ell}^{(b)})\) at each transition \(\ell\).
We then set
\begin{align}
    \delta_\ell
    &=
    \max\!\left\{
    \delta_{\min},
    Q_q\!\left(
    \{\operatorname{rms}(e_{\theta,\ell}^{(b)})\}_{b\in\mathcal D_{\mathrm{cal}}}
    \right)
    \right\}.
\label{eq:pseudo-huber-calibration}
\end{align}
Here, \(Q_q\) is a fixed empirical quantile over the calibration set \(\mathcal D_{\mathrm{cal}}\); we freeze \(\delta_\ell\) for the remainder of training.
The calibration hyperparameters are fixed before the multi-seed comparison and shared across mazes and agent counts.
The underlying SOC objective remains \cref{eq:adjoint-objective}; only the squared regression surrogate \cref{eq:cmds-ddim-am-loss} is replaced.
The gradient identity and critical-point guarantee for the basic squared objective \cref{eq:cmad-basic-am}, established by \citet{domingoenrich2025adjoint}, do not automatically extend to \cref{eq:maze-am-pseudo-huber-loss}.
This modified regression objective is studied empirically and need not preserve the minimiser of squared adjoint matching; its connection to the original stochastic-control objective remains an open question.
\Cref{tab:robust-loss-ablation} reports the full squared-\(\ell_2\)-versus-pseudo-Huber regression comparison, held-out joint success, mean \(\pm\) standard deviation across \(10\) independent training seeds per (maze, \(N\)) cell.
All runs use identical hyperparameters tuned for \(\ell_2\)-AM, identical control architecture (\Cref{apdx:maze-axial-arch}), and evaluation protocol.
The pseudo-Huber matching loss \eqref{eq:pseudo-huber} generally improves mean success and substantially reduces variance in challenging settings, such as the Large maze with 4 agents.

\begin{table}[!h]
\centering
\small
\setlength{\tabcolsep}{4pt}
\renewcommand{\arraystretch}{1.1}

\resizebox{\linewidth}{!}{%
\begin{tabular}{lcccccccc}
\toprule
& \multicolumn{2}{c}{Medium (\(N=2\))}
& \multicolumn{2}{c}{Medium (\(N=4\))}
& \multicolumn{2}{c}{Large (\(N=2\))}
& \multicolumn{2}{c}{Large (\(N=4\))} \\
\cmidrule(lr){2-3}
\cmidrule(lr){4-5}
\cmidrule(lr){6-7}
\cmidrule(lr){8-9}
Matching loss
& Pool 1 \(\uparrow\)
& Pool 2 \(\uparrow\)
& Pool 1 \(\uparrow\)
& Pool 2 \(\uparrow\)
& Pool 1 \(\uparrow\)
& Pool 2 \(\uparrow\)
& Pool 1 \(\uparrow\)
& Pool 2 \(\uparrow\) \\
\midrule
Squared \(\ell_2\)
& \(0.940\,{\color{gray}\scriptstyle\pm0.020}\)
& \(0.884\,{\color{gray}\scriptstyle\pm0.048}\)
& \(\mathbf{0.806}\,{\color{gray}\scriptstyle\pm0.155}\)
& \(0.667\,{\color{gray}\scriptstyle\pm0.151}\)
& \(0.833\,{\color{gray}\scriptstyle\pm0.041}\)
& \(0.792\,{\color{gray}\scriptstyle\pm0.042}\)
& \(0.723\,{\color{gray}\scriptstyle\pm0.157}\)
& \(0.606\,{\color{gray}\scriptstyle\pm0.152}\) \\
\rowcolor{pastelblue!25}
Pseudo-Huber
& \(\mathbf{0.988}\,{\color{gray}\scriptstyle\pm0.012}\)
& \(\mathbf{0.933}\,{\color{gray}\scriptstyle\pm0.032}\)
& \(0.805\,{\color{gray}\scriptstyle\pm0.098}\)
& \(0.665\,{\color{gray}\scriptstyle\pm0.169}\)
& \(\mathbf{0.907}\,{\color{gray}\scriptstyle\pm0.037}\)
& \(\mathbf{0.833}\,{\color{gray}\scriptstyle\pm0.045}\)
& \(\mathbf{0.873}\,{\color{gray}\scriptstyle\pm0.027}\)
& \(\mathbf{0.744}\,{\scriptstyle\pm0.036}\) \\
\bottomrule
\end{tabular}}%
\vspace{.5em}
\caption{
\textbf{Adjoint matching training-seed variance and its mitigation.}
Held-out joint success, mean \(\pm\) population standard deviation across \(10\) independent training seeds per (maze, \(N\)) cell.
}
\label{tab:robust-loss-ablation}
\end{table}

\subsubsection{Training Variance}\label{apdx:maze-training-variance}
We report the final \method{}+TRG training variance.
We train \method{}+TRG with 5 different seeds for each setting and report the mean and standard deviation of joint success on each pool.
\begin{table}[!h]
\centering
\small
\setlength{\tabcolsep}{4pt}
\renewcommand{\arraystretch}{1.1}
\resizebox{\linewidth}{!}{%
\begin{tabular}{lcccccccc}
\toprule\toprule
& \multicolumn{2}{c}{Medium (\(N=2\))}
& \multicolumn{2}{c}{Medium (\(N=4\))}
& \multicolumn{2}{c}{Large (\(N=2\))}
& \multicolumn{2}{c}{Large (\(N=4\))} \\
\cmidrule(lr){2-3}
\cmidrule(lr){4-5}
\cmidrule(lr){6-7}
\cmidrule(lr){8-9}

Method
& Pool 1 \(\uparrow\) & Pool 2 \(\uparrow\)
& Pool 1 \(\uparrow\) & Pool 2 \(\uparrow\)
& Pool 1 \(\uparrow\) & Pool 2 \(\uparrow\)
& Pool 1 \(\uparrow\) & Pool 2 \(\uparrow\) \\
\midrule

\method{}+TRG
& \(\mathbf{0.998}\,{\color{gray}\scriptstyle\pm0.003}\)
& \(\mathbf{0.983}\,{\color{gray}\scriptstyle\pm0.015}\)
& \(\mathbf{0.984}\,{\color{gray}\scriptstyle\pm0.005}\)
& \(\mathbf{0.947}\,{\color{gray}\scriptstyle\pm0.006}\)
& \(\mathbf{0.973}\,{\color{gray}\scriptstyle\pm0.009}\)
& \(\mathbf{0.906}\,{\color{gray}\scriptstyle\pm0.015}\)
& \(\mathbf{0.975}\,{\color{gray}\scriptstyle\pm0.006}\)
& \(\mathbf{0.969}\,{\color{gray}\scriptstyle\pm0.005}\) \\
\bottomrule
\end{tabular}%
}
\caption{\textbf{\method{}+TRG training variance.}
    Held-out joint success rate (Pool 1: train-cell instances, Pool 2: instances from entirely held-out maze cells), mean \(\pm\) population standard deviation across 5 independent training seeds, evaluated on 128 held-out instances per split.}
\label{tab:seed-variance}
\end{table}

\subsubsection{Baseline Suite and Protocol Matching}\label{apdx:maze-baselines}
We compare against independent reference sampling, Best-of-\(256\) reward selection, FK-Steering, and DPS test-time reward guidance.
All baselines use the same frozen single-agent prior, trajectory representation, reward, agent radius \(r_{\mathrm{agent}}=0.12\), \(\hat X_{1\mid\ell}\)-clip of \(4\), and fixed evaluation sets of \(128\) instances per split.
DPS and FK-Steering are tuned for each task.

\subsubsection{Qualitative Examples}\label{apdx:maze-qualitative}
\begin{figure*}
\centering
\begingroup
\renewcommand{\thesubfigure}{\Roman{subfigure}}
\captionsetup[subfigure]{
    labelformat=parens,
    labelsep=space,
    justification=centering
}

\begin{subfigure}[t]{0.7\textwidth}
    \centering
    \includegraphics[width=\linewidth]{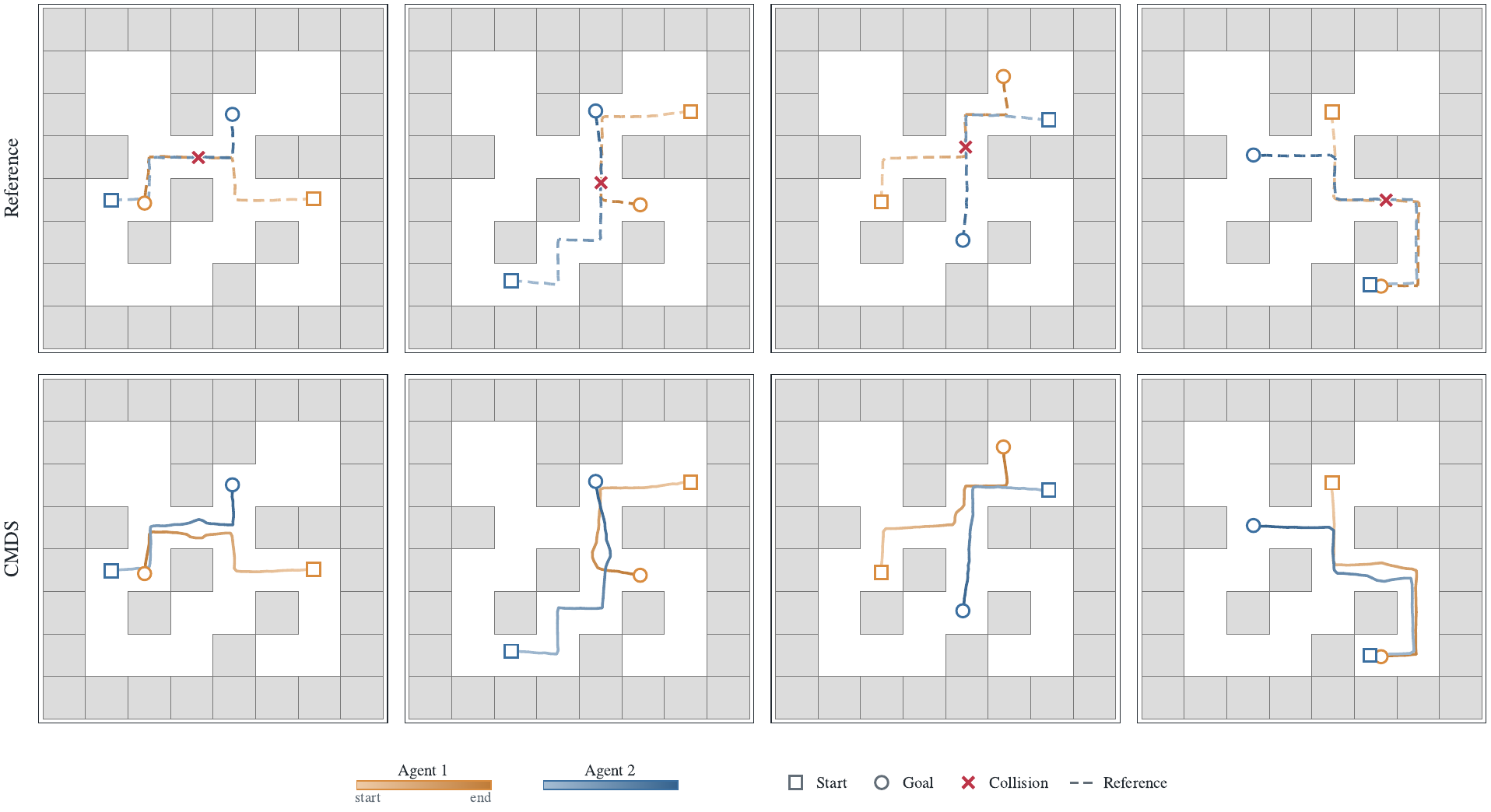}
    \caption{Medium PointMaze ($N=2$).}
    \label{fig:maze-qualitative-medium-n2}
\end{subfigure}

\medskip

\begin{subfigure}[t]{0.7\textwidth}
    \centering
    \includegraphics[width=\linewidth]{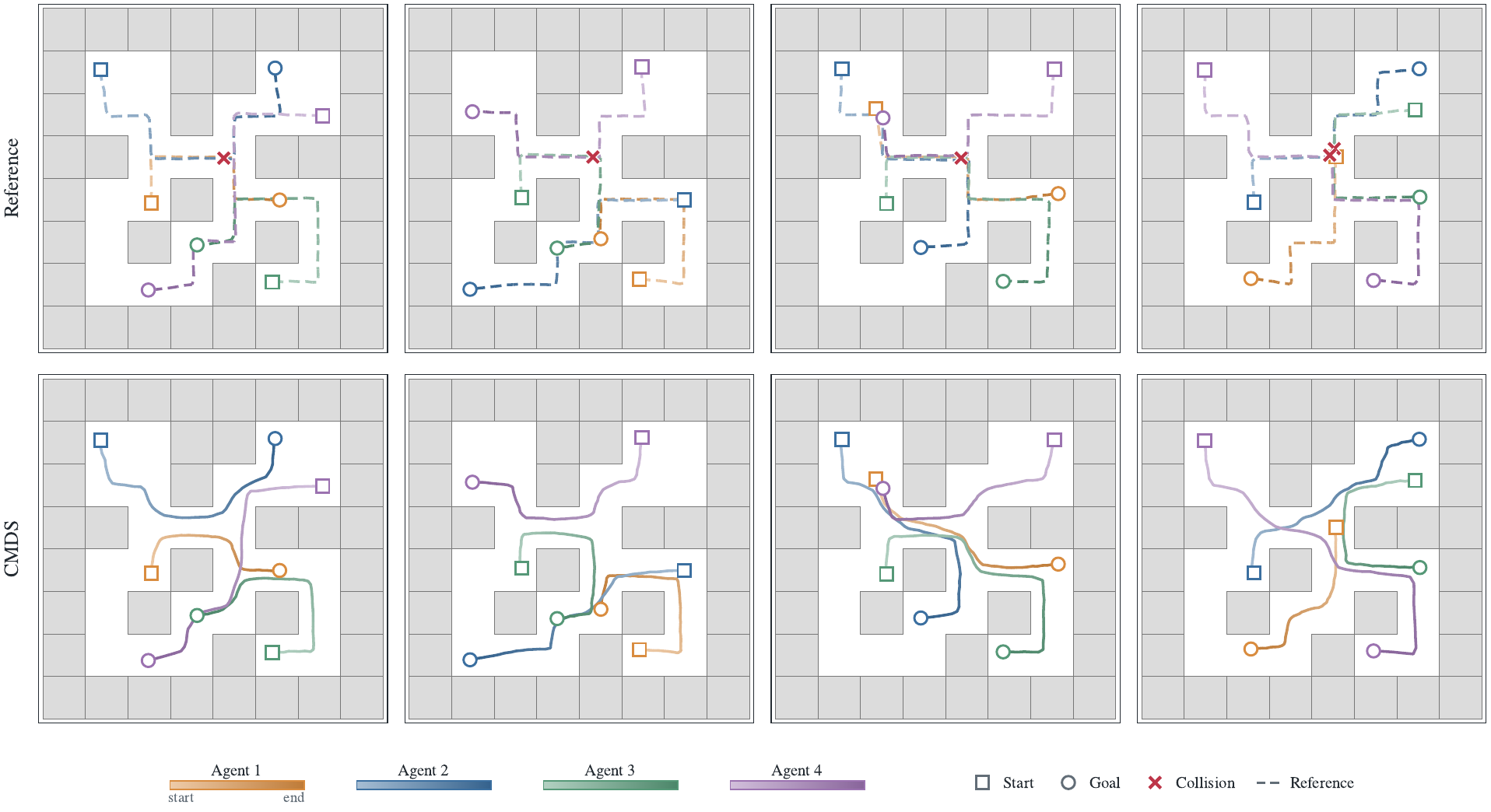}
    \caption{Medium PointMaze ($N=4$).}
    \label{fig:maze-qualitative-medium-n4}
\end{subfigure}

\medskip

\begin{subfigure}[t]{0.7\textwidth}
    \centering
    \includegraphics[width=\linewidth]{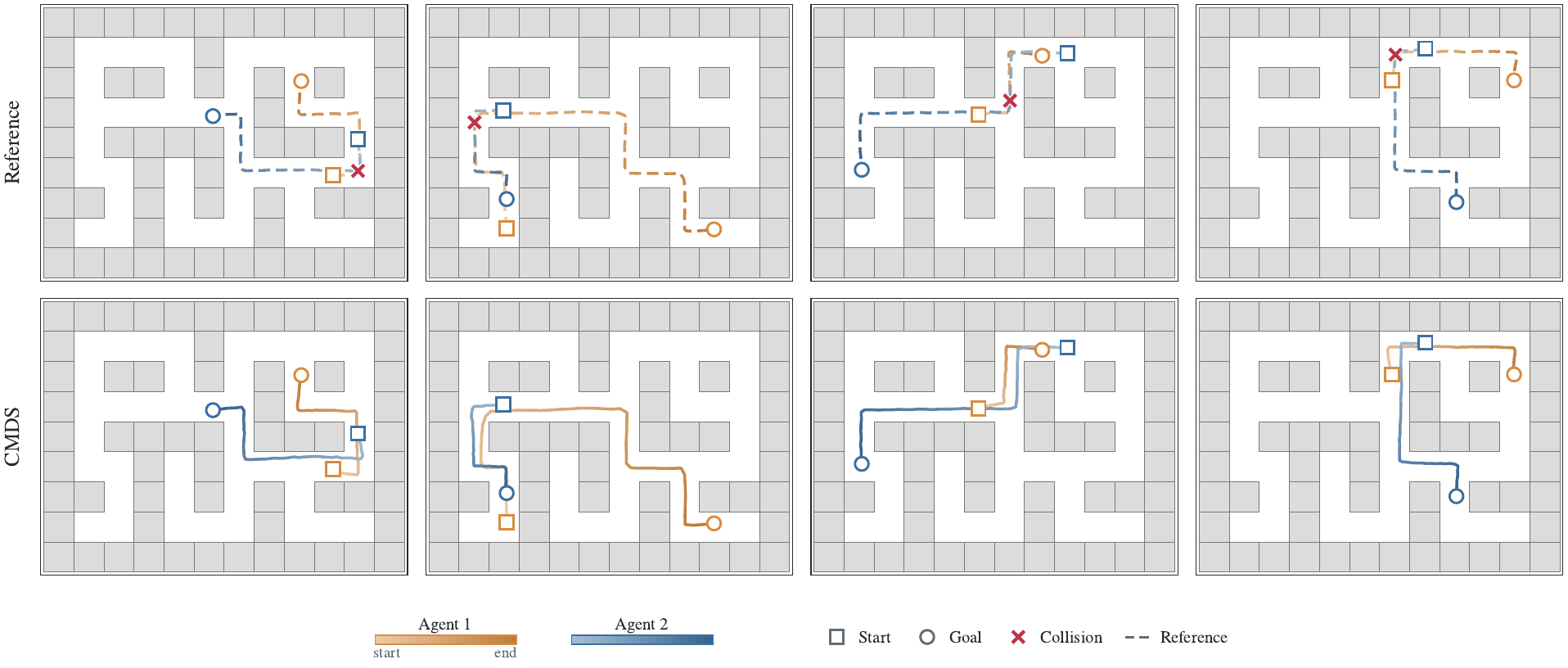}
    \caption{Large PointMaze ($N=2$).}
    \label{fig:maze-qualitative-large-n2}
\end{subfigure}

\medskip

\begin{subfigure}[t]{0.7\textwidth}
    \centering
    \includegraphics[width=\linewidth]{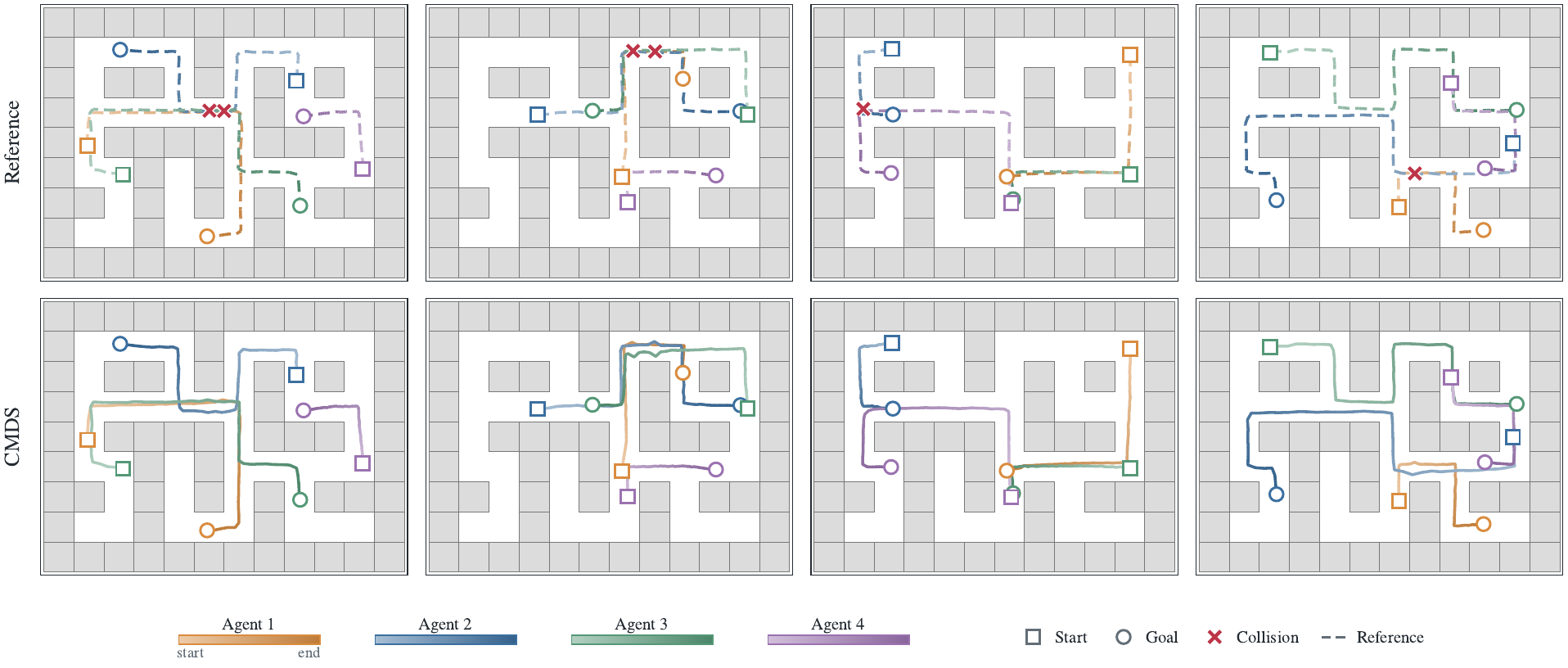}
    \caption{Large PointMaze ($N=4$).}
    \label{fig:maze-qualitative-large-n4}
\end{subfigure}

\caption{\textbf{Qualitative maze-coordination examples.}
For each maze and agent count, the top row is the reference process trajectory (dashed) and the bottom row is the trajectory produced with the trained \method{} control (solid), for the same \(6\) held-out instances and the same initial noise.
}
\label{fig:maze-qualitative-panel}
\endgroup
\end{figure*}

\Cref{fig:maze-qualitative-medium-n2,fig:maze-qualitative-medium-n4,fig:maze-qualitative-large-n2,fig:maze-qualitative-large-n4} give \(6\) held-out examples per (maze, \(N\)) cell, with the highest variation of the trajectory before and after introducing the control.

\subsection{Multi-Arm KUKA Coordination from Frozen Single-Arm Priors}\label{apdx:kuka-collision}
\begin{takehome}
\textbf{Key Takeaways:}
\textbf{Frozen energy-based priors can be coordinated.}
\method{} learns a shared control over independent seven-joint KUKA priors while keeping their pretrained weights fixed.
Adjoint training differentiates the reverse dynamics through the learned energy using input Hessian--vector products, evaluated one transition at a time.
\textbf{A two-arm controller generalises without retraining.}
Trained to coordinate two frozen single-arm priors, \method{} provides zero-shot transfer to shifted or enlarged obstacles, wider robot base spacing, and three to five arms.
It achieves higher mean strict success than the evaluated guidance baselines in all these transfer settings.
\end{takehome}

\subsubsection{Task, representation, and data}\label{apdx:kuka-data}
The KUKA task follows \citet{luo2024potential} and consists of two robotic arms whose single-arm trajectory priors are independently pretrained for
obstacle-avoiding motion planning.
Arm \(i\) generates joint angles \(q^i=(q^i_1,\ldots,q^i_H)\in\mathbb R^{H\times7}\), with \(H=56\).
All arms share a path-progress coordinate, and \(q^i_1,q^i_H\) are fixed to the requested start and goal.
The planner generates joint-angle paths without assigning physical
times to their waypoints or enforcing joint-velocity limits.
Evaluation checks geometric constraints, including joint limits and
collisions along trajectories.
With the saved prior-normalisation bounds \(q_{\mathrm{lo}},q_{\mathrm{hi}}\), the normalised full path is \(x=2(q-q_{\mathrm{lo}})/(q_{\mathrm{hi}}-q_{\mathrm{lo}})-1\), with component-wise operations and an affine inverse.
Only the \(H-2\) interior knots are sampled; endpoints are reinserted whenever the full trajectory is reconstructed.
%
%
We denote the joint native state by \(X=(X^1,\ldots,X^N)\), where \(X^i\in\mathbb R^{(H-2)\times7}\) contains arm \(i\)'s normalised interior coordinates.
For fixed task conditions \(\kappa\), comprising the endpoints, arm bases, and obstacles, the reconstructed assembly is \(Y=\varphi(X)=q=(q^1,\ldots,q^N)\); the dependence of \(\varphi\) on \(\kappa\) is suppressed.
The two-arm state has \(14(H-2)\approx750\) free coordinates.
For \(N\) arms, \(D=7N(H-2)\); norms on the native state and control are Euclidean after vectorisation, without averaging over arms or coordinates.

The arm bases are positioned at \((-0.5,0,0)\) and \((0.5,0,0)\)
metres, with both bases' local coordinate axes aligned with the
world axes.
The scene contains a shared table beneath the arms and five cubic
obstacles, each with side length \(0.44\,\mathrm m\) and edges
aligned with the world axes.

Single-arm priors receive obstacle centres expressed in each arm's local base coordinates; the controller and collision reward use world-frame (i.e. shared coordinate system) forward kinematics and the actual obstacle extents.
The pretrained conditioning interface remains fixed, while the joint control observes the transformed scene.

\paragraph{Priors.}
We instantiate the priors from the pretrained checkpoints of \citet{luo2024potential}. 
The Dual-arm checkpoint is used only for evaluation and comparison with \method{}\footnote{\url{https://github.com/devinluo27/potential-motion-plan-release}}.

\paragraph{Controller-training and evaluation populations.}
A layout specifies the obstacle arrangement, while a planning task specifies start and goal configurations for both arms
within that layout.
The released dataset provides separate training and test layouts.
We divide its 2,500 training layouts into 2,250 layouts for learning the controller and 250 layouts reserved for validation.
Validation is used to monitor training and select controller settings and guidance strengths. 
Test layouts are used to report performance.
These three groups contain no shared layouts.

For controller training, we extract 32 start--goal pairs from each of the 2,250 training layouts, giving 72,000 tasks.
Removing tasks whose start or goal configurations violate joint limits or contain collisions leaves 50 thousand training tasks.

For validation, we select two tasks from each of 100 of the 250 reserved layouts.
For testing, we select two tasks from each of 300 released test layouts.
Applying the same endpoint checks leaves 134 validation tasks and 413 test tasks, with the latter spanning 264 layouts.
All methods use the same retained tasks, selected before planner evaluation. 
Collision-free start and goal configurations do not guarantee that a collision-free path exists between them.

\paragraph{Zero-shot transfer.}
Each transformation below is applied to all test configurations.
Obstacle perturbations are deterministic per layout and shared across methods.
All controller and prior weights and guidance strengths are held fixed. 
The resulting number of tasks after feasibility filtering is detailed in \cref{tab:kuka-populations}.

\begin{table}[tb]
\centering\small
\renewcommand{\arraystretch}{1.13}
    \begin{tabular}{@{}p{0.20\linewidth}p{0.64\linewidth}r@{}}
    \toprule
    Population & Construction & Evaluation tasks \\
    \midrule
    In-domain & Released dual-arm scene and endpoints & 413 \\
    Shifted obstacles & Horizontal Gaussian shifts, standard deviation 0.10\,m & 191 \\
    Larger obstacles & Each box half-extent multiplied by 1.2 & 121 \\
    Wider bases & Base positions moved to \(x=\pm0.6\)\,m & 208 \\
    Three arms & Third base at \((0,0.8,0)\)\,m, identity rotation & 172 \\
    \bottomrule
    \end{tabular}
    \vspace{0.25em}
\caption{\textbf{KUKA evaluation populations.}}
\label{tab:kuka-populations}
\end{table}

For wider bases, joint-space endpoints and world-frame obstacle positions remain fixed, so world-space end-effector endpoints change.

\paragraph{Transfer from two to five arms.}\label{apdx:kuka-arm-count}
\method{} is architecturally compatible with an arbitrary number of arms without retraining.
For the arm-count comparison in the right-hand block of \cref{tab:kuka-generalization}, we select twenty start--goal tasks from each of the 300 test layouts, giving 6,000 tasks.
Selection is fixed before running any planner.
We keep the original two arms and obstacles unchanged, then add arms at \((0,0.8,0)\), \((0,-0.8,0)\), and \((0.8,0.8,0)\) metres, all with the same base orientation as the original arms.
Within each layout, we pair consecutive tasks in the fixed selection order. 
We assign endpoints to the added arms using other two-arm tasks from the same obstacle layout.
We pair the twenty selected tasks in their fixed order: the first with the second, the third with the fourth, and so on.
For each task, arm three receives the start and goal joint angles of the paired task's first arm; arm four receives those of its second arm.
Arm five receives the first arm's start and goal joint angles from a third task.
We select this third task from all released tasks in the same layout, sorted by task identifier, taking the first entry other
than the current task, its paired task, or either of the two tasks used in the original two-arm benchmark.
Only endpoint joint angles are reused; trajectories are generated anew for each arm count.

Checking joint limits and endpoint collisions leaves 4 thousands 1,5 thousand, 500, and 200 tasks for two, three, four, and five arms, respectively.
We compare all arm counts on the same 209 tasks whose endpoints pass these checks in every setting.
These results are conditional on endpoint feasibility for all four counts; they do not describe the full retained population at each count or certify that a feasible connecting path exists.

\subsubsection{Energy-based priors and discrete adjoint matching}\label{apdx:kuka-adjoint}

\paragraph{Energy parameterisation.}
The released network \(f_\psi\) defines the energy parameterisation of \citet{du2023reduce,luo2024potential}.
For a normalised full path \(x\), diffusion index \(\zeta\), and scene conditioning \(c\), the noise predictor is
\begin{align}
    E_\psi(x,\zeta,c)=\tfrac12\|f_\psi(x,\zeta,c)\|_2^2,
    \qquad
    \epsilon_\psi(x,\zeta,c)=\nabla_x E_\psi(x,\zeta,c).
    \label{eq:kuka-energy}
\end{align}
The prior applies classifier-free guidance with weight \(\omega_{\mathrm{cfg}}=2\), which gives \(\epsilon_\psi^{\mathrm{cfg}}=\epsilon_\psi^{\varnothing}+\omega_{\mathrm{cfg}}(\epsilon_\psi^c-\epsilon_\psi^{\varnothing})\).
These are learned trajectory energies, distinct from the geometric collision reward defined below.

\paragraph{Reverse transition and control.}
We use the state and transition-index convention of \cref{alg:cmds-discrete-am}, but retain \(K=10\) for the number of DDIM transitions in this subsection, since \(n_{\mathrm{cand}}\) denotes the number of initial joint trajectory candidates per task in the KUKA experiments.
The rollout runs from the initial noise state \(X_0\) to the terminal sample \(X_K\), with transitions indexed by \(\ell=0,\ldots,K-1\) and native diffusion time-steps \(\zeta_\ell=90-10\ell\).
Let \(\bar\alpha_\zeta\) denote the checkpoint's cumulative noise-schedule coefficient at native time-step \(\zeta\), and define
\[
    \bar\alpha_{\mathrm{src},\ell}=\bar\alpha_{\zeta_\ell},
    \qquad
    \bar\alpha_{\mathrm{dst},\ell}=
    \begin{cases}
        \bar\alpha_{\zeta_{\ell+1}}, & 0\leq\ell<K-1,\\
        1, & \ell=K-1.
    \end{cases}
\]
The frozen noise prediction is evaluated on each arm's full normalised path with fixed endpoints, then restricted to its free interior.
We write the concatenated prediction as \(\epsilon_{\mathrm b,\ell}(X_\ell)=(\epsilon_{\mathrm b,\ell}^i(X_\ell^i))_{i=1}^N\), with evaluation at \(\zeta_\ell\), scene conditioning, and classifier-free guidance included.
The reference transition first clips its clean estimate and recomputes the corresponding epsilon:
\begin{equation}
    \begin{aligned}
    \hat X_{1\mid\ell}^{\mathrm{clip}}
    &=\operatorname{clip}_{[-1,1]}
    \left(
        \frac{X_\ell-\sqrt{1-\bar\alpha_{\mathrm{src},\ell}}\,
        \epsilon_{\mathrm b,\ell}(X_\ell)}
        {\sqrt{\bar\alpha_{\mathrm{src},\ell}}}
    \right),\\
    \widetilde\epsilon_{\mathrm b,\ell}(X_\ell)
    &=\frac{X_\ell-\sqrt{\bar\alpha_{\mathrm{src},\ell}}\,
        \hat X_{1\mid\ell}^{\mathrm{clip}}}
        {\sqrt{1-\bar\alpha_{\mathrm{src},\ell}}},\\
    \varsigma_\ell^2
    &=\eta^2\frac{1-\bar\alpha_{\mathrm{dst},\ell}}
        {1-\bar\alpha_{\mathrm{src},\ell}}
        \left(1-\frac{\bar\alpha_{\mathrm{src},\ell}}
        {\bar\alpha_{\mathrm{dst},\ell}}\right),\\
    \mu_{\mathrm b,\ell}(X_\ell)
    &=\sqrt{\bar\alpha_{\mathrm{dst},\ell}}\,
        \hat X_{1\mid\ell}^{\mathrm{clip}}
        +\sqrt{1-\bar\alpha_{\mathrm{dst},\ell}-\varsigma_\ell^2}\,
        \widetilde\epsilon_{\mathrm b,\ell}(X_\ell).
    \end{aligned}
    \label{eq:kuka-ddim-reference}
\end{equation}
For the learned-control sampler, \(\eta=1\), so \(\varsigma_\ell>0\) for \(0\leq\ell<K-1\), whereas \(\varsigma_{K-1}=0\).
For an epsilon correction \(\delta\epsilon_{\theta,\ell}(X_\ell)\), define on the stochastic transitions \(0\leq\ell<K-1\),
\begin{equation}
    \begin{aligned}
    \gamma_\ell
    &=\sqrt{1-\bar\alpha_{\mathrm{dst},\ell}-\varsigma_\ell^2}
        -\sqrt{\frac{\bar\alpha_{\mathrm{dst},\ell}}
        {\bar\alpha_{\mathrm{src},\ell}}}
        \sqrt{1-\bar\alpha_{\mathrm{src},\ell}},\\
     w_{\theta,\ell}(X_\ell)
    &=\frac{\gamma_\ell}{\varsigma_\ell}
        \delta\epsilon_{\theta,\ell}(X_\ell),\\
    \mu_\ell^\theta(X_\ell)
    &=\mu_{\mathrm b,\ell}(X_\ell)+\varsigma_\ell w_{\theta,\ell}(X_\ell),\\
     X_{\ell+1}
    &=\mu_\ell^\theta(X_\ell)+\varsigma_\ell\Xi_\ell,
        \qquad \Xi_\ell\sim\mathcal N(0,I_D).
    \end{aligned}
    \label{eq:kuka-control}
\end{equation}
At the deterministic final transition, no learned correction is applied:
\[
    X_K=\mu_{\mathrm b,K-1}(X_{K-1}),
    \qquad
    w_{\theta,K-1}\equiv0.
\]
The final control is set to zero by convention, not computed through division by \(\varsigma_{K-1}\).
Unlike the modified endpoint in \cref{apdx:ddim-adjoint-matching}, KUKA retains the native deterministic endpoint; it contributes neither control energy nor a matching term.

\paragraph{Adjoint matching with energy-based priors.}
We use the discrete adjoint recursion and matching loss of \Cref{apdx:ddim-adjoint-matching}.
The control cost is averaged over arms, and the collision reward has weight \(\beta=40\).
Adjoints propagate through all ten denoising transitions, while control costs and matching terms involve only the nine stochastic transitions.
Since \(\epsilon_\psi=\nabla_x E_\psi\), the required vector--Jacobian products through the noise predictor are Hessian--vector products through the energy.
We compute these by locally recomputing each transition with input differentiation enabled and checkpoint parameters frozen.
The clean predictions, reward-gradient directions, and auxiliary geometry supplied to the controller are detached during this
differentiation.
The calibrated pseudo-Huber matching loss is specified in \Cref{apdx:kuka-training}.

\subsubsection{Collision reward, controller, and training}\label{apdx:kuka-training}

\paragraph{Differentiable collision reward.}
We divide each trajectory segment into four subsegments by interpolating the joint angles, then use forward kinematics
to compute the positions and orientations of the robot links at each sampled configuration.
We approximate each link with four spheres to measure distances between arms and to obstacles and the table\footnote{
    Distances are estimated using a sampled-axis convex-hull approximation with a \(2\,\mathrm{mm}\) offset.
}.
Intended contact between the fixed mounting link and the table is excluded.

For clearances \(d\), safety margin \(m\), and activation width \(\delta=0.01\,\mathrm{m}\), we define
\begin{equation}
    p(d;m)=\left[\frac{m+\delta-d}{\delta}\right]_+^2,
    \qquad
    P(d;m)=\operatorname{mean}p(d;m)+\max p(d;m),
    \label{eq:kuka-clearance-penalty}
\end{equation}
where \([x]_+=\max(x,0)\).
The mean penalises violations across the sampled points, while the maximum emphasises the worst violation for each arm or arm pair.
For an assembled trajectory \(Y=\varphi(X)=q\), the reward combines inter-arm, obstacle, table, and self-collision penalties:
\begin{equation}
    r_\kappa(Y)=-\left[
    \dfrac{1}{\binom N2}\sum_{i<j}P(d^{ij}_{\mathrm{arm}};m)
    +\dfrac1N\sum_i\left(
    2P(d^i_{\mathrm{obs}};0)
    +\tfrac12P(d^i_{\mathrm{table}};0)
    +2P(d^i_{\mathrm{self}};0)
    \right)\right].
    \notag
    \label{eq:kuka-reward}
\end{equation}
The clearances are evaluated on \(Y\) in scene \(\kappa\); these arguments are suppressed on the right-hand side.
Penalties activate within 1\,cm of contact; the additional inter-arm margin \(m\) is zero in these experiments.
Start and goal configurations are fixed exactly.
We use no endpoint, path-length, or curvature penalty.
The reward and control cost do not guarantee collision-free, smooth, or dynamically feasible motion.

\paragraph{Shared control.}
The controller follows the same implementation detailed in \cref{apdx:maze-axial-arch}.
Inputs include the noisy state, detached clean prediction, normalised reward gradient, endpoint and base conditions, and local kinematic features.
During cross-arm attention, the controller uses the relative positions of each pair of arms, their clearance, goals, base positions, and motion to adjust how much attention they give each other.
Consequently, the same controller can process an additional arm without changing parameter dimensions.

The controller is parameterised using the Tweedie reward gradient, \method{}+TRG; see \cref{eq:cmds_trg}, and uses the pseudo-Huber loss in \cref{apdx:maze-robust-loss}.
Detailed parameters are provided in \cref{tab:kuka-training}.

\begin{table}[tb]
\centering\small
\renewcommand{\arraystretch}{1.12}
    \begin{tabular}{@{}p{0.40\linewidth}p{0.54\linewidth}@{}}
    \toprule
    Setting & Final configuration \\
    \midrule
    Trajectory / reverse steps & 56 knots / 10 DDIM steps \\
    Stochasticity / prior conditioning & \(\eta=1\) / \(\omega_{\mathrm{cfg}}=2\) \\
    Controller & Width 344, 4 heads, 2 axial blocks \\
    Feed-forward / pair-bias hidden width & \(4\times\) controller width / 32 \\
    Optimiser & Adam, \((\beta_1,\beta_2)=(0.9,0.999)\), \(\epsilon=10^{-8}\) \\
    Updates / batch / microbatch & 20,000 / 256 joint queries / 128 \\
    Learning rate & 400-step warm-up to \(10^{-4}\), cosine decay to \(10^{-5}\) \\
    EMA / gradient-norm clipping & 0.99 / 0.5 \\
    Reward coefficient / active matching steps & \(\beta=40\) / all 9 stochastic transitions \\
    Robust calibration & First 500 updates; 95th percentile; floor \(10^{-3}\) \\
    Primary / supplementary training seed & 0 / 1; final-update EMA weights \\
    \bottomrule
    \end{tabular}
\caption{\textbf{KUKA controller-training configuration.} 
    Values are taken from the final saved training configuration; the priors remain frozen.
}
\label{tab:kuka-training}
\end{table}

\subsubsection{Baselines and PBDM refinement}\label{apdx:kuka-baselines}

\paragraph{Independent and joint priors.}
Single-arm KUKA samples each arm independently using the same frozen prior as \method{}; Dual KUKA uses the released joint prior.
Single-arm sampling uses \(\eta=1\); Dual KUKA retains its native \(\eta=0\) sampler. 
We also evaluated single-arm sampling with $\eta=0$, but it did not improve performance.
Except for PBDM, methods return the first candidate that passes the common collision and endpoint checks.

\paragraph{Direct collision guidance.}
Guided single-arm and joint priors use the same collision reward as \method{}, with gradient scale 0.6 on the first nine DDIM transitions.
The reward gradient is evaluated at the predicted clean path, without backpropagating through the prior.
The guidance strength was selected from \(\{0.1,0.3,0.6,1.0,1.7\}\) by validation success across \(n_{\mathrm{cand}}\in\{1,2,3,4\}\) initial noises, then fixed for all tests.

\paragraph{PBDM selection and refinement.}
PBDM uses the released joint prior and planning procedure \citep{luo2024potential}.
The release defaults to 20 initial candidates with refinement disabled; our main PBDM results enable its optional refinement with the stated candidate budget \(n_{\mathrm{cand}}\) equal to 1 or 4, matching the number of initial noise samples drawn by the other methods.

\begin{enumerate}[leftmargin=*]
    \item Generate \(n_{\mathrm{cand}}\) candidates and return the first accepted by PBDM's own checker.
    \item If none passes, try the first \(\min(n_{\mathrm{cand}},6)\)  candidates in order, allowing up to \(R=5\) repair attempts each.
    Repair partially re-noises a path and applies three DDIM steps.
    Return the first accepted repaired path, or report failure.
    \item Evaluate the returned path with our common strict checker.
    Success requires passing both checks; strict rejection does not trigger another candidate or repair.
\end{enumerate}
The budget \(n_{\mathrm{cand}}\) excludes repair attempts, but planning time includes all repair and checking.

\paragraph{Single-arm priors with adapted PBDM.}
This baseline uses independent single-arm priors.
During repair, the released partial denoiser runs separately for each arm.
Joint segments are accepted using PBDM's rules, with three repair attempts and three denoising steps per attempt.
The adaptation is applied to three to five arms.

\subsubsection{Strict evaluation, timing, and uncertainty} \label{apdx:kuka-evaluation}

\paragraph{Strict success.}

An independent PyBullet mesh checker tests inter-arm, obstacle, table, and non-adjacent self collisions.
Each trajectory segment has at least eight subdivisions, with joint increments no larger than 0.04 radians.
Success requires no detected collisions and finite joint angles within joint limits, with no additional inter-arm clearance margin.

\paragraph{Planning time.}
Each task generates its \(n_{\mathrm{cand}}\) candidates as one batch on a GH200 GPU using warm CUDA graphs.
Timing includes scene setup, generation, transfer, checking, and any repair through the final decision.
It excludes model and asset loading, graph preparation, diagnostics, logging, and clean-up.

\begin{figure}[p]
\centering
\includegraphics[width=\linewidth]{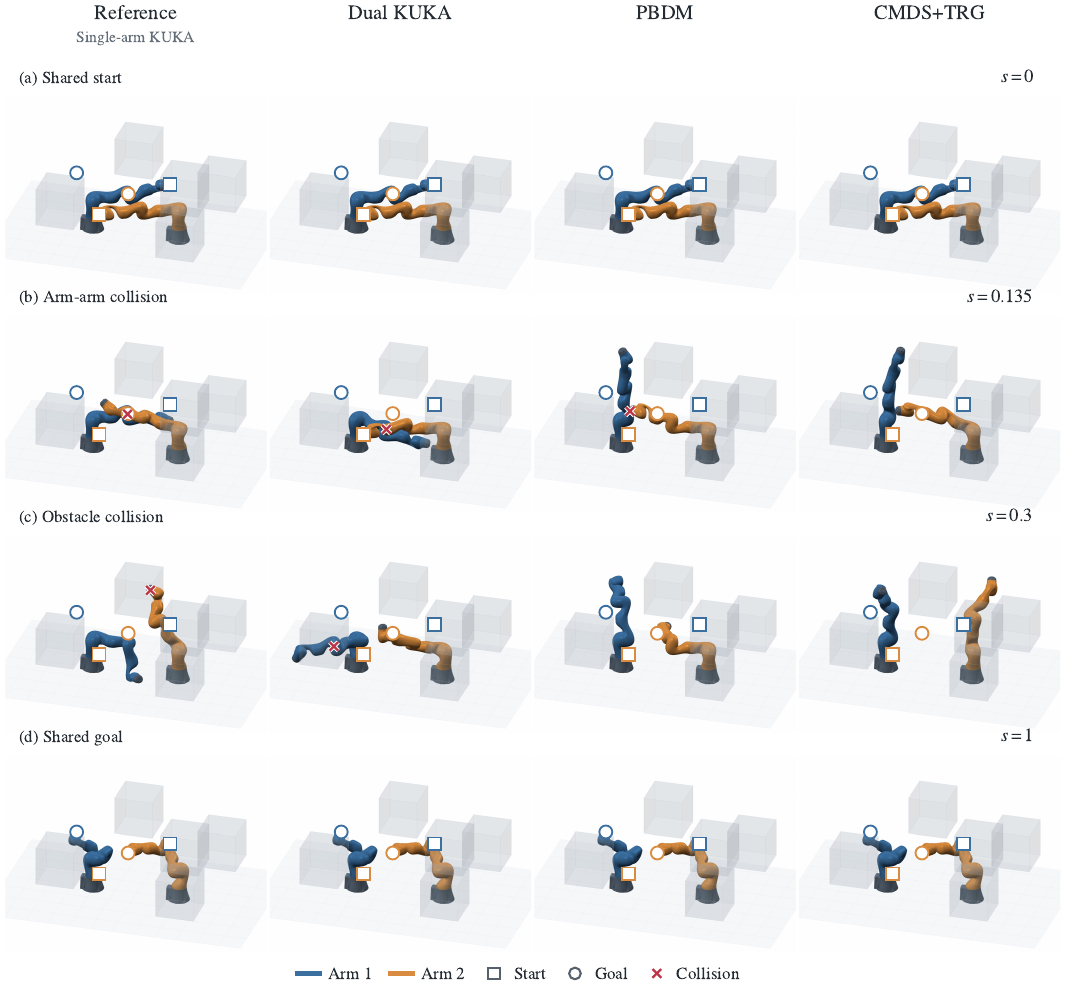}
\caption{\textbf{Synchronized KUKA configurations.} Reference (two independent single-arm priors), Dual KUKA, PBDM and CMDS+TRG on the same task at common path progress $s\in\{0, 0.135, 0.3, 1\}$: shared start, arm-arm collision, obstacle collision, shared goal. Squares mark starts, circles goals, red crosses collisions. Only CMDS+TRG remains collision-free.
}
\label{fig:kuka-storyboard}
\end{figure}

\begin{figure}[p]
\centering
\includegraphics[width=\linewidth]{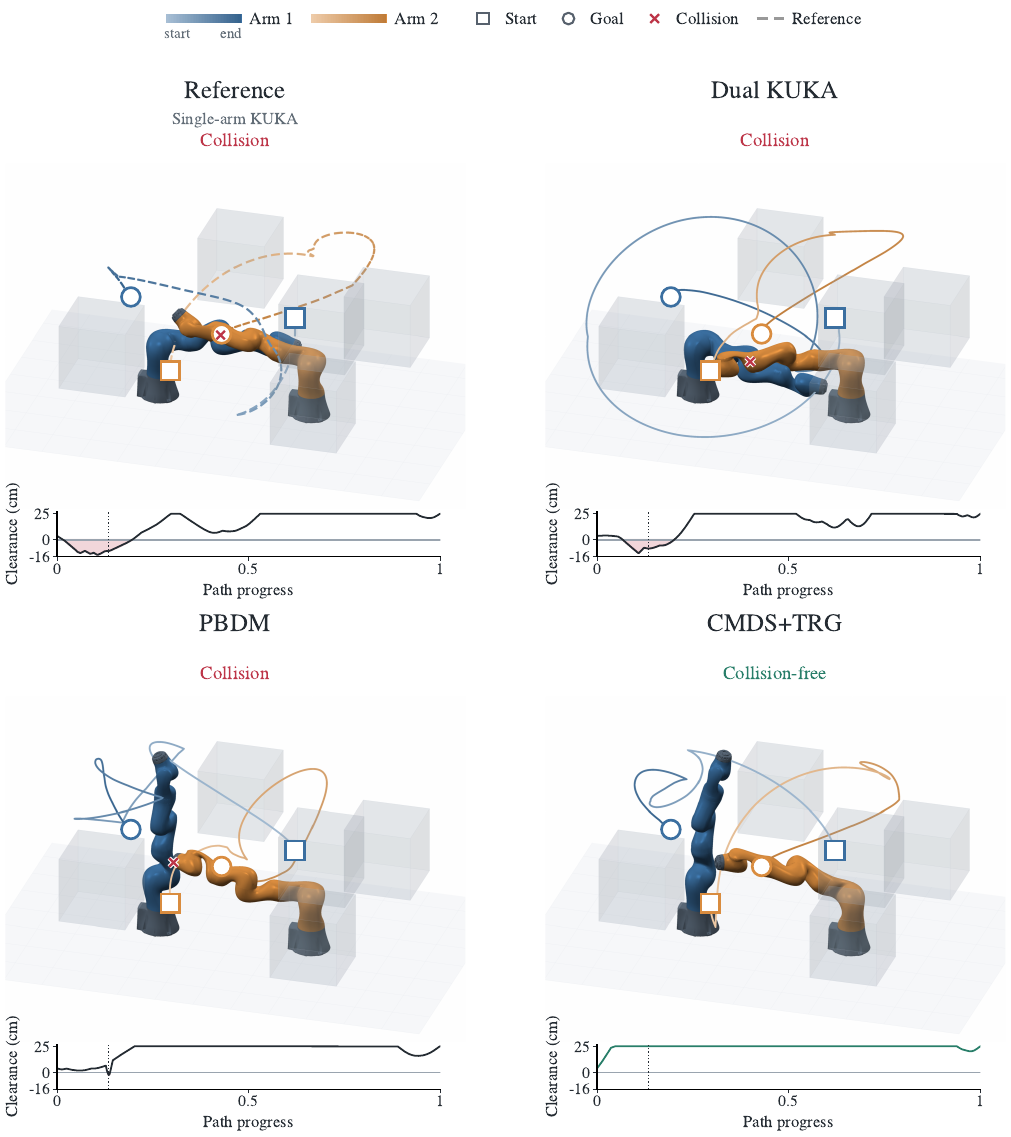}
\caption{\textbf{KUKA poses and inter-arm clearance.} Reference (two independent single-arm priors), Dual KUKA, PBDM and CMDS+TRG on the same held-out task at normalized path progress $s=0.135$ (dotted line), with the inter-arm clearance along each path below. Wrist paths fade from start to goal; squares mark starts, circles goals, red crosses collisions at the shown pose. Shaded negative clearance indicates collision; only CMDS+TRG stays positive over the full path.
}
\label{fig:kuka-clearance}
\end{figure}

\subsection{Goal-Directed Multi-Person Motion Coordination}\label{apdx:mdm_multi_planning}

\begin{takehome}
\textbf{Key Takeaways:}
\textbf{Frozen single-person priors can be coordinated into multi-person motion.}
A single shared control coordinates four independently text-conditioned MDM processes to reach prescribed destinations, avoid inter-person collisions, and maintain kinematic validity, without paired interaction data. \textbf{Amortisation makes coordination efficient.}
\method{} achieves $90.0\%$ joint success compared with $75.0\%$ for PCD++, while requiring $0.54$\,s per assembly versus $145.34$\,s for PCD++.
\end{takehome}
We coordinate four text-conditioned human motions using a shared control and a frozen single-person prior.
People cross a common region towards opposite destinations, requiring collision avoidance and consistent articulated motion.

\subsubsection{Task, representation, and frozen MDM prior}
\label{app:motion_task}
Each person \(i\in\{1,\ldots,N\}\), with \(N=4\), receives a caption \(c_i\), a planar start \(s_i\in\mathbb R^2\), an initial yaw \(\psi_i\), and a destination \(g_i=-s_i\).
Starts are arranged on a randomly rotated ring with nominal radius \(2.147\,\mathrm m\), radial jitter \(\pm0.15\,\mathrm m\), and angular jitter \(\pm10^\circ\).
Yaw points towards the destination with an additional \(\pm10^\circ\) jitter.
Scene conditions remain fixed during generation, and waiting and lateral detours are allowed.
A motion contains \(F=120\) frames at \(20\,\mathrm{Hz}\), represented by \(263\) normalised HumanML features per frame.

\paragraph{Notation and assembly.}
We retain the discrete state and indexing convention of Algorithm~1: the joint component state is
\(X_\ell=(X_\ell^1,\ldots,X_\ell^N)\in\mathbb R^{N\times F\times263}\), the reverse-time grid is \(\zeta_\ell=K-1-\ell\), and the generated motion is \(X_K\), with \(K=50\).
Thus \(X_0\) denotes Gaussian noise, rather than a clean motion, and \(\ell\) is distinct from the physical frame index \(f\in\{0,\ldots,F-1\}\).
We use \(\tau\) for possibly fractional frame indices on the densified collision-checking grid.
Each reverse transition updates the entire motion clip; the terminal reward is evaluated after denoising and can inspect all physical frames.
The task condition is \(\kappa=\{(c_i,s_i,g_i,\psi_i)\}_{i=1}^N\), together with the active-person and frame masks.
A deterministic geometric decoder \(D_i(\cdot;s_i,\psi_i)\) denormalises person \(i\)'s features, reconstructs \(22\) joints, and applies the scene transform, enforcing \(p_i(0)=s_i\) by construction.
For a fixed task, the assembly retains both the native features and the decoded world geometry:
\begin{equation}
    \varphi(X)
    \coloneqq\bigl(X,J_\kappa(X)\bigr),
    \qquad
    J_\kappa(X)
    \coloneqq\bigl(D_i(X^i;s_i,\psi_i)\bigr)_{i=1}^N.
    \label{eq:motion_assembly}
\end{equation}
We suppress the fixed scene dependence of \(\varphi\) in the notation.
Retaining native features is necessary for the rotation, displacement, and contact-consistency costs, which cannot be evaluated from world-space joint positions alone.
State variables, noise, and optimisation are restricted to active people and frames; padded coordinates remain zero throughout.
For motion-feature tensors, \(\|\cdot\|_2\) denotes the Euclidean norm after vectorisation, without averaging over people, frames, or features.

\paragraph{Frozen denoising and reverse transitions.}
We use the released HumanML MDM checkpoint trained for \(50\) diffusion steps, together with its frozen CLIP ViT-B/32 text encoder.
Each person's frozen denoiser is conditioned on its own caption, not on the other people or the scene's starts and goals, with classifier-free guidance scale \(2.5\).
This is a separately trained native \(50\)-step model, not a respaced \(1{,}000\)-step checkpoint.
Sampling retains its cosine schedule, fixed posterior variance, and unclipped clean predictions.
Write \(\alpha_t=1-\beta_t\), \(\bar\alpha_t=\prod_{j=0}^t\alpha_j\), and \(\bar\alpha_{-1}=1\) for the saved native schedule.
For a candidate joint state \(Z=(Z^1,\ldots,Z^N)\), define
\begin{equation}
    \begin{aligned}
        \mathcal D_\ell^i(Z^i)
        &\coloneqq\frac{Z^i-\sqrt{1-\bar\alpha_{\zeta_\ell}}\,
            \epsilon_{\mathrm b}^i(Z^i,\zeta_\ell)}
            {\sqrt{\bar\alpha_{\zeta_\ell}}},
        &&0\leq\ell<K,\\
        \mathcal D_\ell(Z)
        &\coloneqq\bigl(\mathcal D_\ell^1(Z^1),\ldots,
            \mathcal D_\ell^N(Z^N)\bigr),
        &\qquad\mathcal D_K(Z)&\coloneqq Z.
    \end{aligned}
    \label{eq:motion-baseline-tweedie}
\end{equation}
Thus \(\hat X_{1\mid\ell}=\mathcal D_\ell(X_\ell)\) is the clean-motion estimate, and \(\hat Y_{1\mid\ell}=\varphi(\hat X_{1\mid\ell})\) is its assembly.
MDM predicts the clean state directly; the expression above is its equivalent epsilon parameterisation, with classifier-free guidance included in \(\epsilon_{\mathrm b}^i\).
The denoising map \(\mathcal D_\ell\) is distinct from the geometric decoder \(D_i\).
Although Algorithm~1 uses DDIM, the reverse maps here are MDM's native DDPM transitions:
\begin{equation}
    F_\ell^{\mathrm b}(X_\ell,\Xi_\ell)
    \coloneqq\mu_{\mathrm b,\ell}(X_\ell)+\varsigma_\ell\Xi_\ell,
    \qquad
    \begin{cases}
        \varsigma_\ell>0,&0\leq\ell<K-1,\\
        \varsigma_{K-1}=0.&
    \end{cases}
    \label{eq:motion-baseline-transition}
\end{equation}
Here \(\mu_{\mathrm b,\ell}\) is the frozen classifier-free-guided DDPM posterior mean, acting component-wise.
With \(\epsilon_{\mathrm b}(Z,t)=(\epsilon_{\mathrm b}^i(Z^i,t))_{i=1}^N\), the native mean and effective variance are
\begin{equation}
    \begin{aligned}
        \mu_{\mathrm b,\ell}(Z)
        &=\frac{1}{\sqrt{\alpha_{\zeta_\ell}}}
            \left[
                Z-\frac{\beta_{\zeta_\ell}}{\sqrt{1-\bar\alpha_{\zeta_\ell}}}
                \epsilon_{\mathrm b}(Z,\zeta_\ell)
            \right],\\
        \varsigma_\ell^2
        &=\frac{\beta_{\zeta_\ell}
            (1-\bar\alpha_{\zeta_\ell-1})}
            {1-\bar\alpha_{\zeta_\ell}}.
    \end{aligned}
    \label{eq:motion_ddpm_coefficients}
\end{equation}
The final transition has \(\zeta_{K-1}=0\) and is deterministic.
No denoiser evaluation or additional grid point \(\zeta_K\) is needed for \(\mathcal D_K=\mathrm{Id}\).

\paragraph{Controlled transition.}
We use \(w_{\theta,\ell}\) for the normalised discrete control, consistently with Algorithm~1.
For the stochastic transitions, \method{} uses
\begin{equation}
    \begin{aligned}
        X_{\ell+1}
        &=F_\ell^\theta(X_\ell,\Xi_\ell)\\
        &\coloneqq\mu_{\mathrm b,\ell}(X_\ell)
            +\varsigma_\ell w_{\theta,\ell}(X_\ell)
            +\varsigma_\ell\Xi_\ell,
        \qquad \Xi_\ell\sim\mathcal N(0,I),
        \quad 0\leq\ell<K-1.
    \end{aligned}
    \label{eq:motion_transition}
\end{equation}
Unlike the epsilon-residual parameterisation in Algorithm~1, the motion controller outputs \(w_{\theta,\ell}\) directly, supplying the mean correction \(\varsigma_\ell w_{\theta,\ell}\).
For \(\varsigma_\ell>0\), this is the same normalisation
\(w_{\theta,\ell}=(F_\ell^\theta-F_\ell^{\mathrm b})/\varsigma_\ell\), with both maps evaluated at the same state and noise.
The final update is \(X_K=\mu_{\mathrm b,K-1}(X_{K-1})=\mathcal D_{K-1}(X_{K-1})\), with no learned correction, control-energy term, or matching loss at that transition.
In particular, the positive-noise terminal modification used for DDIM in Algorithm~1 is not used here.

\paragraph{DDPM--DDIM control correspondence.}
%
Both samplers use the same control normalisation, but different network outputs.
In Algorithm~1, an epsilon residual \(\delta\epsilon_{\theta,\ell}\) induces the DDIM mean shift \(\gamma_\ell\delta\epsilon_{\theta,\ell}\), giving \(w_{\theta,\ell}=(\gamma_\ell/\varsigma_\ell)\delta\epsilon_{\theta,\ell}\) for that sampler.
For the native DDPM kernel, the corresponding signed epsilon coefficient and equivalent residual are
\begin{equation}
    \begin{aligned}
        \gamma_\ell^{\mathrm{DDPM}}
        &\coloneqq-\dfrac{\beta_{\zeta_\ell}}
            {\sqrt{\alpha_{\zeta_\ell}}\sqrt{1-\bar\alpha_{\zeta_\ell}}},\\
        \delta\epsilon_{\theta,\ell}^{\mathrm{equiv}}
        &\coloneqq\dfrac{\varsigma_\ell}{\gamma_\ell^{\mathrm{DDPM}}}
            w_{\theta,\ell},
        \qquad 0\leq\ell<K-1.
    \end{aligned}
    \label{eq:motion_ddpm_ddim_control}
\end{equation}
The sign is retained: \(\gamma_\ell^{\mathrm{DDPM}}\delta\epsilon_{\theta,\ell}^{\mathrm{equiv}}=\varsigma_\ell w_{\theta,\ell}\).
The motion implementation requires no such conversion because the controller outputs \(w_{\theta,\ell}\) directly.
For adjacent native timesteps, the same denoiser and schedule, and no clipping, stochastic DDIM with \(\eta=1\) reproduces the DDPM mean and posterior variance on stochastic transitions.
This does not identify the complete rollouts: Algorithm~1 retains noise at its modified DDIM endpoint, whereas MDM uses its original control-free deterministic endpoint.
Thus the adaptation changes the sampling kernel, output parameterisation, and endpoint treatment, rather than the underlying stochastic-control formulation.

\subsubsection{Cooperative control architecture}
\label{app:motion_architecture}
The controller forms one token per person and physical frame.
Its motion input concatenates the current noisy features, the prior's clean-motion estimate, and the decoded world-space joints:
\begin{equation}
    \begin{aligned}
        \hat J_{\ell,i,f}
        &\coloneqq D_i(\hat X_{1\mid\ell}^{\,i};s_i,\psi_i)_f,\\
        z_{\ell,i,f}
        &\coloneqq\bigl[
            (X_\ell^i)_f,\,
            (\hat X_{1\mid\ell}^{\,i})_f,\,
            \operatorname{vec}(\hat J_{\ell,i,f})
        \bigr]
        \in\mathbb R^{263+263+66}.
    \end{aligned}
    \label{eq:motion_token}
\end{equation}
The \(66\) skeleton coordinates represent the \(22\) joints in metres.

We project \(z_{\ell,i,f}\) to width \(344\) and add separate projections of the \(512\)-dimensional CLIP embedding, the static condition
\([s_i,g_i,\sin\psi_i,\cos\psi_i,F/196]\in\mathbb R^7\), and \(32\)-dimensional sinusoidal embeddings of the physical frame \(f\) and native diffusion index \(\zeta_\ell\), then apply SiLU.
A local temporal encoder attends to each person's entire clip.
Two subsequent blocks alternate temporal attention within a person, attention across people at the same frame, and a feed-forward network.
Communication is aligned in physical time, with whole-motion context supplied by the temporal passes.
Weights are shared across people, making the controller equivariant to a consistent permutation of people and their conditions.

\paragraph{Geometric communication.}
Person attention receives an additive bias computed from predicted planar root geometry.
At a fixed reverse transition and physical frame, suppress these indices and write
\(\Delta p_{ij}=\hat p_j-\hat p_i\),
\(d_{ij}^{\mathrm{root}}=\|\Delta p_{ij}\|_2\), and
\(\Delta v_{ij}=\hat v_j-\hat v_i\), where velocities use \(20\,\mathrm{Hz}\) backward differences.
The bias is
\begin{equation}
    b_{ij}
    =\operatorname{MLP}_{8\to32\to4}\!\left(
        \left[
            \Delta p_{ij},\,
            d_{ij}^{\mathrm{root}},\,
            \Delta v_{ij},\,
            \frac{-\Delta p_{ij}^{\top}\Delta v_{ij}}
                {\max(d_{ij}^{\mathrm{root}},10^{-4})},\,
            g_j-g_i
        \right]
    \right).
    \label{eq:motion_geometry_bias}
\end{equation}
Its four outputs are added to the corresponding attention-head logits \(q_i^\top k_j/\sqrt{86}\).
The same bias is used in both interaction blocks.
The root distance \(d_{ij}^{\mathrm{root}}\) is distinct from the signed body-capsule clearance defined below.

\paragraph{Control output.}
Separate linear heads read the local encoder output \(H_{\mathrm{local}}\) and the second interaction block output \(H_{\mathrm{joint}}\):
\begin{equation}
    \begin{aligned}
        w_{\theta,\ell}
        ={}&W_{\mathrm{local}}H_{\mathrm{local}}+b_{\mathrm{local}}\\
        &+W_{\mathrm{joint}}H_{\mathrm{joint}}+b_{\mathrm{joint}}
        \in\mathbb R^{N\times F\times263}.
    \end{aligned}
    \label{eq:motion_control_output}
\end{equation}
This \(4{,}863{,}042\)-parameter control supplies joint awareness to the separate denoising processes.
Both output heads and the final layer of the geometric-bias MLP are zero-initialised, so the initial controller leaves the frozen sampling process unchanged.

\subsubsection{Assembly reward}
\label{app:motion_reward}
For a completed joint motion \(X\), write \(Y=\varphi(X)\).
The total cost combines collision avoidance, goal attainment, directional progress, and physical consistency:
\begin{equation}
    \begin{aligned}
        C_\kappa(X)={}&
            \lambda_{\mathrm{col}}E_{\mathrm{col}}
            +\lambda_{\mathrm{goal}}E_{\mathrm{goal}}
            +\lambda_{\mathrm{prog}}E_{\mathrm{prog}}\\
        &+\lambda_{\mathrm{bone}}E_{\mathrm{bone}}
            +\lambda_{\mathrm{FK}}E_{\mathrm{FK}}
            +\lambda_{\mathrm{disp}}E_{\mathrm{disp}}\\
        &+\lambda_{\mathrm{contact}}E_{\mathrm{contact}}
            +\lambda_{\mathrm{floor}}E_{\mathrm{floor}}
            +\lambda_{\mathrm{speed}}E_{\mathrm{speed}}.
    \end{aligned}
    \label{eq:motion_cost}
\end{equation}
All terms are evaluated using \(X\) and \(\varphi(X)\), with dependence on the fixed task suppressed on the right-hand side.
We distinguish \(C_\kappa^{\mathrm{train}}\), used to train \method{}, from \(C_\kappa^{\mathrm{base}}\), used for baseline selection and guidance.
They use identical terms and normalisations, but different goal weights: \(\lambda_{\mathrm{goal}}=50\) for control training and \(\lambda_{\mathrm{goal}}=1\) for the baselines.
The corresponding assembly-level rewards are
\begin{equation}
    r_\kappa^{\mathrm{train}}(\varphi(X))
    \coloneqq-C_\kappa^{\mathrm{train}}(X),
    \qquad
    r_\kappa^{\mathrm{base}}(\varphi(X))
    \coloneqq-C_\kappa^{\mathrm{base}}(X).
    \label{eq:motion_rewards}
\end{equation}
The subscripted cost weights are distinct from the outer SOC reward multiplier \(\lambda\), specified in \cref{app:motion_training}.
Their recorded values are reported in \cref{tab:motion_settings}.
All geometric distances below are in metres.

\paragraph{Body collision.}
Each person is represented by \(15\) fixed-radius capsule primitives, including a spherical head primitive.
For capsule centreline segment \(S_{ia}(\tau;Y)\) and radius \(\rho_a\), the signed surface clearance is
\begin{equation}
    d_{ij}^{ab}(\tau;Y)
    \coloneqq\operatorname{dist}\bigl(S_{ia}(\tau;Y),S_{jb}(\tau;Y)\bigr)
        -\rho_a-\rho_b.
    \label{eq:motion_clearance}
\end{equation}
We divide each native frame interval into five linear subintervals and retain the final endpoint, giving the synchronised grid
\(\mathcal T=\{n/5:n=0,\ldots,5(F-1)\}\), with
\(|\mathcal T|=5(F-1)+1=596\).
The collision penalty is
\begin{equation}
    E_{\mathrm{col}}
    =\frac{1}{\binom{N}{2}|\mathcal T|}
        \sum_{i<j}\sum_{\tau\in\mathcal T}\sum_{a,b=1}^{15}
        \left(\frac{[0.03-d_{ij}^{ab}(\tau;Y)]_+}{0.05}\right)^2,
    \qquad [z]_+\coloneqq\max(z,0).
    \label{eq:motion_collision}
\end{equation}
For \(N=4\), the denominator is \(6|\mathcal T|\), as in the implemented cost.
The penalty averages over person pairs and dense times, sums all \(225\) capsule pairs, and becomes zero at \(3\,\mathrm{cm}\) clearance.
This soft-cost margin is distinct from both PCD++'s nonnegative-clearance constraint and evaluation's \(5\,\mathrm{mm}\) penetration tolerance.
\Cref{tab:motion_capsules} specifies the primitive endpoints and radii.

\begin{table}[t]
    \centering
    \small
    \caption{
        Body-capsule geometry for collision costs and checks.
        Endpoints are zero-based HumanML joint indices; the repeated head endpoint defines a sphere.
    }
    \label{tab:motion_capsules}
    \setlength{\tabcolsep}{8pt}
    \begin{tabular}{@{}llr@{}}
        \toprule
        \toprule
        Primitive & Joint endpoints & Radius (m) \\
        \midrule
        Pelvis & \((1,2)\) & 0.120 \\
        Torso & \((0,9)\) & 0.150 \\
        Shoulder span & \((16,17)\) & 0.100 \\
        Upper torso--head & \((9,15)\) & 0.100 \\
        Head & \((15,15)\) & 0.110 \\
        Upper arms & \((16,18),\ (17,19)\) & 0.055 \\
        Forearms & \((18,20),\ (19,21)\) & 0.045 \\
        Thighs & \((1,4),\ (2,5)\) & 0.085 \\
        Lower legs & \((4,7),\ (5,8)\) & 0.065 \\
        Feet & \((7,10),\ (8,11)\) & 0.055 \\
        \bottomrule
    \end{tabular}
\end{table}

\paragraph{Destination and progress.}
Let \(p_i(f;Y)\) be person \(i\)'s planar root position, and set the goal tolerance to \(\varepsilon_{\mathrm g}=0.5\,\mathrm m\).
On the final five frames, \(\mathcal W_i=\mathcal W=\{F-5,\ldots,F-1\}=\{115,\ldots,119\}\), we use
\begin{align}
    E_{\mathrm{goal}}
    &=\frac1N\sum_{i=1}^N
        \frac{\min_{f\in\mathcal W_i}\|p_i(f;Y)-g_i\|_2^2}
            {\varepsilon_{\mathrm g}^2},
    \label{eq:motion_goal}\\
    \bar q_i
    &=\frac1{|\mathcal W_i|}\sum_{f\in\mathcal W_i}
        \frac{(p_i(f;Y)-s_i)^\top(g_i-s_i)}{\|g_i-s_i\|_2^2},
    \qquad
    E_{\mathrm{prog}}
    =\frac1N\sum_{i=1}^N[0.9-\bar q_i]_+^2.
    \label{eq:motion_progress}
\end{align}
People may arrive at different frames within the window.
Progress is the window-mean displacement along the start-to-goal direction; its threshold shapes training but is not an extra success condition.

\paragraph{Physical consistency.}
Let \(Q_{ifj}\in\mathbb R^3\) be canonical decoded joints and \(J_{ifj}\) their world-space counterparts.
Write \(L_{ifb}\) for the \(21\) bone lengths, \(L_b^\star\) for fixed audit references, and \(Q^{\mathrm{FK}}\) for joints reconstructed from the native 6D rotations using those reference lengths.
For denormalised displacement channels \(V_{ifj}\), root-local rotation \(A_{if}\), and \(\Delta Q_{ifj}=Q_{i,f+1,j}-Q_{ifj}\), the costs are
\begin{align}
    E_{\mathrm{bone}}
    &=\left\langle(L/L^\star-1)^2\right\rangle,
    &E_{\mathrm{FK}}
    &=\frac{\langle\|Q-Q^{\mathrm{FK}}\|_2^2\rangle}{3(0.05)^2},
    \label{eq:motion_bone_fk}\\
    E_{\mathrm{disp}}
    &=\frac{\langle\|V-A\Delta Q\|_2^2\rangle}{3(0.05)^2},
    &E_{\mathrm{contact}}
    &=\langle(H-H^\star)^2\rangle,
    \label{eq:motion_displacement_contact}\\
    E_{\mathrm{floor}}
    &=\frac{\langle[-J_y]_+^2\rangle}{(0.05)^2},
    &E_{\mathrm{speed}}
    &=\left\langle[\|20\Delta Q\|_2/v_{\max}-1]_+^2\right\rangle.
    \label{eq:motion_floor_speed}
\end{align}
Angle brackets average over people, valid native frames or intervals, and the relevant bones or joints.
The factors of three make vector errors coordinate-wise means.
Encoded displacement is compared per frame, without multiplying it by the frame rate.
Here \(v_{\max}\) is the speed threshold in the soft cost.
For denormalised contact channels \(H\), corresponding to joints \(7,10,8,11\), the soft reference is
\(H^\star=\sigma((0.08-h)/0.02)\,\sigma((0.2-v)/0.05)\), using foot height \(h\), speed \(v\), and logistic sigmoid \(\sigma\).
Root speed and jerk are additional hard evaluation checks.

\subsubsection{Training by adjoint matching}
\label{app:motion_training}
Only controller parameters are updated.
Training scenes are generated online, with captions drawn from five families: walking, jogging, high-knee marching, forward jumping, and crouched walking.
Each family has three phrasings.
A quarter of scenes use one shared action family; the remainder sample family assignments conditioned on having at least two distinct families.
Phrasings are sampled per person.
Training uses generated trajectories, without paired interaction data.

\paragraph{Control objective and replay.}
The motion experiment uses reward strength \(\lambda=100\) and includes all stochastic transitions in the matched set \(\mathcal K=\{0,\ldots,K-2\}\).
Its discrete stochastic-control objective is
\begin{equation}
    \mathcal J(\theta)
    =\mathbb E\!\left[
        \lambda C_\kappa^{\mathrm{train}}(X_K)
        +\frac12\sum_{\ell=0}^{K-2}
            \|w_{\theta,\ell}(X_\ell)\|_2^2
    \right].
    \label{eq:motion_objective}
\end{equation}
The expectation covers task conditions, initial noise, and rollout noise.
The norm sums over people, physical frames, and all \(263\) motion features, without averaging; \(\lambda\) scales only the terminal cost.
For a fixed task and state, the controlled and reference stochastic transitions have common covariance \(\varsigma_\ell^2I\) and mean difference \(\varsigma_\ell w_{\theta,\ell}\), giving conditional KL divergence \(\tfrac12\|w_{\theta,\ell}\|_2^2\).
With the same initial Gaussian distribution, the expected sum of these energies equals the discrete path-space KL relative to the frozen sampler at classifier-free guidance scale \(2.5\); the shared deterministic final map adds no conditional KL.
This Gaussian mean-shift interpretation does not cover arbitrary nonzero corrections at zero-noise transitions, including deterministic DDIM with \(\eta=0\).

As in Algorithm~1, set \(\bar\theta=\operatorname{stopgrad}(\theta)\), generate a controlled rollout without gradient tracking, and replay individual transitions backwards.
Suppressing the rollout superscript, the state adjoints satisfy
\begin{equation}
    \begin{aligned}
        a_K
        &=\lambda\nabla_{X_K} C_\kappa^{\mathrm{train}}(X_K),\\
        a_{K-1}
        &=\bigl[D_X\mu_{\mathrm b,K-1}(X_{K-1})\bigr]^\top a_K,\\
        a_\ell
        &=\bigl[D_X\mu_{\mathrm b,\ell}(X_\ell)
            +\varsigma_\ell D_Xw_{\bar\theta,\ell}(X_\ell)\bigr]^\top a_{\ell+1}\\
        &\quad+\bigl[D_Xw_{\bar\theta,\ell}(X_\ell)\bigr]^\top
            w_{\bar\theta,\ell}(X_\ell),
        \qquad \ell=K-2,\ldots,0.
    \end{aligned}
    \label{eq:motion_adjoint}
\end{equation}
The deterministic final transition remains in adjoint propagation, but contributes neither control energy nor a matching term.
Each transition is differentiated using a fresh state variable at the stored value \(X_\ell\), with \(\bar\theta\) fixed and the next-state adjoint detached.
The controller's clean-motion, world-joint, text, and geometry-derived observations are detached and held fixed during this state differentiation, while its explicit noisy-state input and the frozen-prior mean remain differentiable.
The recursion follows these observation-detach conventions, rather than differentiating through the recomputation of every controller observation; graphs are released between transitions.

\paragraph{Calibrated robust matching.}
Training regresses against detached next-state adjoints using
\begin{equation}
    e_{\theta,\ell}
    \coloneqq w_{\theta,\ell}\bigl(\operatorname{stopgrad}(X_\ell)\bigr)
        +\varsigma_\ell\operatorname{stopgrad}(a_{\ell+1}),
    \qquad \ell\in\mathcal K.
    \label{eq:motion_matching_residual}
\end{equation}
The control is evaluated with the same task and detached observations as in replay.
With \(D=N F\cdot263=4\cdot120\cdot263\) motion-feature coordinates, the robust matching loss is
\begin{equation}
    \mathcal L_{\mathrm{PH}}(\theta)
    =\mathbb E\!\left[
        \sum_{\ell=0}^{K-2}
            D\delta_\ell^2\left(
                \sqrt{1+\frac{\|e_{\theta,\ell}\|_2^2}
                    {D\delta_\ell^2}}-1
            \right)
    \right].
    \label{eq:motion_matching}
\end{equation}
The first \(500\) successful updates use squared matching and collect per-instance RMS residuals \(\|e_{\theta,\ell}\|_2/\sqrt D\).
Each \(\delta_\ell\) is then fixed to their \(95\)th percentile, floored at \(10^{-3}\).
With squared matching, current-policy rollouts, and all stochastic transitions included, the matching gradient agrees with direct differentiation of \cref{eq:motion_objective} under the same observation-detach conventions.
Pseudo-Huber matching downweights large residuals and generally changes that gradient; it leaves the terminal reward, adjoint recursion, and quadratic control energy unchanged.

\begin{table}[t]
    \centering
    \small
    \caption{
        Reward weights and controller training settings for the reported \(N=4\) experiment.
        Baselines share the cost normalisations but use a smaller goal weight.
    }
    \label{tab:motion_settings}
    \begin{tabular}{@{}ll@{}}
        \toprule
        \toprule
        Setting & Recorded value \\
        \midrule
        \((\lambda_{\mathrm{col}},\lambda_{\mathrm{goal}},\lambda_{\mathrm{prog}})\), training
            & \((100,50,1)\) \\
        \((\lambda_{\mathrm{col}},\lambda_{\mathrm{goal}},\lambda_{\mathrm{prog}})\), baselines
            & \((100,1,1)\) \\
        \((\lambda_{\mathrm{bone}},\lambda_{\mathrm{FK}},\lambda_{\mathrm{disp}})\)
            & \((500,5,5)\) \\
        \((\lambda_{\mathrm{contact}},\lambda_{\mathrm{floor}},\lambda_{\mathrm{speed}})\)
            & \((1,200,10)\) \\
        Outer SOC reward multiplier \(\lambda\)
            & \(100\), control training only \\
        Optimiser / updates / learning rate
            & Adam / \(15{,}000\) / constant \(10^{-4}\) \\
        Effective batch size
            & \(8\) scenes: microbatch \(2\), accumulation \(4\) \\
        Gradient-norm clipping / EMA decay
            & \(0.5\) / \(0.9995\) \\
        \bottomrule
    \end{tabular}
\end{table}

Raw and EMA checkpoints were validated during training.
Selection ranked joint success, then terminal cost plus control energy, preferring earlier steps on ties.
Evaluation uses raw step-\(15{,}000\) weights from one seed-\(0\) run.
Inference requires no reward-gradient or projection iterations.

\subsubsection{Baselines}\label{app:motion_baselines}
Baselines share the frozen prior, native DDPM sampler, classifier-free guidance scale, scenes, decoder, and audit.
Selection and guidance use \(C_\kappa^{\mathrm{base}}\), with weights
\((100,1,1,500,5,5,1,200,10)\) in the order of \cref{eq:motion_cost}.
These retain exactly the cost normalisations above; in particular, the baseline goal weight is \(1\), rather than the training value \(50\).

\paragraph{Best-of-256.}
Best-of-256 generates \(256\) independent complete four-person assemblies and returns the one with minimum baseline cost.
Reference samples the same frozen MDM component priors independently and is used as a qualitative reference.

\paragraph{DPS guidance.}
DPS differentiates the joint baseline reward of the clean-motion estimate through the frozen denoiser to the current noisy state:
\begin{equation}
    \begin{aligned}
        v_\ell
        &\coloneqq\nabla_{X_\ell}
            r_\kappa^{\mathrm{base}}\!\left(
                \varphi(\mathcal D_\ell(X_\ell))
            \right),\\
        \widetilde X_{\ell+1}
        &\coloneqq\operatorname{stopgrad}\!\left(
            F_\ell^{\mathrm b}(X_\ell,\Xi_\ell)
            +\eta_{\mathrm g}v_\ell
        \right).
    \end{aligned}
    \label{eq:motion_dps_update}
\end{equation}
DPS sets \(X_{\ell+1}=\widetilde X_{\ell+1}\).
DPS, PCD, and PCD++ all use \(\eta_{\mathrm g}=0.1\) at every transition, including the deterministic final one.
This coefficient directly multiplies the reward gradient: there is no \(\varsigma_\ell^2\) factor and no separate application of the control-training multiplier \(\lambda=100\).
All guidance gradients are with respect to state variables, with denoiser parameters frozen and no gradient propagation across outer diffusion steps.
Unlike the learned control in \cref{eq:motion_transition}, this constant guidance also changes the deterministic endpoint; the subsequent PCD projections are not Gaussian mean shifts either.
These baselines therefore do not inherit the learned controller's quadratic path-KL interpretation.

\paragraph{PCD and PCD++.}
The PCD baseline combines \cref{eq:motion_dps_update} with a root-goal projector enforcing the final-window destination condition.
The projector changes planar root increments and the corresponding stored joint-displacement channels while keeping headings, heights, and poses fixed.
It minimises normalised-feature change within this restricted parameterisation, but does not enforce collision clearance or the kinematic-validity audit.
PCD++ uses the same guidance update and replaces that projector with a joint optimisation of the proposed state.
All \(263\) normalised feature channels of each active person and frame are free variables in this solve.
Feasibility is checked on a freshly recomputed clean estimate at the candidate's next noise level, rather than on the clean estimate used for the preceding guidance update.
At the final transition, the candidate is already a clean motion and \(\mathcal D_K=\mathrm{Id}\).

\paragraph{Joint goal and collision constraints.}
Using the planar roots and signed capsule clearances defined above, set
\begin{equation}
    \begin{aligned}
        c_{\mathrm g,i}(Y)
        &\coloneqq\min_{f\in\mathcal W_i}
            \|p_i(f;Y)-g_i\|_2-\varepsilon_{\mathrm g},\\
        c_{\mathrm c,ijab\tau}(Y)
        &\coloneqq\delta_{\mathrm c}-d_{ij}^{ab}(\tau;Y),\\
        V_\kappa(Y)
        &\coloneqq\max\!\left\{
            0,\ \max_i c_{\mathrm g,i}(Y),\ 
            \max_{\substack{i<j,\,a,b\\\tau\in\mathcal T_{ij}}}
                c_{\mathrm c,ijab\tau}(Y)
        \right\}.
    \end{aligned}
    \label{eq:motion-pcdpp-constraints}
\end{equation}
Here \(\mathcal T_{ij}\) contains the checked times when both people are active; in the reported fixed-length experiment, \(\mathcal T_{ij}=\mathcal T\) for every pair.
We use \(\varepsilon_{\mathrm g}=0.5\,\mathrm m\) and \(\delta_{\mathrm c}=0\), so PCD++ requires nonnegative capsule clearance.
The feasible set is
\begin{equation}
    \mathcal F_\kappa
    \coloneqq\bigl\{
        Y:V_\kappa(Y)=0,\ p_i(0;Y)=s_i\ \text{for all }i
    \bigr\}.
    \label{eq:motion_feasible_set}
\end{equation}
Starts are already enforced by the decoder.
For a guided proposal \(\widetilde X_{\ell+1}\), the projection targets
\begin{equation}
    \underset{Z}{\operatorname{minimise}}\quad
        \tfrac12\|Z-\widetilde X_{\ell+1}\|_2^2
    \qquad\text{subject to}\qquad
        \varphi(\mathcal D_{\ell+1}(Z))\in\mathcal F_\kappa.
    \label{eq:motion-pcdpp-target}
\end{equation}
The distance is Euclidean in the active normalised motion features, not in decoded joint coordinates.
The implemented solver approximates this constrained problem; it does not certify a nearest feasible point.

\paragraph{Penalty and numerical guards.}
The inner objective uses the squared-hinge penalty
\begin{equation}
    \begin{aligned}
        \mathcal P_\kappa(Y;\omega_{\mathrm g},\omega_{\mathrm c})
        \coloneqq{}&\sum_i[c_{\mathrm g,i}(Y)+\omega_{\mathrm g}]_+^2\\
        &+\sum_{i<j}\sum_{\tau\in\mathcal T_{ij}}\sum_{a,b=1}^{15}
            [c_{\mathrm c,ijab\tau}(Y)+\omega_{\mathrm c}]_+^2.
    \end{aligned}
    \label{eq:motion-pcdpp-penalty}
\end{equation}
Unlike the soft collision cost, this penalty uses unnormalised sums of constraint violations.
The small interior guards tighten the penalty only; acceptance uses the original constraints in \cref{eq:motion-pcdpp-constraints}.
At inner iterate \(m\), let \(J^{(m)}\) denote the active decoded world joints in
\(\hat Y^{(m)}=\varphi(\mathcal D_{\ell+1}(Z^{(m)}))\), and let \(\epsilon_{\mathrm{mach}}\) be the machine epsilon of their data type.
The detached coordinate scale and guards are
\begin{equation}
    \begin{aligned}
        S_m
        &\coloneqq\operatorname{stopgrad}\!\left[
            \max\!\left(1,\|J^{(m)}\|_\infty,
                \max_i\|g_i\|_\infty\right)
        \right],\\
        \omega_{\mathrm c}^{(m)}
        &\coloneqq\max(10^{-7},64\epsilon_{\mathrm{mach}}S_m),\\
        \omega_{\mathrm g}^{(m)}
        &\coloneqq\min(10^{-3}\varepsilon_{\mathrm g},\omega_{\mathrm c}^{(m)}).
    \end{aligned}
    \label{eq:motion-pcdpp-guards}
\end{equation}
The infinity norms denote the largest absolute coordinate.
These guards are recomputed at every inner iterate and held fixed when differentiating that iterate's objective:
\begin{equation}
    \mathcal L_m(Z)
    \coloneqq\tfrac12\|Z-\widetilde X_{\ell+1}\|_2^2
        +\rho\,\mathcal P_\kappa\!\left(
            \varphi(\mathcal D_{\ell+1}(Z));
            \omega_{\mathrm g}^{(m)},\omega_{\mathrm c}^{(m)}
        \right).
    \label{eq:motion_pcdpp_inner_objective}
\end{equation}

\paragraph{Projection settings and failure handling.}
PCD++ permits at most \(M=100\) inner Adam updates per outer transition, with step size \(\eta_{\mathrm p}=0.01\), initial penalty \(\rho_0=10\), and cap \(\rho_{\max}=10^7\).
Adam uses \((\beta_1,\beta_2)=(0.9,0.999)\), \(\epsilon=10^{-8}\), and zero weight decay, with fresh optimiser state at every outer transition.
Every five inner updates, the penalty is multiplied by \(10\), up to its cap, if the maximum violation has not decreased by more than \(5\%\) since the previous reference check.
The solver checks all \(15\times15\) capsule pairs for each pair of people at the \(596\) dense times, and checks goals in each person's final five native frames.
It accepts the first verified feasible iterate, with at most \(M\) updates and \(M+1\) checks per transition.
The accepted state is \(Z^{(m)}\), not its denoised estimate, so it remains at the correct noise level for the next DDPM transition.
Non-finite states, objectives, residuals, or gradients terminate the affected assembly with failure.
Failure to find a feasible iterate within the budget also terminates that assembly, without an infeasible fallback.
Failed assemblies count as failures in all binary success metrics.

\Cref{alg:motion-pcdpp} gives the complete procedure.
Each completed outer transition uses one guidance backward pass and at most \(M\) inner-optimisation backward passes, giving at most \(K(1+M)=5050\) backward passes per assembly under the reported settings.
Early acceptance or failure reduces this count; feasibility checks additionally require forward computations.
This counts joint backward evaluations, not separate passes for each person; at the final transition, inner gradients pass through the decoder but not through a denoiser, since \(\mathcal D_K=\mathrm{Id}\).
The solver does not certify convergence or global distance optimality.
Feasibility covers only the stated goal constraints and capsule proxy at checked times, not continuous-time clearance or the complete kinematic-validity audit.

\input{_algo_baselines}

\subsubsection{Additional Experimental Evaluations}\label{app:motion_evaluation}
\begin{figure}[t]
    \centering
    \includegraphics[width=0.95\linewidth]{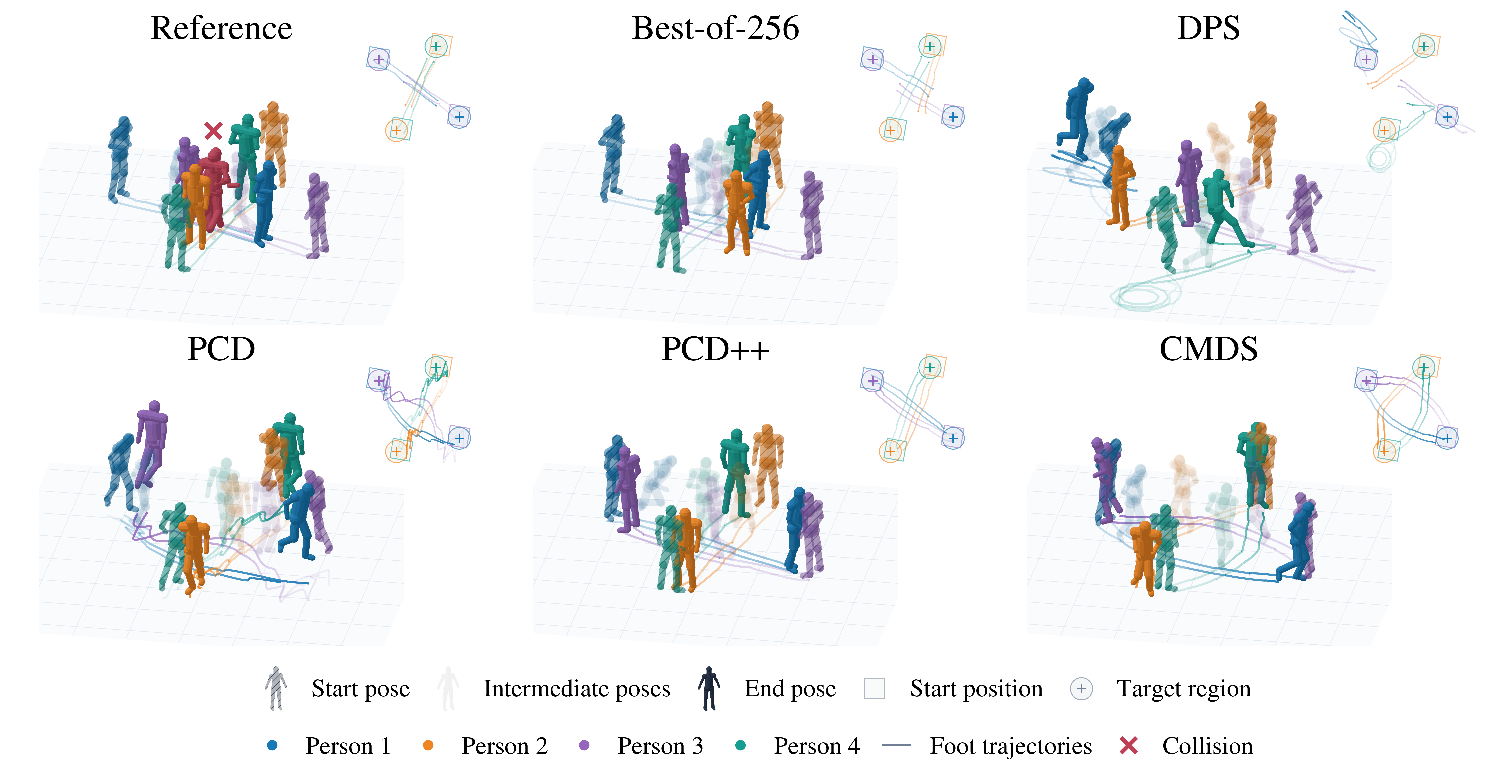}
    \caption{
        \textbf{Extended qualitative comparison of four-person motion generation.}
        We compare uncontrolled diffusion, Best-of-256, DPS, PCD, PCD++, and \method{} on the same four-person task shown in the main paper.
        The prompts are \textit{``A person moves forward at a jogging pace.''} for Person~1 (\figPone) and Person~4 (\figPfour), and
        \textit{``A person walks forward.''} for Person~2 (\figPtwo) and Person~3 (\figPthree).
        Top-down insets show starts and target regions.
        Solid curves show foot trajectories; intermediate poses are faded, while start and final poses are opaque.
        Red crosses mark inter-person collisions.
        In this example, uncontrolled diffusion, Best-of-256, DPS, and PCD do not satisfy the full joint goal-and-collision constraint, whereas PCD++ and \method{} reach all goals without inter-person collisions.
    }
    \label{fig:mdm_appendix}
    \label{fig:mdm_appedix}
\end{figure}

Evaluation uses \(24\) fixed scenes drawn from three training action families: walking, jogging, and crouched walking.
Thirteen scenes are homogeneous and eleven mix families.
Every method is evaluated with five sampling seeds, giving \(120\) attempts per method.
These seeds measure sampling variability for one controller, not training variability or held-out layout generalisation.
Main figures compare uncontrolled MDM, PCD++, and \method{}; the appendix comparison also includes Best-of-256, DPS, and PCD.
Matched views use the same sampling seed, identical scene conditions, and a common camera.
Capsule poses and footpaths show body clearance and goal progress, with faint intermediate poses and emphasised starts and endpoints.
Uncontrolled MDM supplies the visual prior reference; \cref{tab:n4_coordination} aggregates the full evaluation.
\Cref{fig:mdm_appendix} extends the qualitative comparison to all evaluated baselines.

\begin{table*}[t]
    \centering
    \caption{
        \textbf{Four-person locomotion coordination.}
        All methods use the same \(24\) scenes with walking, jogging, and crouched walking, evaluated over five sampling seeds (\(120\) attempts per method).
        Entries are mean \(\pm\) sample standard deviation across the five seed summaries.
        Binary rates weight all \(24\) attempts per seed equally, including generation failures as zero successes; continuous geometry and time use available generated outputs.
        Timing excludes failed-attempt costs.
        Bold denotes the best mean, including ties.
    }
    \label{tab:n4_coordination}
    \small
    \setlength{\tabcolsep}{4pt}
    \resizebox{\textwidth}{!}{%
        \begin{tabular}{lrrrrrrr}
            \toprule
            \toprule
            Method & \shortstack{Joint success $\uparrow$\\(\%)}
                   & \shortstack{Collision-free $\uparrow$\\(\%)}
                   & \shortstack{Peak body overlap $\downarrow$\\(cm)}
                   & \shortstack{All-person goals $\uparrow$\\(\%)}
                   & \shortstack{Goal error $\downarrow$\\(m)}
                   & \shortstack{Kinematic validity $\uparrow$\\(\%)}
                   & \shortstack{Time $\downarrow$\\(s/assembly)} \\
            \midrule

            Best-of-256
            & \(0.0\,{\color{gray}\scriptstyle\pm0.0}\)
            & \(16.7\,{\color{gray}\scriptstyle\pm5.9}\)
            & \(8.860\,{\color{gray}\scriptstyle\pm1.086}\)
            & \(0.0\,{\color{gray}\scriptstyle\pm0.0}\)
            & \(2.456\,{\color{gray}\scriptstyle\pm0.146}\)
            & \(83.3\,{\color{gray}\scriptstyle\pm6.6}\)
            & \(69.895\,{\color{gray}\scriptstyle\pm0.292}\) \\

            DPS
            & \(0.0\,{\color{gray}\scriptstyle\pm0.0}\)
            & \(61.7\,{\color{gray}\scriptstyle\pm10.8}\)
            & \(3.335\,{\color{gray}\scriptstyle\pm1.049}\)
            & \(0.0\,{\color{gray}\scriptstyle\pm0.0}\)
            & \(3.751\,{\color{gray}\scriptstyle\pm0.284}\)
            & \(57.5\,{\color{gray}\scriptstyle\pm6.8}\)
            & \(8.566\,{\color{gray}\scriptstyle\pm0.018}\) \\

            PCD
            & \(4.2\,{\color{gray}\scriptstyle\pm2.9}\)
            & \(8.3\,{\color{gray}\scriptstyle\pm2.9}\)
            & \(19.002\,{\color{gray}\scriptstyle\pm1.865}\)
            & \(\mathbf{100.0}\,{\color{gray}\scriptstyle\pm0.0}\)
            & \(0.500\,{\color{gray}\scriptstyle\pm0.000}\)
            & \(27.5\,{\color{gray}\scriptstyle\pm4.8}\)
            & \(9.985\,{\color{gray}\scriptstyle\pm0.064}\) \\

            PCD++
            & \(75.0\,{\color{gray}\scriptstyle\pm6.6}\)
            & \(\mathbf{98.3}\,{\color{gray}\scriptstyle\pm2.3}\)
            & \(\mathbf{0.000}\,{\color{gray}\scriptstyle\pm0.000}\)
            & \(98.3\,{\color{gray}\scriptstyle\pm2.3}\)
            & \(0.487\,{\color{gray}\scriptstyle\pm0.001}\)
            & \(75.0\,{\color{gray}\scriptstyle\pm6.6}\)
            & \(145.335\,{\color{gray}\scriptstyle\pm4.460}\) \\

            \midrule

            \rowcolor{pastelblue!45}
            \method{}
            & \(\mathbf{90.0}\,{\color{gray}\scriptstyle\pm6.3}\)
            & \(90.0\,{\color{gray}\scriptstyle\pm6.3}\)
            & \(0.740\,{\color{gray}\scriptstyle\pm0.351}\)
            & \(\mathbf{100.0}\,{\color{gray}\scriptstyle\pm0.0}\)
            & \(\mathbf{0.158}\,{\color{gray}\scriptstyle\pm0.008}\)
            & \(\mathbf{100.0}\,{\color{gray}\scriptstyle\pm0.0}\)
            & \(\mathbf{0.541}\,{\color{gray}\scriptstyle\pm0.002}\) \\

            \bottomrule
        \end{tabular}%
    }
\end{table*}

%% file: _algo_baselines.tex
\begin{algorithm}[t]
    \caption{PCD++ with a DDPM rollout and joint Tweedie projection}
    \label{alg:motion-pcdpp}
    \footnotesize
    \begin{algorithmic}[1]
        \Require frozen DDPM maps \(\{F_\ell^{\mathrm b}\}_{\ell=0}^{K-1}\), with \(\varsigma_{K-1}=0\);
        clean-estimate maps \(\{\mathcal D_\ell\}_{\ell=0}^{K}\), with \(\mathcal D_K=\mathrm{Id}\);
        assembly \(\varphi\), task \(\kappa\), and baseline reward \(r_\kappa^{\mathrm{base}}\);
        constraint residual \(V_\kappa\) and guarded penalty \(\mathcal P_\kappa\);
        guidance strength \(\eta_{\mathrm g}\); maximum inner updates \(M\);
        Adam step size \(\eta_{\mathrm p}\); penalty parameters \(\rho_0,\rho_{\max}\)
        \Statex \textbf{Conventions:} optimise only active coordinates; keep padding zero.
        Any non-finite state, objective, residual, or gradient returns \textsc{ProjectionFailure}.
        \State Sample \(X_0^i\sim\mathcal N(0,I)\) and independent rollout noise
        \(\Xi_\ell^i\sim\mathcal N(0,I)\), for all \(i,\ell\), on active coordinates
        \For{\(\ell=0,\ldots,K-1\)}
            \algphase{1. Native DDPM proposal and DPS guidance}
            \State \(\hat X_{1\mid\ell}\gets\mathcal D_\ell(X_\ell),\quad
                \hat Y_{1\mid\ell}\gets\varphi(\hat X_{1\mid\ell})\)
            \State \(v_\ell\gets\nabla_{X_\ell}
                r_\kappa^{\mathrm{base}}(\hat Y_{1\mid\ell})\)
            \bluecomment{Differentiate through the frozen denoiser with respect to its input}
            \State \(\widetilde X_{\ell+1}\gets\operatorname{stopgrad}
                \bigl(F_\ell^{\mathrm b}(X_\ell,\Xi_\ell)+\eta_{\mathrm g}v_\ell\bigr)\)
            \bluecomment{Also guide at \(\ell=K-1\); no variance or training-reward multiplier}
            \algphase{2. Joint optimisation of the proposed state}
            \State \(Z^{(0)}\gets\widetilde X_{\ell+1},\quad\rho\gets\rho_0\);
                initialise fresh Adam state
            \For{\(m=0,\ldots,M\)}
                \State \(\hat Y^{(m)}\gets
                    \varphi\bigl(\mathcal D_{\ell+1}(Z^{(m)})\bigr)\)
                \bluecomment{Fresh clean estimate at the candidate's next noise level; \(\mathcal D_K=\mathrm{Id}\)}
                \State \(q_m\gets V_\kappa(\hat Y^{(m)})\)
                \If{\(q_m=0\)}
                    \State \(X_{\ell+1}\gets\operatorname{stopgrad}(Z^{(m)})\)
                    \bluecomment{Keep the corrected state, not its denoised estimate}
                    \State \textbf{break}
                \EndIf
                \If{\(m=M\)}
                    \State \Return \textsc{ProjectionFailure}
                    \bluecomment{Terminate this assembly; no infeasible fallback}
                \EndIf
                \If{\(m=0\)}
                    \State \(q_{\mathrm{ref}}\gets q_m\)
                \ElsIf{\(m\bmod 5=0\)}
                    \If{\(q_m\geq0.95\,q_{\mathrm{ref}}\)}
                        \State \(\rho\gets\min(10\rho,\rho_{\max})\)
                    \EndIf
                    \State \(q_{\mathrm{ref}}\gets q_m\)
                \EndIf
                \State Compute detached guards \(\omega_{\mathrm g}^{(m)},\omega_{\mathrm c}^{(m)}\)
                    from \(\hat Y^{(m)}\) using \cref{eq:motion-pcdpp-guards}
                \State Define \(\displaystyle
                    \mathcal L_m(Z):=\tfrac12
                    \|Z-\widetilde X_{\ell+1}\|_{\mathrm{active}}^2
                    +\rho\,\mathcal P_\kappa\!\left(
                        \varphi(\mathcal D_{\ell+1}(Z));
                        \omega_{\mathrm g}^{(m)},\omega_{\mathrm c}^{(m)}\right)\)
                \State \(G_m\gets
                    \left.\nabla_Z\mathcal L_m(Z)\right|_{Z=Z^{(m)}}\)
                \bluecomment{Hold the guards fixed; one backward pass per inner update}
                \State \(Z^{(m+1)}\gets
                    \Call{AdamStep}{Z^{(m)},G_m,\eta_{\mathrm p}}\)
            \EndFor
        \EndFor
        \State \Return \(X_K,\ \varphi(X_K)\)
        \bluecomment{At most \(K(1+M)\) backward passes; \(5050\) for the reported settings}
    \end{algorithmic}
\end{algorithm}